\documentclass{article}
\usepackage{multirow}
\usepackage[nonatbib,preprint]{neurips_2025}
\usepackage{hyperref}
\usepackage{url}
\usepackage[utf8]{inputenc}
\usepackage[T1]{fontenc}
\usepackage{hyperref}
\usepackage{latexsym,bm}
\usepackage[tbtags]{amsmath}
\usepackage{graphicx}
\usepackage{subfigure}
\usepackage{flushend}
\usepackage{amssymb}
\usepackage{stfloats}
\usepackage{cite}
\usepackage{stfloats}
\usepackage{comment}
\usepackage{algorithm}
\usepackage{algpseudocode}
\usepackage{array}
\usepackage{amsmath}
\usepackage{bm}
\usepackage{CJK}
\usepackage{setspace}
\usepackage{algorithm}
\usepackage{color}
\usepackage{verbatim}
\usepackage{multirow}
\usepackage{url}
\usepackage{amssymb}
\usepackage{floatflt}
\usepackage{wrapfig}

\usepackage{graphicx}
\graphicspath{ {./figures/} }

\title{Likelihood-Free Variational Autoencoders}

\author{
{
Chen Xu\textsuperscript{1}, \
Qiang Wang\textsuperscript{1}, \
Lijun Sun\textsuperscript{2} \thanks{Corresponding author.}
}\\
{\textsuperscript{1}Beijing University of Posts and Telecommunications,\textsuperscript{2}McGill University}\\
{
\textsuperscript{1}\{xuchen; wangq\}@bupt.edu.cn,
\textsuperscript{2}lijun.sun@mcgill.ca
}
}

\begin{document}

\maketitle

\begin{abstract}
Variational Autoencoders (VAEs) typically rely on a probabilistic decoder with a predefined likelihood, most commonly an isotropic Gaussian, to model the data conditional on latent variables. While convenient for optimization, this choice often leads to likelihood misspecification, resulting in blurry reconstructions and poor data fidelity, especially for high-dimensional data such as images. In this work, we propose \textit{EnVAE}, a novel likelihood-free generative framework that has a deterministic decoder and employs the energy score---a proper scoring rule---to build the reconstruction loss. This enables likelihood-free inference without requiring explicit parametric density functions. To address the computational inefficiency of the energy score, we introduce a fast variant, \textit{FEnVAE}, based on the local smoothness of the decoder and the sharpness of the posterior distribution of latent variables. This yields an efficient single-sample training objective that integrates seamlessly into existing VAE pipelines with minimal overhead. Empirical results on standard benchmarks demonstrate that \textit{EnVAE} achieves superior reconstruction and generation quality compared to likelihood-based baselines. Our framework offers a general, scalable, and statistically principled alternative for flexible and nonparametric distribution learning in generative modeling.
\end{abstract}

\section{Introduction}

Deep generative models are a cornerstone of modern machine learning. Among them, Variational Autoencoders (VAEs) \cite{kingma2013auto} have emerged as a powerful and flexible framework due to their principled probabilistic foundation, training stability, and capacity for learning structured latent representations. Unlike Generative Adversarial Networks (GANs) \cite{goodfellow2014generative} and flow-based models \cite{rezende2015variational}, VAEs provide an explicit variational formulation, making them a versatile tool for a wide range of applications including image synthesis \cite{qian2019make, huang2018introvae}, text-to-speech \cite{kim2021conditional, ren2021portaspeech}, text-to-image generation \cite{gu2022vector}, time series modeling \cite{li2023causal, zhu2023markovian}, and disentangled representation learning \cite{higgins2017beta, chen2018isolating}.

Despite their theoretical appeal, VAEs often struggle to generate high-quality samples, particularly in high-dimensional domains \cite{an2024distributional, bredell2023explicitly}. This limitation stems from the structural nature of the Evidence Lower Bound (ELBO), which is typically optimized by assuming a simple, tractable likelihood---commonly an isotropic Gaussian---for the decoder. Reformulating the ELBO shows it minimizes the KL divergence between the joint distributions $q_{D,\phi}(\mathbf{x}, \mathbf{z})$ and $p_\theta(\mathbf{x}, \mathbf{z})$ \cite{kingma2019}. Owing to the inherent asymmetry of the KL divergence, regions where the distribution $q_{D,\phi}(\boldsymbol{x},\boldsymbol{z})$ exhibits high probability density but $p_\theta(\boldsymbol{x},\boldsymbol{z})$ does not are subjected to a substantial penalty, whereas the converse is not equivalently constrained. As a result, the model compensates by generating an overly dispersed distribution $p_\theta(\boldsymbol{x},\boldsymbol{z})$ and $p_\theta(\boldsymbol{x})$, thereby leading to the production of blurred images.

A wide range of methods have been proposed to address this issue. These efforts include the introduction of more expressive prior distributions \cite{casale2018gaussian,dorta2018structured,an2024distributional,park2019variational,vafaii2024poisson,zhao2020variational,srivastava2017autoencoding,kim2023t,takahashi2018student}, architectural innovations \cite{kim2021conditional,vahdat2020nvae, sonderby2016ladder, zhao2017learning, larsen2016autoencoding, pu2017adversarial}, and post-hoc refinement strategies \cite{xiao2020vaebm, luo2025energy,han2020joint,aneja2021contrastive}. While these approaches often lead to empirical gains, they largely address the symptoms---such as sample blurriness and over-smoothing---without resolving the core limitation: the structural inflexibility imposed by predefined likelihoods and the asymmetry of KL-divergence in ELBO-based training. The assumption of a Gaussian likelihood often implies pixel-wise independence and a unimodal distribution around the predicted mean, which may not accurately reflect the intricate dependencies and multimodal nature of real-world images.

Recent research in generative modeling has demonstrated that proper scoring rules \cite{gneiting2007strictly,pacchiardi2022likelihood} offer a robust alternative to likelihood-based objectives by enabling flexible distribution modeling through Monte Carlo sampling \cite{bouchacourt2016disco,gritsenko2020spectral, chen2024generative, shen2024engression, pacchiardi2024probabilistic}. These approaches have facilitated the development of likelihood-free inference for generative neural networks without requiring adversarial training \cite{pacchiardi2022likelihood,goodfellow2014generative}. Motivated by this insight, we propose \textit{EnVAE}, a likelihood-free Variational Autoencoder that replaces the probabilistic decoder with a deterministic one and leverages the energy score---a proper scoring rule---as a sample-based surrogate for the likelihood. To further improve efficiency, we introduce \textit{FEnVAE}, a computationally lightweight variant of \textit{EnVAE} that uses a local linear approximation to estimate the energy score from a single sample. This formulation significantly reduces computational overhead and integrates seamlessly with existing VAE architectures without architectural modifications. Both models are theoretically grounded and offer conceptually simple, interpretable alternatives to conventional VAEs. Our main contributions are as follows:

\begin{itemize}

\item We introduce \textit{EnVAE}, a novel VAE architecture featuring a deterministic decoder trained using the energy score as a loss function. This approach offers a departure from likelihood-based training and adversarial methods, with the potential to better model complex data distributions and achieve enhanced generative capabilities compared to standard VAEs.
	
\item We propose \textit{FEnVAE}, a fast single-sample variant that leverages local linear approximations to reduce computation. This lightweight method improves generation quality and integrates seamlessly with existing VAE models.

\item We demonstrate that both \textit{EnVAE} and \textit{FEnVAE} deliver improved reconstruction and generation performance, while providing reliable uncertainty quantification and detailed spectral feature recovery. We validate these advantages through extensive experiments and ablation studies. Furthermore, we analyze the computational efficiency and practical applicability of \textit{FEnVAE} to confirm its scalability and ease of integration.
	
\end{itemize}

\section{Related Work}\label{sec:method}

Our method builds upon the rich literature of likelihood-free inference (LFI), with a particular focus on generative models equipped with deterministic decoders, where defining a valid likelihood is often intractable or misaligned with perceptual quality. Traditional solutions frequently resort to adversarial training frameworks, such as Generative Adversarial Networks (GANs)\cite{goodfellow2014generative}, which eliminate the need for explicit likelihood modeling by introducing a discriminator to distinguish real from synthetic samples.

To overcome the limitations of standard VAEs in producing sharp and realistic images, numerous hybrid approaches have been proposed that integrate adversarial objectives into the VAE framework. These models aim to preserve the tractability of variational inference while improving generative quality. For example, Adversarial Autoencoders (AAEs) \cite{makhzani2015adversarial} substitute the KL divergence in the ELBO with a discriminator-based divergence to align the aggregated posterior with the prior. VAE-GAN \cite{larsen2016autoencoding} augments the reconstruction loss with a GAN-style discriminator that distinguishes real images from reconstructions. Other approaches such as BiGAN \cite{donahue2016adversarial} and ALI \cite{dumoulin2016adversarially} introduce bidirectional adversarial training to jointly learn encoder and decoder mappings. While these models have demonstrated empirical success, they often sacrifice the interpretability and stability of likelihood-based approaches due to adversarial training, and introduce significant optimization complexity \cite{salimans2016improved}.

An alternative to adversarial training in generative modeling involves the use of proper scoring rules \cite{gneiting2007strictly}, such as the energy score, which directly compare the generated distribution to the true data distribution without relying on adversarial frameworks. Training neural networks with the energy score offers a more stable and computationally efficient alternative to the often unstable and resource-intensive adversarial training employed by GANs \cite{pacchiardi2022likelihood}. These methods have been successfully applied in probabilistic time series forecasting \cite{pacchiardi2024probabilistic, zheng2024mvg}, multivariate weather prediction \cite{chen2024generative}, and extrapolation tasks in neural networks \cite{shen2024engression}.

Building on this line of research, we propose to integrate the energy score---a strictly proper scoring rule that measures the discrepancy between the generated and true data \cite{gneiting2007strictly,szekely2013energy}---directly into the VAE training process, thereby enabling likelihood-free optimization while preserving the benefits of amortized inference and scalability, and avoiding the instability typically associated with adversarial training. To the best of our knowledge, this is the first approach to unify proper scoring rules with variational autoencoders employing deterministic decoders, offering a principled and efficient alternative for training deep generative models in high-dimensional settings.

\section{Methodology}\label{sec:method}

n this section, we present \textit{EnVAE}, a likelihood-free variational autoencoder that employs the energy score as a sample-based reconstruction objective. We then introduce \textit{FEnVAE}, a computationally efficient variant that leverages a local linearity approximation to enable single-sample training. Finally, we outline key implementation details and provide architectural diagrams to illustrate the proposed models.

\subsection{Variational Autoencoder with Likelihood-free Inference}

We generalize the standard variational inference paradigm by adopting a likelihood-free perspective and propose \textit{EnVAE}. Instead of making explicit distribution assumptions as in the standard VAE, we adopt a likelihood-free inference framework \cite{pacchiardi2022likelihood} to replace the likelihood-based objective with the energy score to achieve flexible and robust output.

Unlike traditional probabilistic decoders that define an explicit likelihood model $p_\theta(\boldsymbol{x}|\boldsymbol{z})$, we consider a deterministic decoder represented by a mapping $g_\theta: \mathcal{Z} \to \mathcal{X}$. For continuous data, we assume $\mathcal{Z} = \mathbb{R}^m$, $\mathcal{X} = \mathbb{R}^n$, and often have $n\gg m$. Let $\boldsymbol{z} \sim \mathcal{N}(\boldsymbol{\mu}_z, \boldsymbol{\Sigma}_z)$, we denote the distribution of $\boldsymbol{x}$ by  $\boldsymbol{Q}_{\theta,\phi}(\boldsymbol{x})$, from which samples are generated by transforming a sample $\boldsymbol{z}^*$ by $\boldsymbol{x}^* = g_\theta(\boldsymbol{z}^*)$. In this case, the conditional likelihood collapses to a Dirac delta: $p_\theta(\boldsymbol{x}|\boldsymbol{z}) = \delta(\boldsymbol{x} - g_\theta(\boldsymbol{z}))$ and the density of $\boldsymbol{Q}$ cannot be evaluated.  This formulation thus eliminates the need for a parametric observation model, rendering the standard ELBO inapplicable since $ \log p_\theta(\boldsymbol{x}|\boldsymbol{z}) $ is undefined for a delta distribution.
Thus, instead of maximizing a log-likelihood, we follow \cite{pacchiardi2024probabilistic,shen2024engression,chen2024generative} and  optimize a proper scoring rule $S(\boldsymbol{Q}, \boldsymbol{x})$, which quantifies the quality of a predicted distribution $ \boldsymbol{Q}$ given an observed outcome $\boldsymbol{x}$. The proper scoring rules are minimized in expectation if and only if $ \boldsymbol{Q} $ matches the true data distribution of $ \boldsymbol{x} $~\cite{gneiting2007strictly}. To assess the quality of this sample-based distribution, we employ the energy score, a proper scoring rule that compares a predicted distribution $ \boldsymbol{P} $ over a space $ \mathcal{X} $ against an observed point $ \boldsymbol{x} \in \mathcal{X} $:
\begin{equation}
S_E(\boldsymbol{P},\boldsymbol{x}) = \mathbb{E}_{\boldsymbol{X} \sim \boldsymbol{P}}[\|\boldsymbol{X} - \boldsymbol{x}\|_2^\beta] - \frac{1}{2}\mathbb{E}_{\boldsymbol{X}, \boldsymbol{X}' \sim \boldsymbol{P}}[\|\boldsymbol{X} - \boldsymbol{X}'\|_2^\beta],
\end{equation}
where $ \beta \in (0, 2] $ controls sensitivity to outliers (typically $ \beta = 1 $) and when $\beta=2$, it degenerates into a Euclidean distance under independent Gaussian distributions (see Appendix~\ref{A.E}). The energy score essentially quantifies the discrepancy between the generated samples $\boldsymbol{X}\sim \boldsymbol{P}$ and the observed data $\boldsymbol{x}$. Intuitively, the first term encourages predictions to be close to the observed outcome $ \boldsymbol{x} $, while the second term discourages overly concentrated distributions by penalizing low sample diversity. Additionally, $ S_E $ is a strictly proper scoring rule, meaning that the expected score is uniquely minimized when $ \boldsymbol{P} $ equals the true data distribution~\cite{gneiting2007strictly}. This offers a principled, likelihood-free alternative to log-likelihood maximization. For the sample-based predictive distribution $ \boldsymbol{Q}_{\theta,\phi}(\boldsymbol{x}) $, the energy score becomes:
\begin{equation}
S_E(\boldsymbol{Q}_{\theta,\phi}(\boldsymbol{x}), \boldsymbol{x}) = \mathbb{E}_{\boldsymbol{z} \sim q_\phi(\boldsymbol{z}|\boldsymbol{x})}[\|g_\theta(\boldsymbol{z}) - \boldsymbol{x}\|_2^\beta] - \frac{1}{2}\mathbb{E}_{\boldsymbol{z}, \boldsymbol{z}' \sim q_\phi(\boldsymbol{z}|\boldsymbol{x})}[\|g_\theta(\boldsymbol{z}) - g_\theta(\boldsymbol{z}')\|_2^\beta].
\end{equation}
This can be estimated using $M$ generated samples $\{\boldsymbol{x}^*_{i}\}_{i=1}^M$ as:
\begin{equation}\label{eq:energyscore}
\hat{S}_E = \frac{1}{M} \sum_{i=1}^M \|\boldsymbol{x}^*_{i} - \boldsymbol{x}\|_2^\beta - \frac{1}{2M(M-1)} \sum_{i=j}^M \sum_{j:j\neq i} \|\boldsymbol{x}^*_{i} - \boldsymbol{x}^*_{j}\|_2^\beta,
\end{equation}
where $\boldsymbol{x}^*_{i}=g_\theta(\boldsymbol{z}_i)$, $\boldsymbol{z}_i\sim q_\phi(\boldsymbol{z}|\boldsymbol{x})$ for $i=1,\ldots,M$. Minimizing $S_E$ encourages accuracy (reconstructions near the true $\boldsymbol{x}$) while also encouraging the distribution of reconstructions to be well clustered. By minimizing the energy score, we aim to learn a generative model that accurately reflects the underlying data distribution.

Although the energy score balances reconstruction accuracy and dispersion, without further regularization, the encoder’s output distribution $q_\phi(\boldsymbol{z}| \boldsymbol{x})$ may drift arbitrarily from the prior, leading to overfitting or latent collapse.
To prevent this,  we still retain a KL term $\mathcal{D}_{\text{KL}}[q(\boldsymbol{z}|\boldsymbol{x})||p(\boldsymbol{z})]$ exactly as in the standard VAE. This means the new framework maintains the variational inference aspect for the latent variable, we are still encouraging the approximate posterior to align with the prior. Thus, in terms of latent space behavior, this model is similar to a $\beta$-VAE (with $\beta$ set by $\alpha$) or the Wasserstein Autoencoder (which also has a latent distribution matching term). The difference is purely in how we measure reconstruction quality. Thus, for a single data point $x$, the full \textit{EnVAE} objective is:
\begin{equation}\label{eq:EnVAE-full}
\mathcal{L}(\theta,\phi; \boldsymbol{x})
= S_E\bigl(\boldsymbol{Q}_{\theta,\phi}(\boldsymbol{x}), \boldsymbol{x}\bigr)
 + \alpha \mathcal{D}_{\mathrm{KL}}\bigl(q_\phi(\boldsymbol{z}| \boldsymbol{x}) \| p(\boldsymbol{z})\bigr),
\end{equation}
where $\alpha>0$ trades off the energy score against the KL regularizer.
In practice, using $ M $ Monte Carlo samples to approximate the energy score and the closed-form KL divergence for Gaussian posteriors, we obtain the following training objective:
\begin{align}\label{eq:fullloss}
\hat{\mathcal{L}}(\theta,\phi; \boldsymbol{x})
= {\frac{1}{M}\sum_{i=1}^M \lVert \boldsymbol{x}_i^* - \boldsymbol{x}\rVert_2^\beta
- \frac{1}{2M(M-1)}\sum_{i=1}^M\sum_{j:j\neq i}\lVert \boldsymbol{x}_i^* - \boldsymbol{x}_j^*\rVert_2^\beta}
 + \alpha \mathcal{D}_{\mathrm{KL}}\bigl(q_\phi(\boldsymbol{z}| \boldsymbol{x}) \| p(\boldsymbol{z})\bigr).
\end{align}
In summary, this approach minimizes a statistical distance, the energy distance, between the model’s output distribution and the data, rather than optimizing a lower bound on likelihood. Unlike ELBO-based objectives, which can suffer from mode-covering due to the KL divergence and reliance on explicit likelihood forms, the energy score sidesteps these issues by relying only on model samples and a proper scoring rule.

\subsection{A Fast Single-sampling Variant Mediated by the Local Linearity Approximation}

While minimizing the energy score offers a flexible and principled alternative to likelihood-based training, it introduces significant computational overhead. This is primarily due to the intractability of the energy score, which requires approximation via $M$ Monte Carlo samples under the conventional formulation. To address this issue, we propose \textit{FEnVAE}, a fast single-sample variant of \textit{EnVAE}, in which a closed-form approximation of the energy score is derived by leveraging the local linearity of the generative function.

For observed data $\boldsymbol{x}$ and samples $\boldsymbol{ x}^*=g_\theta(\boldsymbol{z}^*)$ drawn from the target distribution $\boldsymbol{F}$ of output, we focus on the reconstruction loss term in Eq.~\eqref{eq:fullloss}, denoted as:
\begin{equation}\label{e:lossstar}
	{\mathcal{L}}'(\theta; \boldsymbol{x}) = \mathbb{E}_{\boldsymbol{z} \sim \mathcal{N}(\boldsymbol{\mu}_z,\boldsymbol{\Sigma}_z)}[{||{g_\theta(\boldsymbol{z})} - \boldsymbol{x}||^\beta}]  - \frac{1}{2}\mathbb{E}_{\boldsymbol{z}, \boldsymbol{z}' \sim \mathcal{N}(\boldsymbol{\mu}_z,\boldsymbol{\Sigma}_z)}[{{||{g_\theta(\boldsymbol{z})} - {g_\theta(\boldsymbol{z}')}||^\beta} }].
\end{equation}
Based on the manifold hypothesis, the function $ g_\theta(\cdot) $ of the VAE decoder can be viewed as a smooth mapping from the low-dimensional manifold (the latent space) to the high-dimensional data space\cite{tenenbaum2000global}. Within a sufficiently small neighborhood, the nonlinear geometry of the manifold can be approximated by a linear subspace, which is spanned by the Jacobian of $ g_\theta(\cdot) $ at that point \cite{lee2003smooth}. Thus, we expand the decoder’s generative function $ g_\theta(\boldsymbol{z}) $ at the point $ \boldsymbol{z} = \boldsymbol{\mu}_z $:
\begin{align}\label{e:tl}
	g_\theta(\boldsymbol{z}) = \sum_{n=0}^{\infty} \frac{1}{n!} \nabla_{\boldsymbol{z}}^{n} g_\theta(\boldsymbol{z}) \big|_{\boldsymbol{z} = \boldsymbol{\mu}_z} \left[ (\boldsymbol{z} - \boldsymbol{\mu}_z)^{\otimes n} \right]
\end{align}
where $\nabla_{\boldsymbol{z}}^n g_\theta(\boldsymbol{z}) \big|_{\boldsymbol{z} = \boldsymbol{\mu}_z}$ denotes the $n$-th order derivative tensor of $g_\theta$ evaluated at $\boldsymbol{z} = \boldsymbol{\mu}_z$, and $(\boldsymbol{z} - \boldsymbol{\mu}_z)^{\otimes n}$ denotes the $n$-fold tensor outer product. Note that $ \boldsymbol{\mu}_{z} $ is the mean of the posterior distribution $q_{\phi}(\boldsymbol{z}|\boldsymbol{x})$ inferred by the encoder given the input sample $ \boldsymbol{x} $.

We assume that $g_\theta(\boldsymbol{z})$ can be treated as a linear function in the neighborhood of $\boldsymbol{z} = \boldsymbol{\mu}_z$, and $g_\theta(\boldsymbol{z})$ can be simplified to its first-order Taylor approximation:
\begin{equation}
	g_\theta(\boldsymbol{z})  = g_\theta(\boldsymbol{\mu}_z) + \nabla_{\boldsymbol{z}} g_\theta(\boldsymbol{z}) \big|_{\boldsymbol{z} = \boldsymbol{\mu}_z} (\boldsymbol{z} - \boldsymbol{\mu}_z)+\mathcal{O}(\|\boldsymbol{z} - \boldsymbol{\mu}_z\|^2),\  \text{as} \ (\boldsymbol{z} - \boldsymbol{\mu}_z)\to \boldsymbol{0}\,
\end{equation}

In a VAE, the approximate posterior distribution  $q_{\phi}\left(\boldsymbol{z}|\boldsymbol{x}\right)$ is typically designed to be sufficiently sharp, with small posterior variance $\boldsymbol{\Sigma}_z$, so that samples drawn from this distribution are tightly concentrated around the mean  $\boldsymbol{\mu}_z$. As a result, the decoder output for a latent sample $\boldsymbol{z}^*$ close enough to $\boldsymbol{\mu}_z$ can be effectively approximated using a first-order Taylor expansion::
\begin{equation}\label{e:TL}
	\boldsymbol{ x}^*\approx  g_\theta(\boldsymbol{\mu}_z) + \nabla_{\boldsymbol{z}} g_\theta(\boldsymbol{z}) \big|_{\boldsymbol{z} = \boldsymbol{\mu}_z} (\boldsymbol{z}^* - \boldsymbol{\mu}_z)
	= g_\theta(\boldsymbol{\mu}_z) + \boldsymbol{J}_{\boldsymbol{\mu}_z}(\boldsymbol{z}^* - \boldsymbol{\mu}_z),
\end{equation}
where $\boldsymbol{J}_{\boldsymbol{\mu}_z} \in \mathbb{R}^{n\times m}$ denotes the first-order partial derivative (Jacobian matrix) of $g_\theta(\boldsymbol{z})$ at $\boldsymbol{z} = \boldsymbol{\mu}_z$. We then substitute the linearized form of the decoder output in Eq.~\eqref{e:TL} into the second term of Eq.~\eqref{e:lossstar}. The resulting expression for the reconstruction loss becomes:
\begin{align}
	{\mathcal{L}}'(\theta; \boldsymbol{x}) =\mathbb{E}_{\boldsymbol{z} \sim \mathcal{N}(\boldsymbol{\mu}_{z},\boldsymbol{\Sigma}_z)}[||{g_\theta(\boldsymbol{z})} - \boldsymbol{x}||^\beta] 	- \frac{1}{{2}}\mathbb{E}_{\boldsymbol{z}, \boldsymbol{z}' \sim \mathcal{N}(\boldsymbol{\mu}_{z},\boldsymbol{\Sigma}_z)}[||\boldsymbol{J}_{\boldsymbol{\mu}_z}(\boldsymbol{z} - \boldsymbol{z}')||^\beta].
\end{align}

Considering that the difference between two latent samples drawn from the posterior $\boldsymbol{z}_i^* - \boldsymbol{z}_j^*=\boldsymbol{\Sigma}_z^{\frac{1}{2}}(\boldsymbol{\epsilon}_i-\boldsymbol{\epsilon}_j)$ can be generated by ${\boldsymbol{\hat \epsilon}=\boldsymbol{\epsilon}_i-\boldsymbol{\epsilon}_j \sim \mathcal{N}(0,2\boldsymbol{I})}$, we have:
\begin{align}\label{e:LossJ}
	{\mathcal{L}}'(\theta; \boldsymbol{x}) &=\mathbb{E}_{\boldsymbol{z} \sim \mathcal{N}(\boldsymbol{\mu}_{z},\boldsymbol{\Sigma}_z)}[||{g_\theta(\boldsymbol{z})} - \boldsymbol{x}||^\beta] - \frac{1}{{2}}\mathbb{E}_{\boldsymbol{\hat \epsilon} \sim \mathcal{N}(0,2\boldsymbol{I})}[||\boldsymbol{J}_{\boldsymbol{\mu}_z}\boldsymbol{\Sigma}_z^{\frac{1}{2}}\boldsymbol{\hat \epsilon}||^\beta].
\end{align}

\begin{figure*}
\centering
\subfigure[ ]{\includegraphics[width=2.8in]{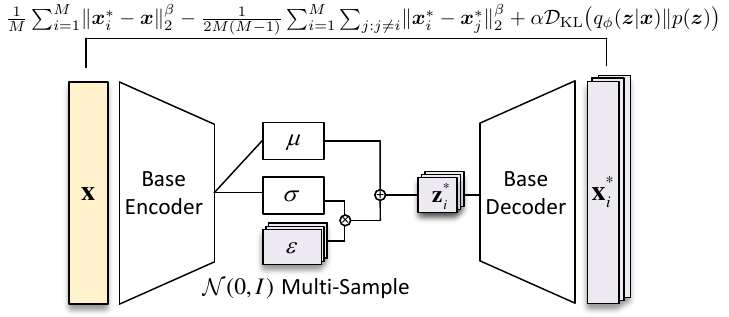}}
\subfigure[ ]{\includegraphics[width=2.5in]{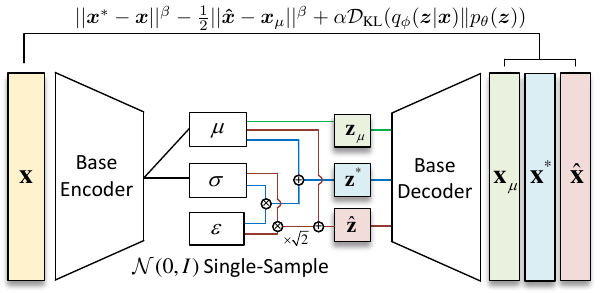}}
\caption{Implementations of \textit{EnVAE} and \textit{FEnVAE}. (a). The architecture of \textit{EnVAE}. Multiple samples of the latent variables are drawn and decoded. (b). The architecture of \textit{FEnVAE}. A single sample of the latent variable is drawn and decoded into three outputs.
}\label{Arc}
\end{figure*}

This first-order approximation eliminates the pairwise operations in Eq.~\eqref{e:lossstar}.
However, the Jacobian matrix poses computation and storage challenges. We replace the explicit computation of the Jacobian matrix back with the decoder $g_\theta(\boldsymbol{z})$, simplifying the computation while reintroducing the previously omitted higher-order term $\mathcal{O}(\|\boldsymbol{z} - \boldsymbol{\mu}_z\|^2)$, thereby locally restoring the nonlinear modeling capability. Based on Eq.~\eqref{e:TL}, we can obtain:
\begin{align}\label{e:Jz}
g_\theta(\boldsymbol{\mu}_{z}+\boldsymbol{\Sigma}_z^{\frac{1}{2}}\boldsymbol{\hat \epsilon})=  &  g_\theta(\boldsymbol{\mu}_{z}) + \nabla_{\boldsymbol{z}} g_\theta(\boldsymbol{z}) \big|_{\boldsymbol{z} = \boldsymbol{\mu}_{z}} \boldsymbol{\Sigma}_z^{\frac{1}{2}}\boldsymbol{\hat \epsilon}	= g_\theta(\boldsymbol{\mu}_{z}) + \boldsymbol{J}_{\boldsymbol{\mu}_z}\boldsymbol{\Sigma}_z^{\frac{1}{2}}\boldsymbol{\hat \epsilon}.
\end{align}
Substituting Eq.~\eqref{e:Jz} into Eq.~\eqref{e:LossJ}, we obtain the loss function mediated by the local linearity approximation:
\begin{align}\label{e:LossM}
	{\mathcal{L}}'(\theta; \boldsymbol{x}) &=\mathbb{E}_{\boldsymbol{z} \sim \mathcal{N}(\boldsymbol{\mu}_{z},\boldsymbol{\Sigma}_z)}[||{g_\theta(\boldsymbol{z})} - \boldsymbol{x}||^\beta]- \frac{1}{{2}}\mathbb{E}_{\boldsymbol{\hat \epsilon} \sim \mathcal{N}(0,2\boldsymbol{I})}[||g_\theta(\boldsymbol{\mu}_{z}+\boldsymbol{\Sigma}_z^{\frac{1}{2}}\boldsymbol{\hat \epsilon})-g_\theta(\boldsymbol{\mu}_{z})||^\beta].
\end{align}

In this context, we leverage the first-order Taylor approximation of $g_\theta(\boldsymbol{z})$ to transform the pairwise terms in the original energy score into the term represented explicitly using the variance of the latent variables. In Eq.~\eqref{e:LossM}, both terms require samples $\boldsymbol{\epsilon}$ generated from the standard Gaussian distribution, with $\boldsymbol{z}=\boldsymbol{\mu}_z+\boldsymbol{\Sigma}_z^{\frac{1}{2}}\boldsymbol{\epsilon}$ and $\boldsymbol{\hat \epsilon} =\sqrt{2}\boldsymbol{\epsilon}$. To reduce computational cost, we propose reusing the same standard Gaussian noise $\boldsymbol{\epsilon}$ in both terms in the reparameterization trick \cite{kingma2013auto}, effectively coupling the reconstruction and dispersion components through a shared random draw. This eliminates the need for pairwise sample generation and allows us to restructure the computation: rather than averaging over multiple samples in a single forward pass, we can achieve the same effect by using single samples across multiple iterations. This approach enables more frequent parameter updates and improves training efficiency. The resulting single-sample loss function is expressed as:
\begin{align}\label{e:Loss1}
	{\mathcal{L}^*}(\theta, \phi;\boldsymbol{x}) =\underbrace{||{g_\theta(\boldsymbol{z}^*)} - \boldsymbol{x}||^\beta}_{\text{Mean Loss}}	- \underbrace{{\frac{1}{2}}||g_\theta(\boldsymbol{\mu}_{z}+\sqrt{2}(\boldsymbol{ z}^*-\boldsymbol{\mu}_{z}))-g_\theta(\boldsymbol{\mu}_{z})||^\beta}_{\text{Uncertainty Loss}}+\underbrace{\alpha\mathcal{D}_\text{KL}(q_\phi(\boldsymbol{z}|\boldsymbol{x}) \| p_\theta(\boldsymbol{z}))}_{\text{KL Divergence Loss}},
\end{align}
where $\alpha$ is a tunable coefficient that balances the reconstruction fidelity and the latent regularization. While \textit{EnVAE} requires multiple samples to approximate the energy score, \textit{FEnVAE} explicitly constructs both the predictive mean and variability using a single latent sample and its structured perturbation. This formulation captures uncertainty in the output space without requiring pairwise evaluations, and transforms the original sampling-based objective into a closed-form surrogate based on the latent distribution's statistical properties. As a result, \textit{FEnVAE} improves perceptual modeling and accelerates convergence during training.

\subsection{Implementation of $\textit{EnVAE}$ and $\textit{FEnVAE}$}

Both \textit{EnVAE} and \textit{FEnVAE} are incorporated into the loss function as additional constraints, without altering the underlying model architecture. These modules are active only during training and introduce no additional computational overhead at inference time. In practice, we simplify the posterior distribution of the latent variables $p(\boldsymbol{z}|\boldsymbol{x})$ as an independent Gaussian distribution, e.g., $\boldsymbol{\Sigma}_z=\operatorname{diag}\{\sigma_1^2,\ldots,\sigma_m^2\}$.

For \textit{EnVAE}, training involves drawing multiple latent samples, followed by decoding and computing the loss according to Eq.~\eqref{eq:fullloss}, as illustrated in Figure \ref{Arc}(a).
In the case of \textit{FEnVAE}, the base model is extended to produce two additional outputs: one corresponding to the decoded mean of the latent distribution, and another corresponding to a perturbed latent sample whose standard deviation is scaled by  $\sqrt{2}$. The loss is then computed based on Eq.~\eqref{e:Loss1}, measuring the distance between these additional outputs, as shown in Figure \ref{Arc}(b).

\section{Experiment}\label{sec:results}

\begin{wraptable}{r}{0.55\textwidth}
\vspace{-0.6cm}
\caption{Different reconstruction loss performances.}
\vspace{0.2cm}
\label{tab:1}
\renewcommand {\arraystretch}{1.2}
\resizebox{0.55\textwidth}{!}{
\begin{tabular}{l c c c c c c c c c}
\hline
\hline
\multirow{2}{*}{Model} & \multicolumn{2}{c}{CelebA 64} & & \multicolumn{2}{c}{CIFAR-10} & & \multicolumn{2}{c}{LSUN Church 64} \\
\cline{2-3} \cline{5-6} \cline{8-9}
& $\text{FID}_{re}\downarrow$ & $\text{FID}_{gen}\downarrow$ & & $\text{FID}_{re}\downarrow$ & $\text{FID}_{gen}\downarrow$ & & $\text{FID}_{re}\downarrow$ & $\text{FID}_{gen}\downarrow$ \\
\hline
L2 & $94.826$ & $98.691$ & & $103.440$ & $107.630$ & & $163.361$ & $169.375$ \\
L1 & $98.798$ & $103.061$ & & $107.467$ & $109.322$ & & $177.322$ & $183.313$ \\
CE & $92.816$ & $95.858$ & & $100.471$ & $103.975$ & & $165.372$ & $170.236$ \\
SSIM & $89.871$ & $92.628$ & & $98.495$ & $99.530$ & & $157.442$ & $162.389$ \\
InfoVAE & $85.093$ & ${86.861}$ & & $99.520$ & $98.182$ & & $152.521$ & $155.461$ \\
FFL & $85.989$ & $89.233$ & & ${95.577}$ & $97.957$ & & ${150.272}$ & ${153.511}$ \\
Student-t & $90.573$ & $91.411$ & & $97.922$ & $101.019$ & & $160.126$ & $164.801$ \\
DistVAE & ${84.961}$ & ${87.126}$ & & $95.861$ & ${97.101}$ & & ${151.937}$ & ${153.981}$ \\
\hline
EnVAE (ours) & $\boldsymbol{75.236}$ & $\boldsymbol{78.941}$ & & $\boldsymbol{87.667}$ & $\boldsymbol{90.132}$ & & $\boldsymbol{139.632}$ & $\boldsymbol{143.064}$ \\
FEnVAE (ours) & $76.024$ & $80.012$ & & $88.940$ & $92.167$ & & $142.226$ & $145.613$ \\
\hline
Impr. (EnVAE) & $\boldsymbol{11.45\%}$ & $\boldsymbol{9.12\%}$ & & $\boldsymbol{8.28\%}$ & $\boldsymbol{7.74\%}$ & & $\boldsymbol{7.08\%}$ & $\boldsymbol{6.81\%}$ \\
Impr. (FEnVAE) & $10.52\%$ & $7.89\%$ & & $6.94\%$ & $5.08\%$ & & $5.35\%$ & $5.14\%$ \\
\hline
\end{tabular}}
\end{wraptable}
In this section, we assess the image reconstruction and generation performance of our proposed models, \textit{EnVAE} and \textit{FEnVAE}, in comparison with commonly used reconstruction losses, as well as the performance of large-scale VAE-based baseline models after integrating our proposed models (refer to Appendix~\ref{A.G}). The experiments are conducted on three datasets: CIFAR-10, CelebA 64, and LSUN Church 64. After integrating \textit{EnVAE} and \textit{FEnVAE}, the modified models are referred to as “+EnL” and “+FEnL,” respectively. Additionally, we compare the performance and efficiency of \textit{FEnVAE} with \textit{EnVAE} and Vanilla VAE, visualize the effectiveness of our model in variance modeling and capturing image frequency-domain features, and analyze the validity of the underlying assumptions of \textit{FEnVAE}. The code is available at \url{https://github.com/ChenXu02/EnVAE}.

\subsection{Image Generation and Reconstruction Analysis}

\begin{wraptable}{r}{0.55\textwidth}
\vspace{-0.6cm}
\caption{Comparison of performance after integration.}
\vspace{0.2cm}
\label{tab:2}
\renewcommand {\arraystretch}{1.2}
\resizebox{0.55\textwidth}{!}{
\begin{tabular}{l c c c c c c c c c}
\hline
\hline
\multirow{2}{*}{Model} & \multicolumn{2}{c}{CelebA 64} & & \multicolumn{2}{c}{CIFAR-10} & & \multicolumn{2}{c}{LSUN Church 64} \\
\cline{2-3} \cline{5-6} \cline{8-9}
& $\text{FID}_{re}\downarrow$ & $\text{FID}_{gen}\downarrow$ & & $\text{FID}_{re}\downarrow$ & $\text{FID}_{gen}\downarrow$ & & $\text{FID}_{re}\downarrow$ & $\text{FID}_{gen}\downarrow$ \\
\hline
DT-VAE & $24.163$ & $25.041$ & & $30.229$ & $30.912$ & & $25.620$ & $26.832$ \\
+ EnL & $22.792$ & $23.117$ & & $28.069$ & $29.761$ & & $23.533$ & $24.712$ \\
+ FEnL & $22.781$ & $24.512$ & & $27.641$ & $29.936$ & & $25.730$ & $25.986$ \\
\hline
NCP-VAE & $5.382$ & $5.763$ & & $24.492$ & $25.125$ & & $21.791$ & $22.691$ \\
+ EnL & $5.306$ & $5.504$ & & $23.011$ & $23.722$ & & $20.071$ & $20.914$ \\
+ FEnL & $5.267$ & $5.712$ & & $23.932$ & $23.411$ & & $21.551$ & $21.161$ \\
\hline
EC-VAE & $2.701$ & $2.775$ & & $5.242$ & $5.311$ & & $4.201$ & $4.332$ \\
+ EnL & $2.592$ & $2.665$ & & $5.247$ & $5.225$ & & $4.187$ & $4.193$ \\
+ FEnL & $2.615$ & $2.706$ & & $5.439$ & $5.296$ & & $4.152$ & $4.275$ \\
\hline
\end{tabular}}
\end{wraptable}
The quantitative evaluation of the reconstruction and generation performance of our proposed models, \textit{EnVAE} and \textit{FEnVAE}, with respect to different reconstruction loss terms is summarized in Table~\ref{tab:1}. The experimental results demonstrate that our likelihood-free models (\textit{EnVAE} and \textit{FEnVAE}) offer notable advantages over methods employing predefined distribution-based reconstruction losses and explicit frequency-domain constraints. While complex models such as InfoVAE, FFL, and DistVAE can alleviate certain types of distortion through predefined priors or frequency-domain regularization, their performance remains limited by the assumptions inherent in such priors. In contrast, our likelihood-free approaches overcome these limitations, leading to improved fidelity in the generated outputs.

We integrate our model into existing VAE-based frameworks, as shown in Table~\ref{tab:2}, and observe a consistent positive impact. Although the improvement is relatively modest—largely due to the inherent expressive capacity of these large-scale models—our likelihood-free reconstruction loss still contributes to measurable enhancements. This indicates that while post-processing with energy-based models (EBMs) can convert VAE outputs into a likelihood-free form, the intermediate representations produced by the VAE, along with constraints embedded in the loss function (e.g., the likelihood term), continue to hinder the model from reaching its full potential. By incorporating our approach, these suboptimal constraints can be relaxed or even eliminated, enabling more flexible and coherent output generation across the entire modeling pipeline.

\begin{wrapfigure}{r}{0.5\textwidth}
\centering
\vspace{-0.4cm}
\includegraphics[width=2.7in]{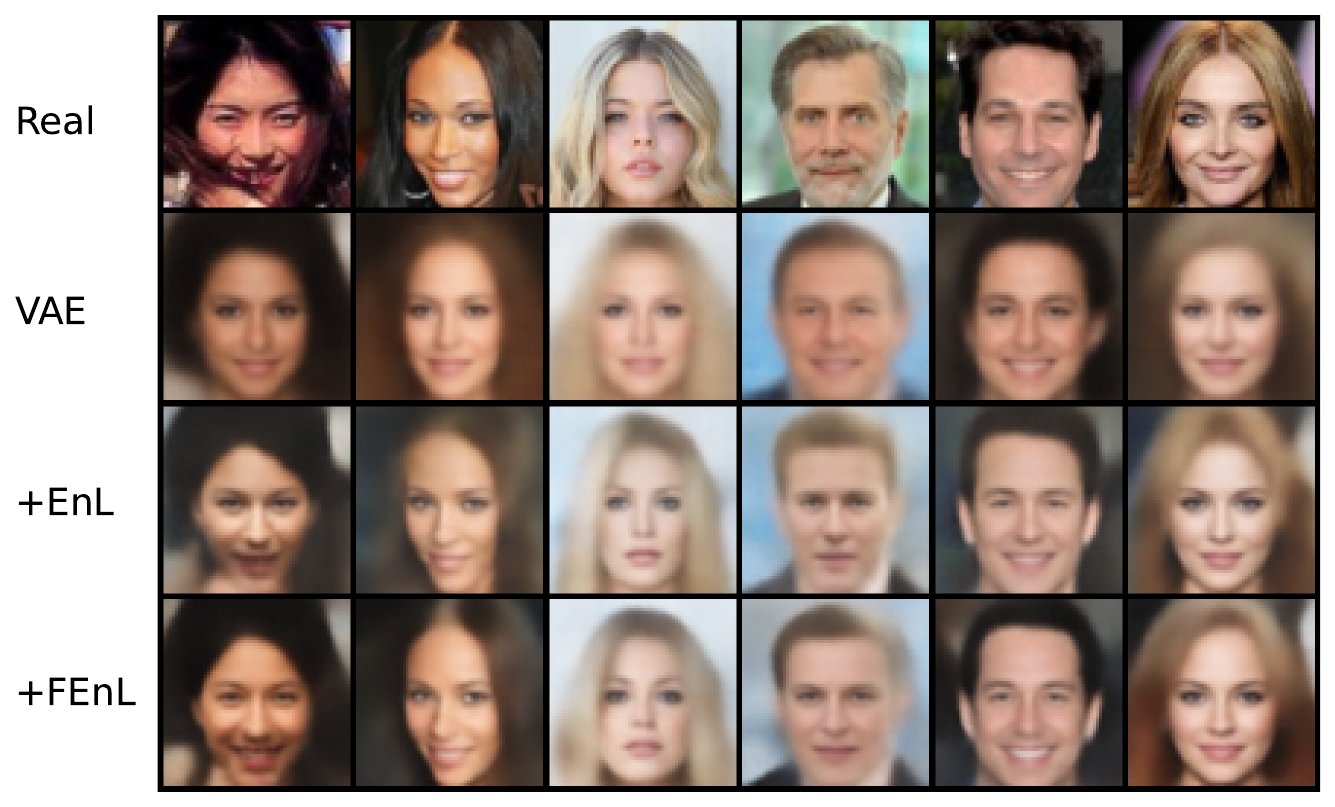}
\vspace{-0.2cm}
\caption{Reconstruction image samples.}
\label{fig:2}
\vspace{-0.3cm}
\end{wrapfigure}
\begin{wrapfigure}{r}{0.5\textwidth}
\centering
\vspace{-0.4cm}
\includegraphics[width=2.7in]{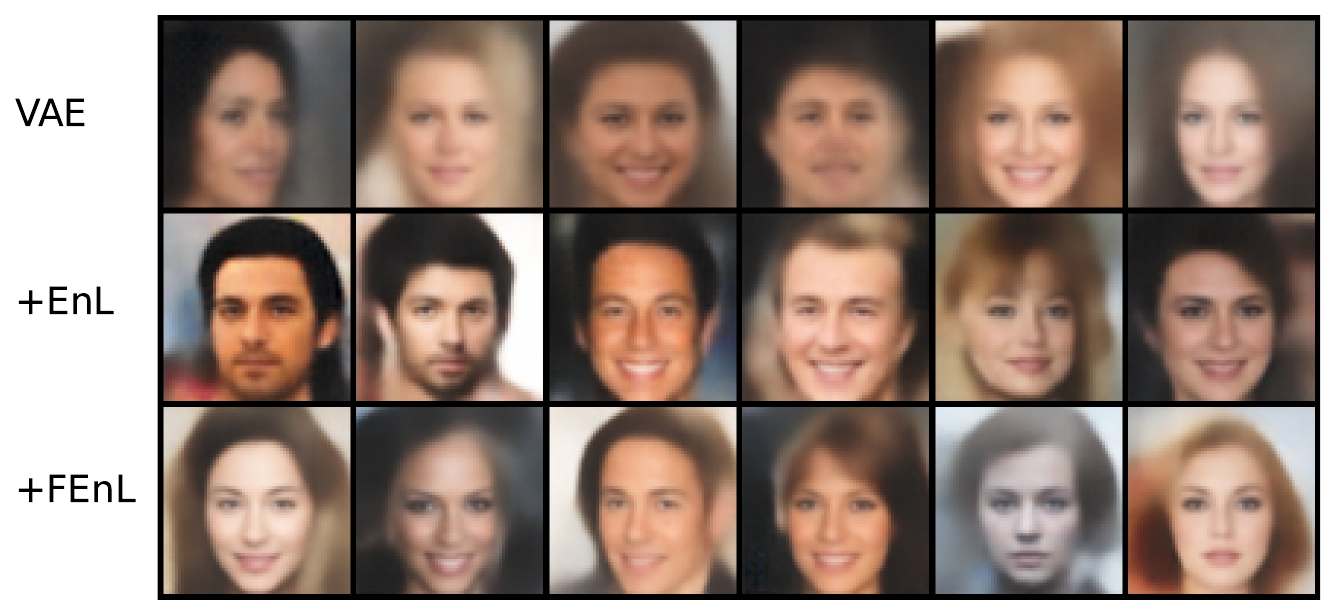}
\vspace{-0.2cm}
\caption{Generation image samples.}
\label{fig:3}
\vspace{-0.2cm}
\end{wrapfigure}

We visualized the reconstructed and generated images on the CelebA dataset based on the Vanilla VAE with our model plugin. As shown in Figure \ref{fig:2}, the reconstructions from the Vanilla VAE exhibit noticeable blurriness and a loss of identity-specific details. In contrast, our models significantly enhance the definition of facial features, improve lighting and shadow rendering, and maintain sharper contours, thereby preserving identity more effectively.
Figure~\ref{fig:3} displays images generated from latent samples. Consistent with the reconstruction results, our models produce faces with more realistic and coherent features, exhibiting improved facial structure and greater visual clarity. In contrast, images generated by the Vanilla VAE often lack fine details and appear noticeably blurry. These results highlight the effectiveness of our proposed approach in significantly enhancing the generative capabilities of VAEs, leading to higher-quality and more interpretable outputs. Additional results can be found in Appendix~\ref{Ari}, \ref{Agi}, and \ref{lsw}.

\subsection{Reconstruction and Residual Error Distribution Analysis}

\begin{figure*}
\vspace{-0.6cm}
\centering
\subfigure[ ]{\includegraphics[width=2.6in]{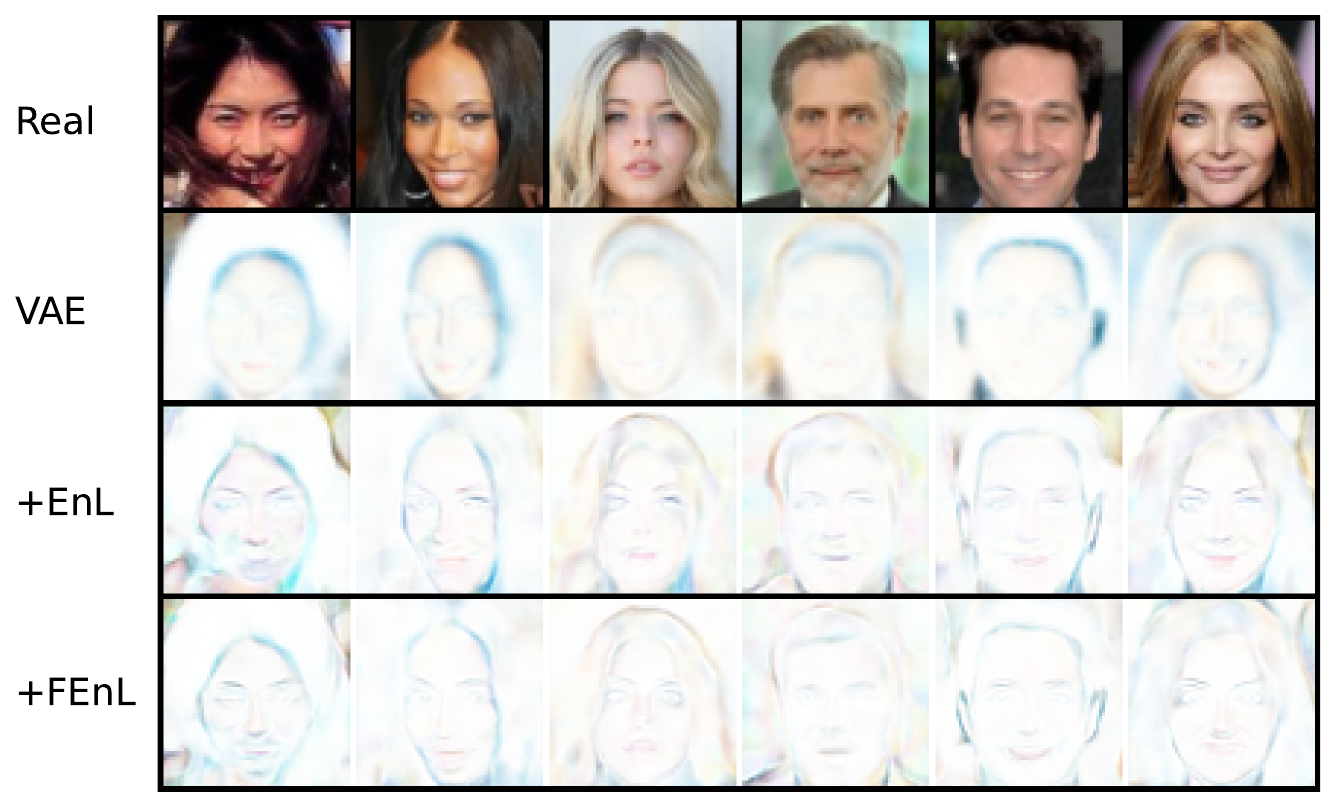}}
\subfigure[ ]{\includegraphics[width=2.7in]{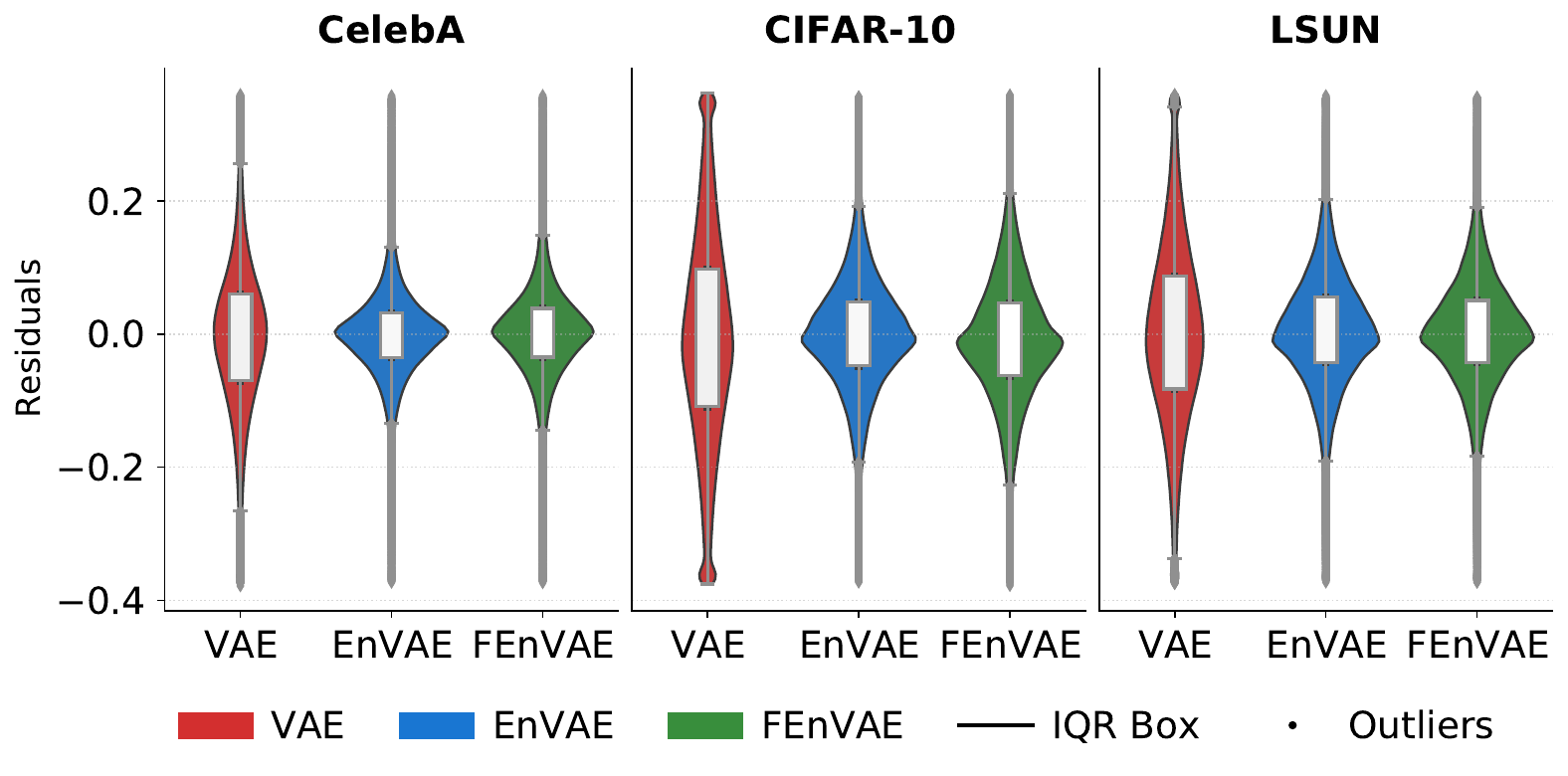}}
\vspace{-0.4cm}
\caption{Uncertainty visualization and residual distribution of reconstructed images. (a). Uncertainty visualization (RGB). (b). Pixel-wise normalized residual distribution of reconstructed images.}\label{fig:5}
\vspace{-0.3cm}
\end{figure*}

We evaluate pixel-wise reconstruction uncertainty by performing 50 independent latent space samplings for a given image using both the Vanilla VAE and our proposed models. Each sample is decoded into an image, and pixel-wise variance is computed across the 50 reconstructions, as illustrated in Figure~\ref{fig:5}(a).
The results reveal that, under identical latent sampling conditions, our models produce variance maps that are more refined and structurally coherent, with sharper boundaries and improved detail preservation. In contrast, the Vanilla VAE generates more diffuse variance maps that fail to capture key structural contours of the image.
These findings indicate that our likelihood-free modeling approach more effectively captures both low-frequency (i.e., low-variance regions) and high-frequency (i.e., high-variance regions) components, despite using the same latent dimensionality, model capacity, and neural architecture as the baseline VAE.

Furthermore, we present the pixel-wise residual (normalized) distribution of the $50$ reconstructed images sampled from the same latent variable distribution, as shown in Figure \ref{fig:5}.(b). Compared to the model Vanilla VAE, both \textit{EnVAE} and \textit{FEnVAE} exhibit enhanced reconstruction fidelity. On CIFAR-10 and LSUN, the Vanilla VAE shows a broad error range (-0.4 to 0.4), indicating instability in capturing fine-grained details. In contrast, \textit{EnVAE} significantly narrows this distribution, reducing tailing and increasing flexibility, concentrating errors closer to zero (-0.2 to 0.2), due to its likelihood-free assumption, which improves preservation of local features. The residual distribution of the \textit{FEnVAE} model across the three datasets is similar to that of \textit{EnVAE}, indicating that its local linear approximation does not have a significant impact on the residual distribution.

\subsection{Efficiency and Applicability Analysis of \textit{FEnVAE}}

\subsubsection{Comparison of Generation Performance and Computational Efficiency}

\begin{figure}
\centering
\includegraphics[width=5.5in]{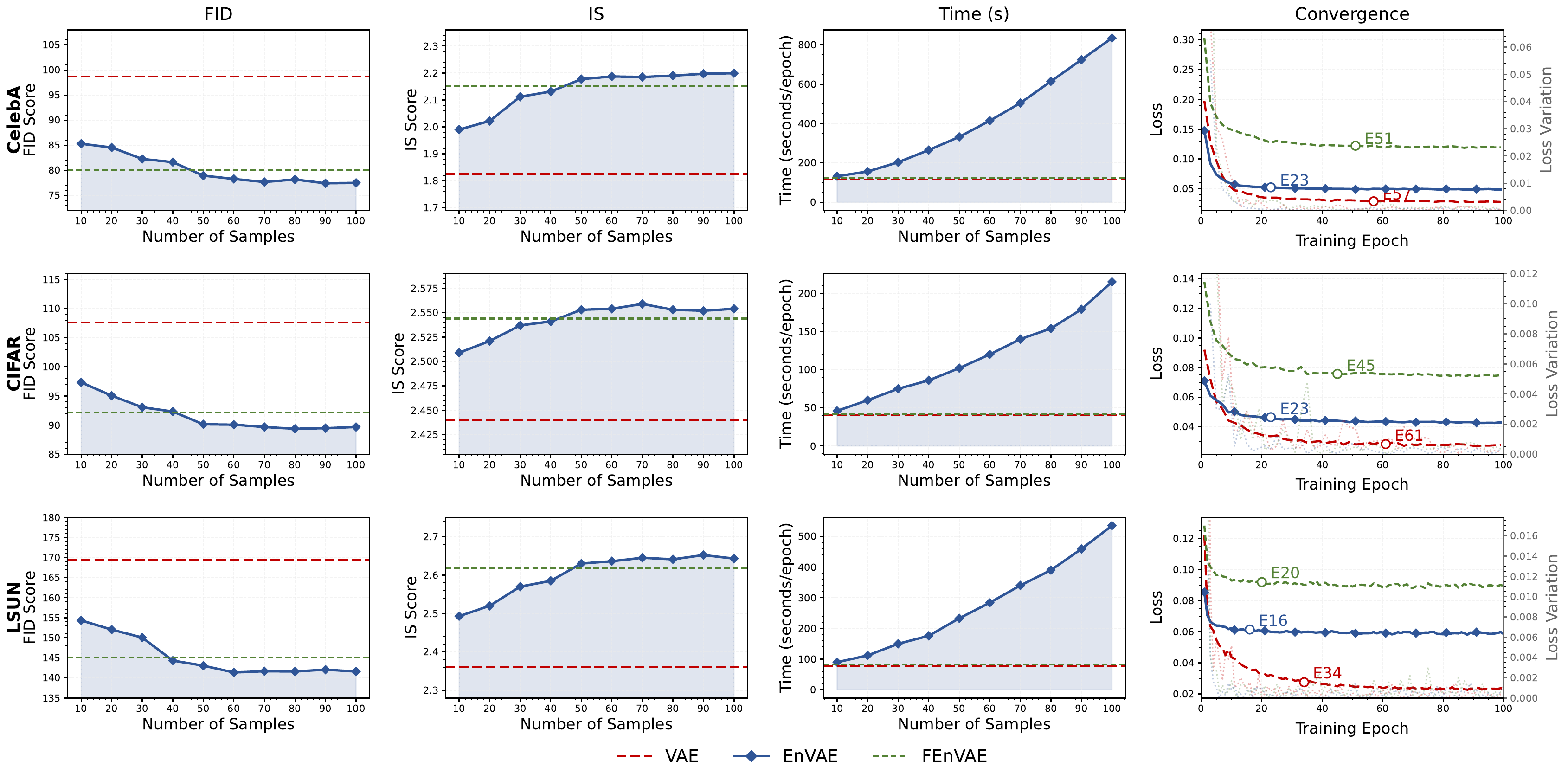}
\vspace{-0.7cm}
\caption{Comparison of performance and efficiency between \textit{EnVAE} and \textit{FEnVAE}}
\vspace{-0.4cm}
\label{fig:6}
\end{figure}
We evaluate \textit{EnVAE} across different sampling numbers and compare its training time, performance, and convergence with Vanilla VAE and \textit{FEnVAE}, as shown in Figure~\ref{fig:6}. As the number of samples increases, \textit{EnVAE} achieves lower FID and higher IS, reflecting improved generation quality and diversity. Even with a few samples, it outperforms the Vanilla VAE significantly. However, performance gains diminish as sampling increases, eventually plateauing. In contrast, \textit{FEnVAE} achieves near-optimal performance without requiring multiple samples, highlighting its efficiency.

As shown in the third column of Figure~\ref{fig:6}, \textit{EnVAE} incurs increasing computational overhead with more samples, scaling roughly linearly. In contrast, \textit{FEnVAE}, with its single-sample strategy, maintains constant overhead—only slightly higher than VAE due to two extra decoding steps. Regarding convergence (last column), \textit{EnVAE} converges fastest, Vanilla VAE slowest, and \textit{FEnVAE} lies in between. \textit{EnVAE}'s efficiency stems from optimizing multiple samples per step. Despite slightly less stable training (as seen from the dashed lines), likely due to the adversarial interplay of mean and uncertainty losses under the single-sample strategy, \textit{FEnVAE}'s epoch-wise speed and near-optimal performance make it the most efficient overall.

\subsubsection{Validity of Local Linear Approximation in \textit{FEnVAE}.}

\begin{figure*}
\centering
\subfigure[ ]{\includegraphics[width=1.35in]{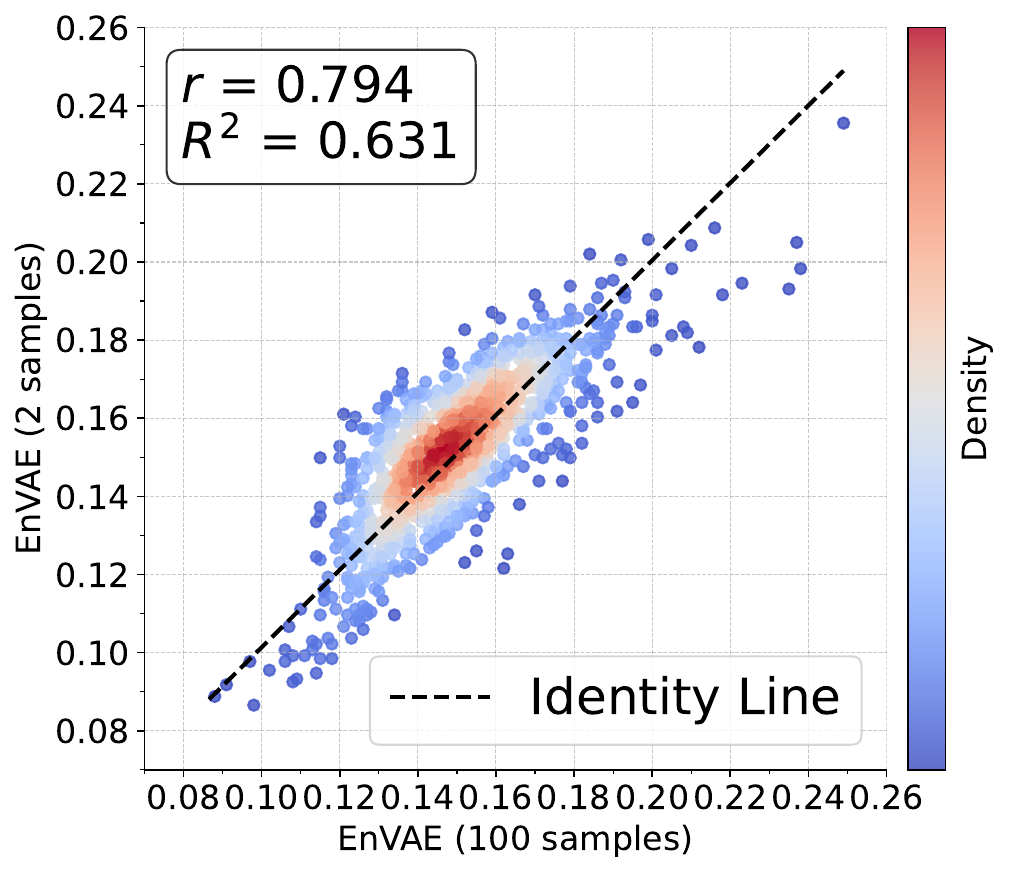}}
\subfigure[ ]{\includegraphics[width=1.35in]{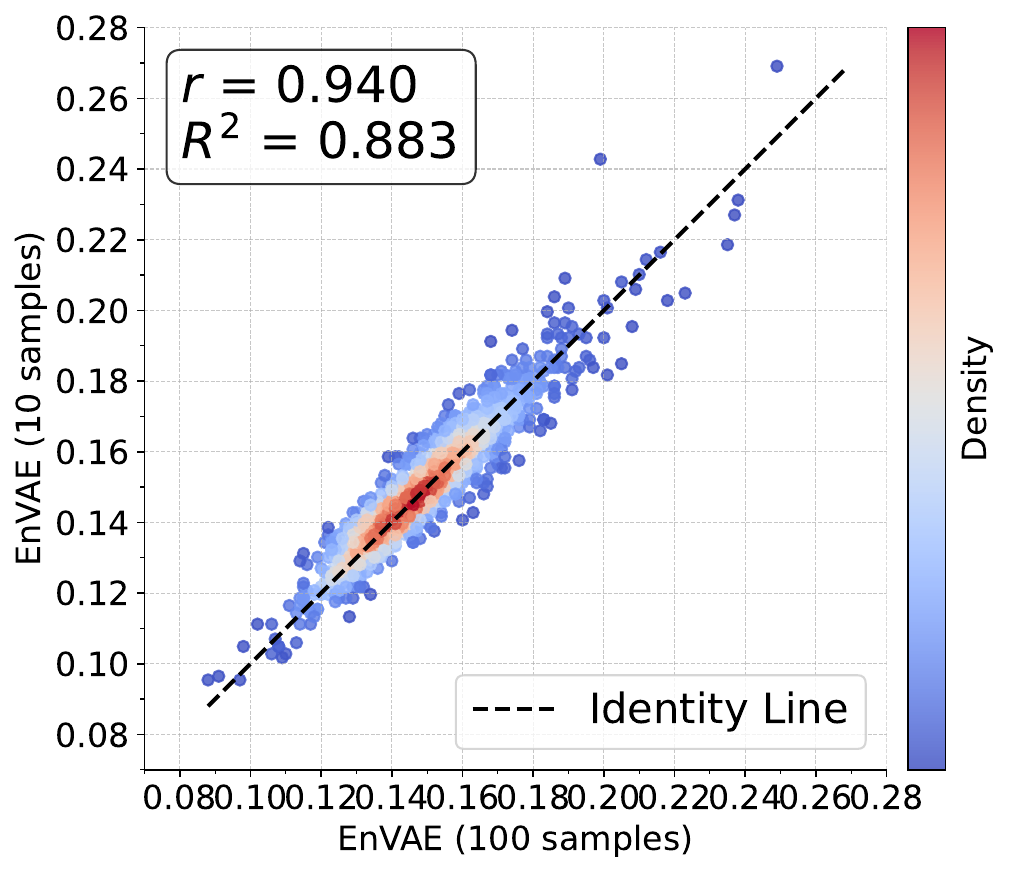}}
\subfigure[ ]{\includegraphics[width=1.35in]{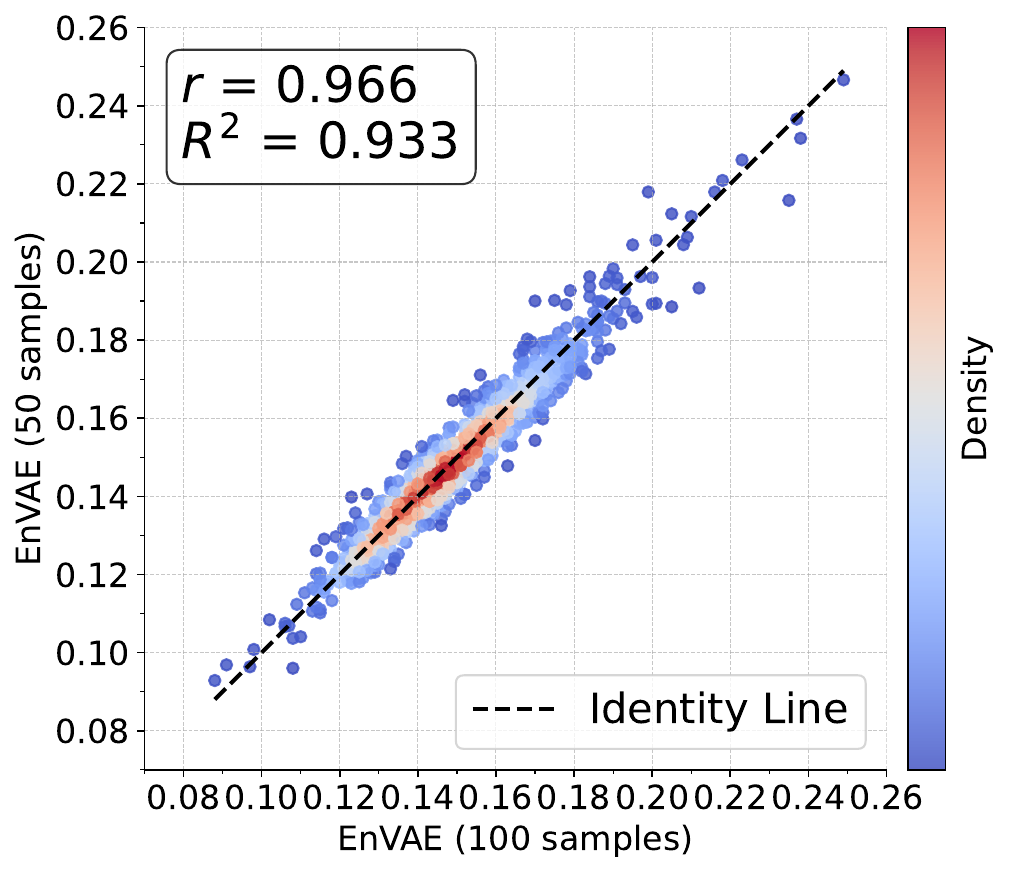}}
\subfigure[ ]{\includegraphics[width=1.35in]{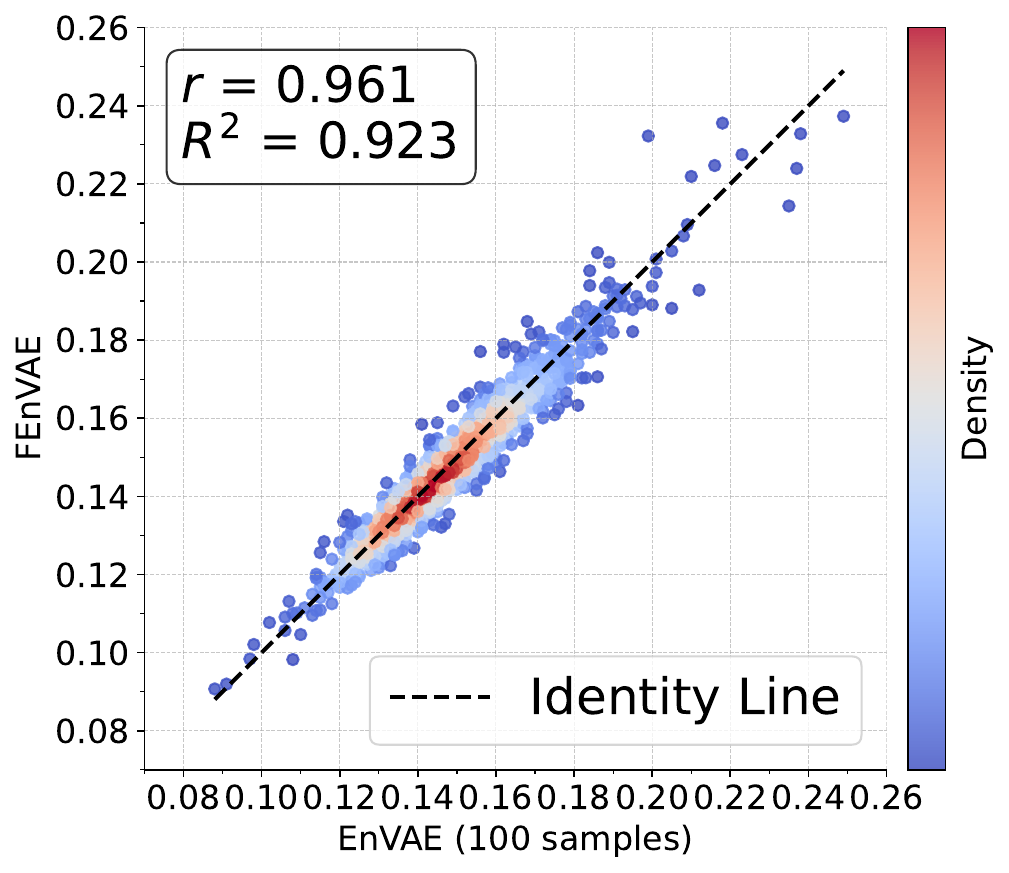}}
\vspace{-0.5cm}
\caption{Correlation of the uncertainty loss. Models are tested on 1000 samples.}\label{fig:7}
\vspace{-0.6cm}
\end{figure*}

In \textit{FEnVAE}, the validity of the local linearity approximation depends on two factors: 1) the magnitude of variable variances and 2) the smoothness of the generative function. To explore this, we compared the average variance of samples across different datasets and assessed the model's smoothness at various latent variable dimensions, with smoothness measured using the Lipschitz coefficient (details in Appendix \ref{A.H}).

\begin{wraptable}{r}{0.5\textwidth}
\centering
\vspace{-0.1cm}
\small
\caption{Performance comparison across datasets.}
\vspace{0.1cm}
\label{tab:31}
\renewcommand {\arraystretch}{1.2}
\resizebox{0.5\textwidth}{!}{
\begin{tabular}{c c c c c c  }
\hline
\hline
{Dataset} &Var.& Lip.& $\text{E\_FID}_{gen}\downarrow$ &$\text{F\_FID}_{gen}\downarrow$ &Impr.  \\
\hline
CelebA & $0.0925$ &$16.85$& $78.941$ &$80.012$& $-1.36\%$ \\
\hline
CIFAR & $0.3025$&$10.97$&$90.132$&$92.167$&$-2.26\%$  \\
\hline
LSUN & $0.2601$ &$14.27$&$143.064$ &$145.613$ &$-1.78\%$  \\
\hline
\end{tabular}}
\vspace{-0.3cm}
\end{wraptable}
Table \ref{tab:31} presents the performance comparison between \textit{FEnVAE} and \textit{EnVAE} on three datasets, with a latent space dimension of 64. We observed that the variance of the latent variables differs across datasets. On the CelebA dataset, the combined effect of the variance and the Lipschitz constant is relatively small, leading to \textit{FEnVAE} performing similarly to \textit{EnVAE}. However, on the LSUN and CIFAR datasets, these effects become more significant, resulting in a relative performance decrease. This indicates that \textit{FEnVAE} may show slight variations in adaptability across different datasets.

\begin{wraptable}{r}{0.5\textwidth}
\centering
\vspace{-0.5cm}
\small
\caption{Performance under different dimension.}
\vspace{0.1cm}
\label{tab:32}
\renewcommand {\arraystretch}{1.2}
\resizebox{0.5\textwidth}{!}{
\begin{tabular}{c c c c c c  }
\hline
\hline
{Dim.} &Var.& Lip.& $\text{E\_FID}_{gen}\downarrow$ &$\text{F\_FID}_{gen}\downarrow$ &Impr.  \\
\hline
8 & $0.0041$ &$26.03$& $102.383$ &$101.185$ & $+1.17\%$ \\
\hline
16 & $0.0069$ &$22.59$& $93.943$ &$93.351 $& $+0.63\%$  \\
\hline
32 & $0.0112$ &$19.18$& $85.945$  & $85.696$ & $+0.29\%$ \\
\hline
64 & $0.0925$ &$16.85$& $78.941$ &$80.012$& $-1.36\%$ \\
\hline
128 & $0.3671$ &$9.72$& $74.758$  & $76.134$ & $-1.84\%$  \\
\hline
\end{tabular}}
\end{wraptable}
Table~\ref{tab:32} reveals a strong correlation between latent dimensionality and the variance of encoded latent variables. When the latent space is low-dimensional, the model's representational capacity is limited, leading to a decline in overall performance. To encode diverse sample features within a constrained space, the model reduces the variance of each sample’s latent representation. This lower variance, in turn, enhances the effectiveness of the local linear approximation in \textit{FEnVAE}, allowing it to closely match and sometimes surpass the performance of \textit{EnVAE}.
Latent dimensionality also affects the smoothness of the generative function, as measured by the Lipschitz coefficient. Lower dimensions result in reduced smoothness, likely because the model compensates for limited capacity by learning more complex, non-linear mappings. However, such smoothness differences across datasets remain minor. We argue that the primary factor determining the validity of \textit{FEnVAE}’s local linearity approximation is the latent variance itself, which is shaped jointly by data complexity and model capacity. Additional data is shown in Appendix \ref{a:sa}.

In addition, we analyze the uncertainty losses term correlation between the full \textit{EnVAE} (using 100 samples) and the models trained with different sample sizes as well as \textit{FEnVAE}, as shown in Figure \ref{fig:7}. We observe that increasing the number of samples during training leads to improved model stability and stronger linear correlation  (higher Pearson correlation coefficient $r$ and coefficient of determination $R^2$) with the full \textit{EnVAE}. Notably, \textit{FEnVAE} achieves a correlation comparable to that of \textit{EnVAE} with many samples using only a single sample, which indirectly validates the effectiveness of our linear approximation.

\section{Conclusion}\label{sec:conclusion}

In this paper, we introduced \textit{EnVAE}, a likelihood-free variational autoencoder that replaces the conventional Gaussian reconstruction assumption with a sample-based energy score objective. To improve computational efficiency, we further proposed \textit{FEnVAE}, a single-sample variant derived under a local linearity approximation, which analytically eliminates the need for pairwise sample comparisons while preserving the expressiveness of the energy score.
Empirical evaluations demonstrate that both \textit{EnVAE} and \textit{FEnVAE} substantially outperform standard baselines. Notably, \textit{FEnVAE} achieves performance comparable to \textit{EnVAE} with significantly reduced computational cost. Visualization and ablation studies further validate the effectiveness of our approach in capturing complex data distributions without relying on explicit likelihoods. Overall, this approach holds the potential to address the limitations of standard VAEs that rely on simple likelihood functions for complex data and offers an alternative to the challenges associated with adversarial training in GANs.

\bibliographystyle{IEEEtran}
\bibliography{reference}

\newpage
\section{Appendix}\label{sec:appendix}

\subsection{Additional Related Work}

\subsubsection{Standard VAEs and Pixel-Wise Likelihoods}

Variational Autoencoders are latent-variable generative models that optimize a tractable evidence lower bound on the intractable marginal likelihood. In standard VAEs for images, the decoder defines an explicit probability distribution $p_{\theta}(\boldsymbol{x}|\boldsymbol{z})$, often a factorized pixel-level likelihood (e.g. independent Bernoulli or Gaussian for each pixel). While this approach achieves high log-likelihood, it can misallocate probability mass and yield blurry or unrealistic samples. In particular, maximizing per-pixel likelihood (equivalently minimizing pixelwise reconstruction error) tends to produce overly smooth images that average out ambiguity. Dosovitskiy and Brox (2016)\cite{dosovitskiy2016generating} famously observed that VAEs generate blurry outputs on complex image datasets. Although more expressive decoders can improve sample fidelity, they often cause the model to ignore the latent code (posterior collapse), since a powerful decoder can model the data distribution without relying on $\boldsymbol{z}$. This tension between pixel-wise likelihood objectives and latent-factor utilization has motivated numerous alternatives to the standard VAE formulation, for example, approaches based on prior design\cite{an2024distributional,takahashi2018student,dai2018diagnosing,rybkin2021simple,barron2019general}, architectural modifications\cite{sonderby2016ladder,vahdat2020nvae,pu2017adversarial,larsen2016autoencoding}, frequency-domain constraints\cite{jiang2021focal,bredell2023explicitly}, and ensemble-based enhancements\cite{luo2025energy,aneja2021contrastive,han2020joint}.

\subsubsection{Implicit and Adversarial VAE Objectives}

Adversarial Autoencoders (AAE)~\cite{makhzani2015adversarial} and Wasserstein Autoencoders (WAE)~\cite{tolstikhin2017wasserstein} replace the KL‑divergence regularizer in the standard variational objective with distribution‑matching criteria while retaining a generic reconstruction cost (e.g.\ an $\ell_2$ distance). In AAEs, a GAN‑style adversarial loss aligns the aggregated posterior $  q(\boldsymbol{z}) = \mathbb{E}_x[q(\boldsymbol{z}| \boldsymbol{x})]$ to a chosen prior by training a discriminator to distinguish encoded samples $\boldsymbol{z}\sim q(\boldsymbol{z})$ from prior samples $\boldsymbol{z}\sim p(\boldsymbol{z})$ and training the encoder to fool it. WAEs similarly keep a simple reconstruction term $c(\boldsymbol{x}, G(E(\boldsymbol{x})))$ but replace the VAE’s KL penalty with an Integral Probability Metric (IPM) between $q(\boldsymbol{z})$ and $p(\boldsymbol{z})$—such as the Wasserstein distance or Maximum Mean Discrepancy—enforcing latent‑space alignment via optimal‑transport or kernel‑moment matching and enabling deterministic encoder–decoder networks that generate high‑quality samples. Adversarially Learned Inference (ALI)~\cite{dumoulin2016adversarially} and BiGAN~\cite{donahue2016adversarial} extend this idea by training a joint encoder–decoder pair adversarially, where a discriminator must distinguish real encoder pairs $(\boldsymbol{x}, E(\boldsymbol{x}))$ from generated pairs $(G(\boldsymbol{z}), \boldsymbol{z})$, effectively aligning the implicit decoder’s output with the data distribution. These VAE/GAN hybrids learn both $G(\boldsymbol{z})$ and $q(\boldsymbol{z}| \boldsymbol{x})$ without any pixel‑level likelihood, achieving strong perceptual quality at the cost of training instability and the loss of a unified ELBO objective. Other works (e.g.\ Mescheder et al.~\cite{mescheder2017adversarial}) have similarly replaced the ELBO’s likelihood or KL term with learned discriminators or moment‑matching losses to avoid restrictive assumptions on $p(\boldsymbol{x}| \boldsymbol{z})$. In summary, adversarial and implicit‑objective VAEs demonstrate that explicit likelihoods are not strictly required for generative modeling, but they introduce new challenges—such as adversarial training and critic hyperparameter tuning—and lack a principled goodness‑of‑fit measure for reconstructions.

\subsubsection{Deterministic Decoders in Generative Models}

Traditional VAEs use stochastic decoders $\boldsymbol{z} \sim q_{\phi}(\boldsymbol{z}|\boldsymbol{x})$ then $\boldsymbol{x} \sim p_{\theta}(\boldsymbol{x}|\boldsymbol{z})$), but there is growing interest in deterministic decoders, where the generation is a deterministic function $\boldsymbol{x}=g_{\theta}(\boldsymbol{z})$ and all randomness comes from the encoder or prior. Deterministic decoders appear in adversarial autoencoders and GAN-based inference models (e.g. ALI/BiGAN), as discussed above, and have known benefits for representation learning. Polykovskiy and Vetrov (2020)\cite{polykovskiy2020deterministic} argue that a VAE with a deterministic decoder (dubbed DD-VAE) is less prone to latent variable collapse: since each latent code must map to a single output, the decoder cannot “over-express” the data distribution without utilizing the latent input. Indeed, with a highly flexible stochastic decoder (such as a PixelCNN autoregressive model\cite{van2016pixel,salimans2017pixelcnn++}), the VAE can achieve low reconstruction error by relying almost entirely on the decoder’s capacity, effectively ignoring $\boldsymbol{z}$. A deterministic decoder, by contrast, makes the latent code the sole source of variation in $\boldsymbol{x}$, which empirically was shown to improve the learned latent manifold and usefulness of $\boldsymbol{z}$ for downstream tasks. Recent work has explored training deterministic decoders while still performing variational encoding. Polykovskiy \& Vetrov\cite{polykovskiy2020deterministic} derived a modified ELBO for discrete sequence VAEs with an argmax decoder, including a continuous relaxation to enable gradient flow through the non-differentiable decoding operation. Their DD-VAE demonstrated that even without sampling in the decoder, one can train a generative autoencoder that produces diverse outputs (diversity coming from the encoder’s stochasticity and the prior) and maintains a structured latent space. Ghosh et al. (2020)\cite{ghosh2019variational} went further and completely removed stochastic elements from the autoencoder, proposing a Regularized Autoencoder (RAE) framework that is “simpler, easier to train, and deterministic, yet has many of the advantages of the VAE”. By interpreting the VAE’s decoder randomness as mere injected noise, they show that one can replace it with explicit regularization (e.g. a penalty on latent norms) to yield equally smooth and meaningful latent spaces. After training such a deterministic autoencoder with regularization, they fit a density (e.g. a Gaussian mixture) to the latent space as an “ex-post” prior, enabling sampling of new data points. The RAE approach achieved sample quality on par with or better than VAEs on image benchmarks, confirming that a deterministic decoder can suffice for generative modeling if the latent space is appropriately regularized.

Inference with a deterministic decoder (as in our method) requires redefining the training objective, since the log-likelihood $\log p_{\theta}(\boldsymbol{x}|\boldsymbol{z})$ is degenerate (Dirac delta). Prior solutions have included adding a small Gaussian noise to the decoder output to approximate a likelihood, or using heuristic reconstruction losses (e.g. $L_2$ or perceptual distance) in place of negative log-likelihood.
Our work departs from these by employing a proper scoring rule as the reconstruction loss, which provides a theoretically justified replacement for $\log p_{\theta}(\boldsymbol{x}|\boldsymbol{z})$. Importantly, the Energy Score we use remains well-defined even when $p_{\theta}(\boldsymbol{x}|\boldsymbol{z})$ is an implicit point mass – in fact, if the model predicts a single outcome $\boldsymbol{x}_{\text{pred}}$ for a given $\boldsymbol{z}$, the Energy Score reduces to $||\boldsymbol{x}_{\text{pred}} - \boldsymbol{x}_{\text{obs}}||$ (up to constant factor), behaving like a distance metric. Thus, optimizing the Energy Score drives the deterministic decoder to produce samples close to the ground truth in the metric space, while the score’s propriety ensures that, in expectation, the correct data distribution is favored. We are not aware of previous VAE variants that unite deterministic decoding with a proper scoring rule objective. Some previous implicit models either lacked an encoder (e.g., GANs) or used adversarial criteria that are not strictly proper measures of fit. By contrast, our method can be seen as a likelihood-free VAE trained end-to-end: the encoder $q_{\phi}(\boldsymbol{z}|\boldsymbol{x})$ is learned to approximate the Energy Score-optimal posterior, and the decoder $h_{\theta}(\boldsymbol{z})$ learns to produce samples that minimize the Energy Score against the true data. This novel combination – deterministic decoder, scoring-rule-based loss, and variational inference – positions our approach at the intersection of the above threads. It leverages the strengths of each: the simplicity and latent fidelity of deterministic decoders, the statistical grounding of proper scoring rules, and the efficiency of amortized variational inference. Crucially, it avoids both explicit likelihood modeling and adversarial min-max training, charting a new path for generative modeling that is likelihood-free, yet principled and practical.

\subsection{Preliminaries of Variational Autoencoders}\label{A.B}

Variational Autoencoders are a class of deep generative models that learn to approximate complex data distributions through an encoder-decoder framework. The model is based on the idea of variational inference and aims to approximate the posterior distribution of latent variables by introducing a tractable variational distribution.

Let $ \boldsymbol{x} \in \mathbb{R}^n $ represent an observed data, and let $ \boldsymbol{z} \in \mathbb{R}^m $ denote the latent variables. The generative process of VAEs defines a joint distribution
\begin{align}
	p_\theta(\boldsymbol{x}, \boldsymbol{z}) = p_\theta(\boldsymbol{x}|\boldsymbol{z})p_\theta(\boldsymbol{z}),
\end{align}
where $ p_\theta(\boldsymbol{x}|\boldsymbol{z}) $ is the likelihood function, and $ p_\theta(\boldsymbol{z}) $ is the prior over latent variables. The objective is to maximize the marginal likelihood
\begin{align}
 p_\theta(\boldsymbol{x}) = \int p_\theta(\boldsymbol{x}|\boldsymbol{z})p_\theta(\boldsymbol{z})d\boldsymbol{z},
\end{align}
which is intractable due to the integral over the latent variables.

To address this, VAEs introduce a variational approximation $ q_\phi(\boldsymbol{z}|\boldsymbol{x}) $ to the true posterior $ p_\theta(\boldsymbol{z}|\boldsymbol{x}) $, thereby enabling tractable inference. The goal is to maximize the Evidence Lower Bound (ELBO) on the marginal log-likelihood, which is equivalent to minimizing the negative ELBO as the loss function:
\begin{align}
\mathcal{L}(\theta, \phi; \boldsymbol{x}) =-\text{ELBO}= -\mathbb{E}_{q_\phi(\boldsymbol{z}|\boldsymbol{x})}[\log p_\theta(\boldsymbol{x}|\boldsymbol{z})] + \mathcal{D}_\text{KL}(q_\phi(\boldsymbol{z}|\boldsymbol{x}) \| p_\theta(\boldsymbol{z})),
\end{align}
where $ \mathbb{E}_{q_\phi(\boldsymbol{z}|\boldsymbol{x})}[\log p_\theta(\boldsymbol{x}|\boldsymbol{z})] $ is the expected log-likelihood (reconstruction term), and $ \mathcal{D}_\text{KL}(q_\phi(\boldsymbol{z}|\boldsymbol{x}) \| p_\theta(\boldsymbol{z})) $ is the Kullback-Leibler (KL) divergence term that regularizes the variational distribution $ q_\phi(\boldsymbol{z}|\boldsymbol{x}) $ towards the prior distribution $ p_\theta(\boldsymbol{z}) $. The ELBO can be seen as a trade-off between reconstructing the data well and regularizing the posterior distribution.

During training, VAE learns the parameters $ \theta $ and $ \phi $ by maximizing the ELBO. In practice, the variational distribution $ q_\phi(\boldsymbol{z}|\boldsymbol{x}) $ is typically chosen to be a simple parametric distribution, such as a Gaussian, and is optimized using stochastic gradient descent.
Once trained, samples from the model can be generated by first sampling from the prior $ p_\theta(\boldsymbol{z}) $ and then decoding the latent variable $ \boldsymbol{z} $ using the decoder $ p_\theta(\boldsymbol{x}|\boldsymbol{z}) $. This allows the model to generate new samples from the same distribution as the training data.
Variational Autoencoders have become a powerful and widely used framework for generative modeling, with applications in various domains, including image generation, anomaly detection, and semi-supervised learning.

\subsection{Case Analysis of $\beta=2$}\label{A.E}

We analyze the case of the energy score with $\beta=2$, deriving insights from the energy score approximation using $M$ Monte Carlo samples as formulated in Eq.~\eqref{eq:fullloss}:
\begin{align}
	\hat{\mathcal{L}}(\theta,\phi; \boldsymbol{x}) = &\frac{1}{{M}}\sum\limits_{i = 1}^M {||{\boldsymbol{x}^*_i} - \boldsymbol{x}||^2}  - \frac{1}{{2{M}(M-1)}}\sum\limits_{i = 1}^M {\sum_{j:j\neq i} {||{\boldsymbol{x}^*_i} - {\boldsymbol{x}^*_j}||^2} } \nonumber\\
	&+\frac{1}{2} \sum_{i=1}^{M} \left( 1 + \log(\boldsymbol{\sigma}_i^2) - \boldsymbol{\mu}_i^2 - \boldsymbol{\sigma}_i^2 \right)\nonumber\\
    =&\sum_k^n \left( \frac{1}{{M}}\sum\limits_{i = 1}^M { || \boldsymbol{x}_{i,k}^{*} - \boldsymbol{x}_k^{*} ||^2} - \frac{1}{{2{M(M-1)}}}\sum\limits_{i = 1}^M {\sum_{j:j\neq i} { || \boldsymbol{x}_{i,k}^{*} - \boldsymbol{x}_{j,k}^{*} ||^2 }}\right) \nonumber\\
    &+\frac{1}{2} \sum_{i=1}^{M} \left( 1 + \log(\boldsymbol{\sigma}_i^2) - \boldsymbol{\mu}_i^2 - \boldsymbol{\sigma}_i^2 \right)\nonumber\\
    =&\sum_k^n \left( \frac{1}{{M}}\sum\limits_{i = 1}^M { || \boldsymbol{x}_{i,k}^{*} - \boldsymbol{x}_{\mu,k}^{*} ||^2}+\frac{1}{{M}}\sum\limits_{i = 1}^M { || \boldsymbol{x}_{\mu,k}^{*} - \boldsymbol{x}_k ||^2} -\frac{1}{{2{M(M-1)}}}\sum\limits_{i = 1}^M {\sum_{j:j\neq i} { || \boldsymbol{x}_{i,k}^{*} - \boldsymbol{x}_{j,k}^{*} ||^2 }}\right) \nonumber\\
    &+\frac{1}{2} \sum_{i=1}^{M} \left( 1 + \log(\boldsymbol{\sigma}_i^2) - \boldsymbol{\mu}_i^2 - \boldsymbol{\sigma}_i^2 \right)\nonumber\\
    =&\sum_k^n \left( \frac{1}{{M}}\sum\limits_{i = 1}^M { || \boldsymbol{x}_{\mu,k} - \boldsymbol{x}_k ||^2}+\text{Var}(x_{i,k}) - \frac{1}{{2}}*2\text{Var}(\boldsymbol{x}_{i,k})\right) \nonumber\\
    &+\frac{1}{2} \sum_{i=1}^{M} \left( 1 + \log(\boldsymbol{\sigma}_i^2) - \boldsymbol{\mu}_i^2 - \boldsymbol{\sigma}_i^2 \right)\nonumber\\
    =&\frac{1}{{M}}\sum\limits_{i = 1}^M { || \boldsymbol{x}_\mu - \boldsymbol{x} ||^2}
    +\frac{1}{2} \sum_{i=1}^{M} \left( 1 + \log(\boldsymbol{\sigma}_i^2) - \boldsymbol{\mu}_i^2 - \boldsymbol{\sigma}_i^2 \right),
\end{align}
where $\boldsymbol{x}_\mu$ is the mean value of reconstructed data. Therefore, when $\beta=2$, the reconstruction loss degenerates into MSE, the loss function of the Vanilla VAE.

\subsection{Datasets and Baselines}\label{A.G}

\subsubsection{Datasets}
\begin{itemize}
\item{ \textbf{CelebA:}}
The CelebA dataset contains over 200,000 celebrity images spanning 10,000 identities. Each image is labeled with 40 attribute labels, such as "smiling" or "wearing glasses." This dataset is widely used for facial recognition and attribute prediction tasks. CelebA provides a large-scale, diverse set of images with various poses, lighting conditions, and facial expressions, making it suitable for evaluating image generation models, especially those dealing with faces and facial features.

\item{ \textbf{CIFAR-10:}}
The CIFAR-10 dataset consists of 60,000 32x32 color images in 10 classes, with 6,000 images per class. The classes include airplane, automobile, bird, cat, deer, dog, frog, horse, ship, and truck. CIFAR-10 is a widely used benchmark for evaluating image classification and generation models. Despite its relatively small image size, CIFAR-10 presents a challenge for generative models due to the complexity and diversity of the object classes.

\item{\textbf{LSUN:}}
The Large-scale Scene Understanding (LSUN) dataset is a large-scale dataset for scene understanding tasks, containing millions of labeled images across a variety of scene categories. Each scene category contains tens of thousands of images, with examples such as bedroom, living room, and dining room. LSUN is particularly useful for evaluating image generation models in the context of complex, high-resolution scenes, making it a suitable benchmark for evaluating models that aim to generate realistic scenes or environments. In this paper, we use data from the church category for evaluation.
\end{itemize}

\subsubsection{Baselines}

Our experiments include two types of evaluations: one compares our model with other methods that assume predefined distributions, following \cite{bredell2023explicitly}; while the other integrates our model into state-of-the-art frameworks for comparison. Therefore, the experimental section consists of two kinds of baselines.

First, we select several commonly used loss terms as reconstruction losses, including:

\begin{itemize}
    \item \textbf{L2 loss:} Measures the squared difference between the reconstructed and ground truth data, which is the same as the reconstruction loss used in Vanilla VAE.
    \begin{equation}
        \mathcal{L}_{\text{L2}} = \| \boldsymbol{x} - \boldsymbol{\hat{x}} \|_2^2.
    \end{equation}
    where $\boldsymbol{x}$ is the ground truth, $\boldsymbol{\hat x}$ is the output data. L2 loss assumes that the reconstruction error follows a Gaussian distribution, making it sensitive to outliers.

    \item \textbf{L1 loss:} Computes the absolute difference between the reconstructed and ground truth data.
    \begin{equation}
        \mathcal{L}_{\text{L1}} = \| \boldsymbol{x} - \boldsymbol{\hat{x}} \|_1.
    \end{equation}
    L1 loss corresponds to the negative log-likelihood of a Laplace-distributed reconstruction error. Compared to L2 loss, it is more robust to outliers due to the heavier tails of the Laplace distribution.

    \item \textbf{Cross-entropy (CE) loss:} Used for matching the probability distributions between the reconstructed and ground truth data. In the case of normalized continuous data, the cross-entropy loss can be interpreted as a form of Kullback-Leibler divergence, where the model aims to minimize the difference between the predicted continuous distribution and the target distribution. For continuous data, the cross-entropy loss is given by:
    \begin{equation}
        \mathcal{L}_{\text{CE}} = - \int p(\boldsymbol{x}) \log q(\boldsymbol{x})   dx,
    \end{equation}
    where $ p(\boldsymbol{x}) $ represents the true distribution of the data and $ q(\boldsymbol{x}) $ is the predicted distribution. This formulation can be viewed as minimizing the divergence between the predicted and true distributions, making it suitable for continuous, normalized data.

    \item \textbf{Structural Similarity Index (SSIM) loss:} Evaluates image quality by comparing luminance, contrast, and structural information.
    \begin{equation}
        \text{SSIM}(\boldsymbol{x}, \boldsymbol{\hat{x}}) = \frac{(2\boldsymbol{\mu}_x\boldsymbol{\mu}_{\hat{x}} + C_1)(2\boldsymbol{\sigma}_{x\hat{x}} + C_2)}
        {(\boldsymbol{\mu}_x^2 + \boldsymbol{\mu}_{\hat{x}}^2 + C_1)(\boldsymbol{\sigma}_x^2 + \boldsymbol{\sigma}_{\hat{x}}^2 + C_2)},
    \end{equation}
    where $ \boldsymbol{\mu}_x $ and $ \boldsymbol{\mu}_{\hat{x}} $ are the means, $ \boldsymbol{\sigma}_x^2 $ and $ \boldsymbol{\sigma}_{\hat{x}}^2 $ are the variances, and $ \boldsymbol{\sigma}_{x\hat{x}} $ is the covariance between $ \boldsymbol{x} $ and $ \boldsymbol{\hat{x}} $. The corresponding SSIM loss is given by:
    \begin{equation}
        \mathcal{L}_{\text{SSIM}} = 1 - \text{SSIM}(\boldsymbol{x}, \boldsymbol{\hat{x}}).
    \end{equation}

\end{itemize}

In addition to the aforementioned evaluation methods, we also introduce four complete models as baselines for comparison: InfoVAE, FFL, Student-t, and DistVAE.

\begin{itemize}

    \item \textbf{InfoVAE\cite{zhao2017infovae}:} InfoVAE extends the traditional variational autoencoder by introducing a flexible divergence term that balances the trade-off between data fidelity and latent space regularization. Instead of solely relying on KL divergence, InfoVAE incorporates alternatives such as Maximum Mean Discrepancy (MMD), enabling more expressive latent representations. This formulation allows the model to retain more mutual information between observed data and latent variables, thereby improving both generative quality and representation learning.

    \item \textbf{Student-t VAE\cite{takahashi2018student}:} The Student-t VAE enhances the traditional VAE by incorporating a Bayesian approach to variance estimation. In the original VAE, the Gaussian assumption can lead to instability, especially when the estimated variance is close to zero. The Student-t VAE mitigates this issue by setting a prior for the variance of the Gaussian decoder and marginalizing it out analytically.

    \item \textbf{DistVAE (Distributional VAE)\cite{an2024distributional}:} DistVAE introduces a distributional learning approach for VAEs that directly estimates the conditional cumulative distribution function (CDF), bypassing the assumption of a Gaussian distribution. This method is nonparametric and leverages the estimation of an infinite number of conditional quantiles. The reconstruction loss is equivalent to the continuous ranked probability score (CRPS) loss.
\end{itemize}

In the integration experiments, we adopt three EBM-based VAEs as the base model, including DT-VAE, NCP-VAE, and EC-VAE:
\begin{itemize}
    \item \textbf{DT-VAE (Divergence Triangle VAE)\cite{han2020joint}:} DT-VAE jointly trains the VAE and the latent Energy-Based Model, allowing these two models to leverage each other's strengths. The objective function includes Kullback-Leibler divergences between the data density, the generator density, and the latent EBM density, seamlessly integrating variational and adversarial learning. The latent EBM acts as a critic for the generator model, improving the generator’s ability to approximate the data distribution.

    \item \textbf{NCP-VAE (Noise Contrastive Prior VAE)\cite{aneja2021contrastive}:} NCP-VAE introduces a noise contrastive prior (NCP), which is learned by contrasting samples from the aggregate posterior with samples from a base prior. This simple post-training mechanism enhances the expressivity of the VAE’s prior.

    \item \textbf{EC-VAE (Energy-Calibrated VAE)\cite{luo2025energy}:} EC-VAE incorporates a conditional Energy-Based Model for calibration of the VAE, enhancing generative performance while maintaining high sampling efficiency. The model generates samples, which are then refined using the conditional EBM to approximate the real data. By minimizing the distance between generated and refined samples during training, the decoder is calibrated.
\end{itemize}
In the integration experiments, we only need to replace the reconstruction loss of the VAE component in the baseline models from MSE to our EnL or FEnL, where corresponding sampling or transformation of the latent space is required.

\subsection{Evaluation Metrics}\label{A.H}

\begin{itemize}
\item{Fréchet Inception Distance (FID)}
FID measures the distributional difference between generated and real images. Specifically, it computes the Fréchet distance between the mean and covariance of features extracted by the Inception network from real and generated images. A lower FID value indicates that the generated images are closer to the real images in terms of their feature distribution, implying higher quality of the generated images. The FID is computed as:

\begin{equation}
    \text{FID}(p, q) = \left\| \boldsymbol{\mu}_p - \boldsymbol{\mu}_q \right\|_2^2 + \text{Tr}(\boldsymbol{\Sigma}_p + \boldsymbol{\Sigma}_q - 2(\boldsymbol{\Sigma}_p \boldsymbol{\Sigma}_q)^{1/2}),
\end{equation}
where $ \boldsymbol{\mu}_p, \boldsymbol{\Sigma}_p $ are the mean and covariance of the real image features, and $ \boldsymbol{\mu}_q, \boldsymbol{\Sigma}_q $ are the mean and covariance of the generated image features.

\item{Inception Score (IS)}
Inception Score is a widely used metric for evaluating the quality and diversity of generated images. It relies on the classification probabilities predicted by an Inception network trained on a large dataset (e.g., ImageNet). A higher IS value indicates that the generated images are more diverse and semantically meaningful. The IS is computed as:
\begin{equation}
    \text{IS} = \exp \left( \mathbb{E}_{\boldsymbol{x} \sim p_g} \left[ D_{\text{KL}} \left( p(\boldsymbol{y}|\boldsymbol{x}) \| p(\boldsymbol{y}) \right) \right] \right),
\end{equation}
where $ p(\boldsymbol{y}|\boldsymbol{x}) $ is the conditional label distribution predicted by the Inception network for a generated image $ \boldsymbol{x} $, and $ p(\boldsymbol{y}) $ is the marginal distribution of predicted labels. $ D_{\text{KL}} $ represents the Kullback-Leibler divergence, which measures how much the conditional distribution deviates from the marginal distribution.

\item{Lipschitz Constant for VAE Encoder Smoothness}
The Lipschitz constant measures the smoothness of the VAE encoder by quantifying the maximum rate of change in its output with respect to input variations. A lower Lipschitz constant implies a smoother encoder, indicating that the local linearity approximation holds more effectively.

Formally, the Lipschitz constant $ Lip $ of an encoder function $ f(x) $ is defined as:
\begin{equation}
    Lip = \sup_{\boldsymbol{x}_1 \neq \boldsymbol{x}_2} \frac{\| f(\boldsymbol{x}_1) - f(\boldsymbol{x}_2) \|}{\| \boldsymbol{x}_1 - \boldsymbol{x}_2 \|},
\end{equation}
where $ \boldsymbol{x}_1, \boldsymbol{x}_2 $ are input data points, and $ f(\boldsymbol{x}) $ represents the encoder output in the latent space. A large $ Lip $ indicates sensitivity to small input perturbations, leading to less smooth latent representations, while a small $ Lip $ ensures local consistency and robustness.
\end{itemize}

\subsection{Additional Reconstruction Images}\label{Ari}

We provide additional reconstruction samples from the three datasets.

\begin{figure}[H]
\centering
\includegraphics[width=5.5in]{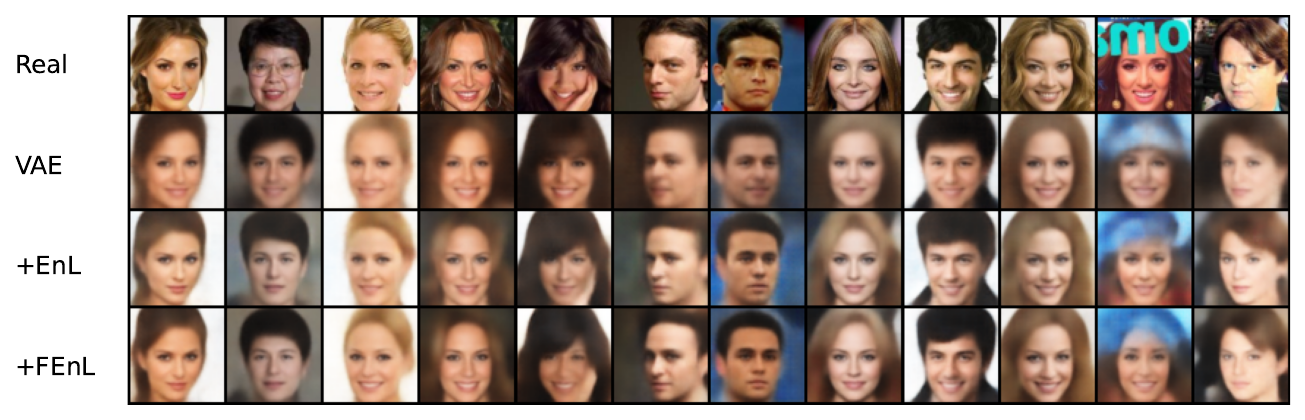}
\vspace{-0.3cm}
\caption{Additional reconstruction images of CelebA 64 dataset.}
\label{1}
\end{figure}

\begin{figure}[H]
\centering
\includegraphics[width=5.5in]{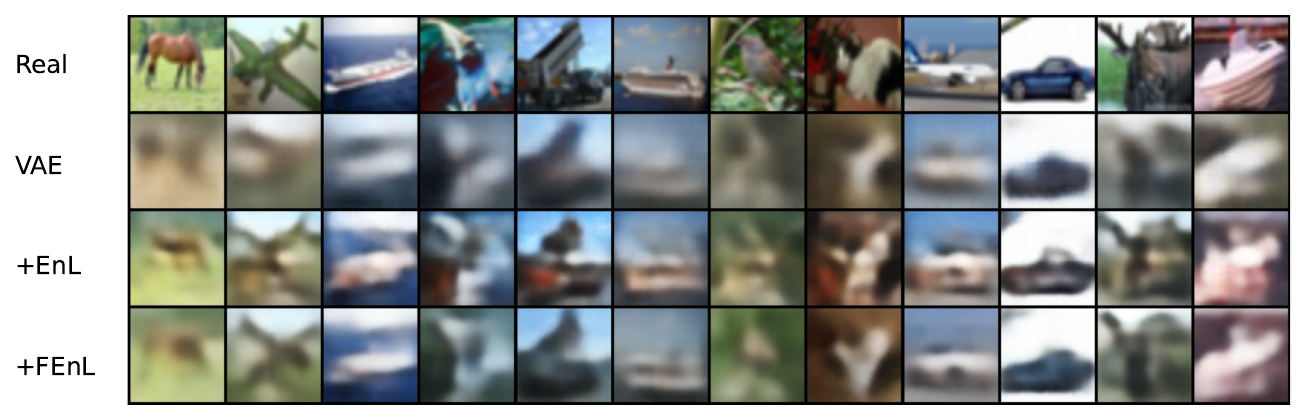}
\vspace{-0.3cm}
\caption{Additional reconstruction images of CIFAR-10 dataset.}
\label{1}
\end{figure}

\begin{figure}[H]
\centering
\includegraphics[width=5.5in]{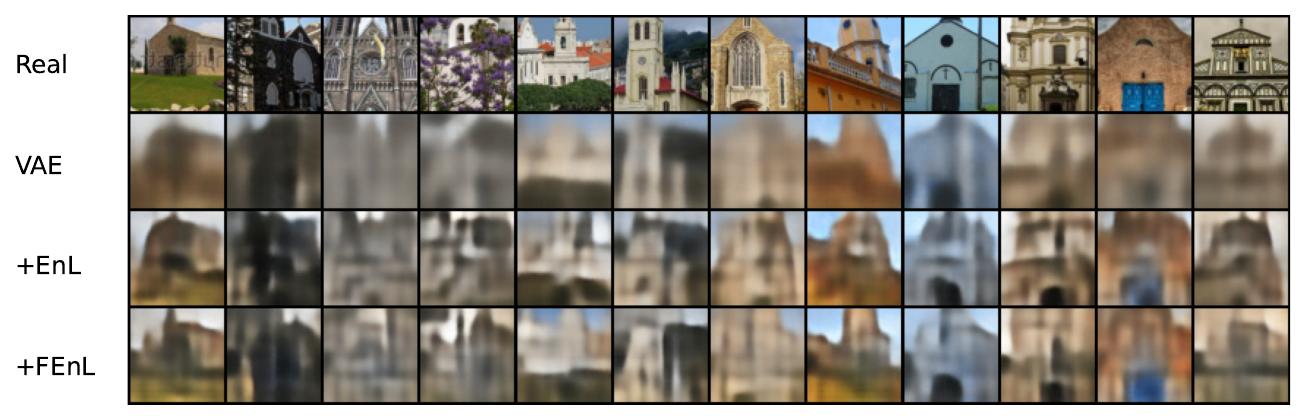}
\vspace{-0.3cm}
\caption{Additional reconstruction images of LSUN 64 dataset.}
\label{1}
\end{figure}

\subsection{Additional generation images}\label{Agi}

We provide additional generation samples from the three datasets.

\begin{figure}[H]
\centering
\includegraphics[width=5.5in]{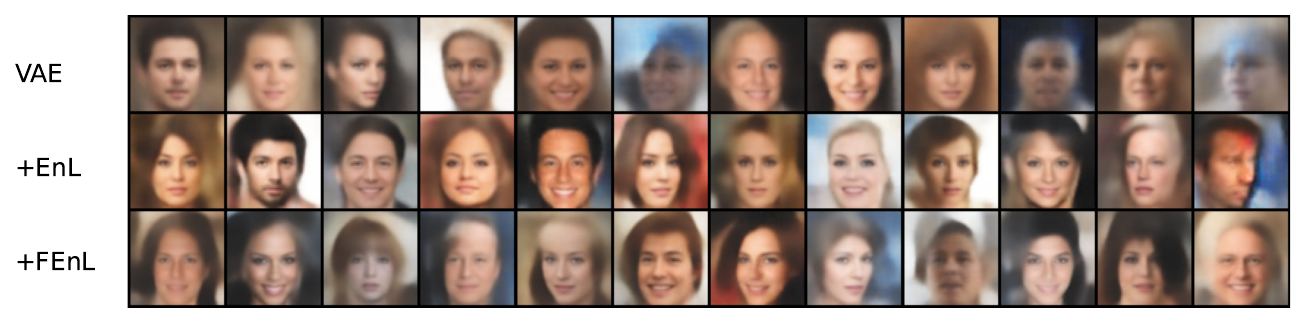}
\vspace{-0.3cm}
\caption{Additional generation images of CelebA dataset.}
\label{1}
\end{figure}

\begin{figure}[H]
\centering
\includegraphics[width=5.5in]{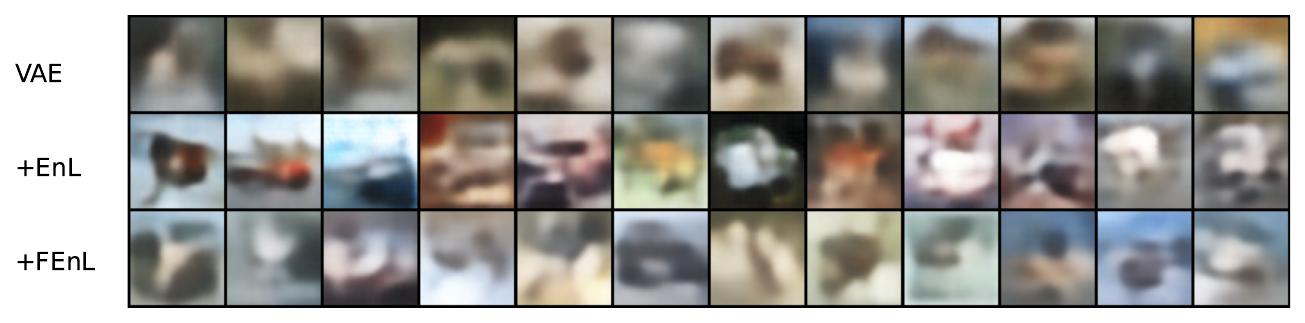}
\vspace{-0.3cm}
\caption{Additional generation images of CIFAR-10 dataset.}
\label{1}
\end{figure}

\begin{figure}[H]
\centering
\includegraphics[width=5.5in]{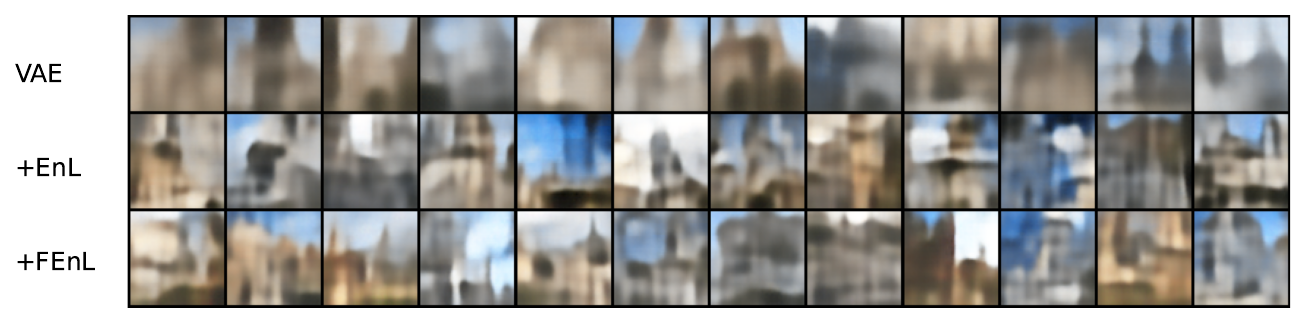}
\vspace{-0.3cm}
\caption{Additional generation images of LSUN 64 dataset.}
\label{1}
\end{figure}

In addition, we provide generation samples obtained by integrating our model into three EBM models.

\begin{figure}[H]
\centering
\vspace{-0.1cm}
\subfigure[ Based on DT-VAE.]{\includegraphics[width=1.81in]{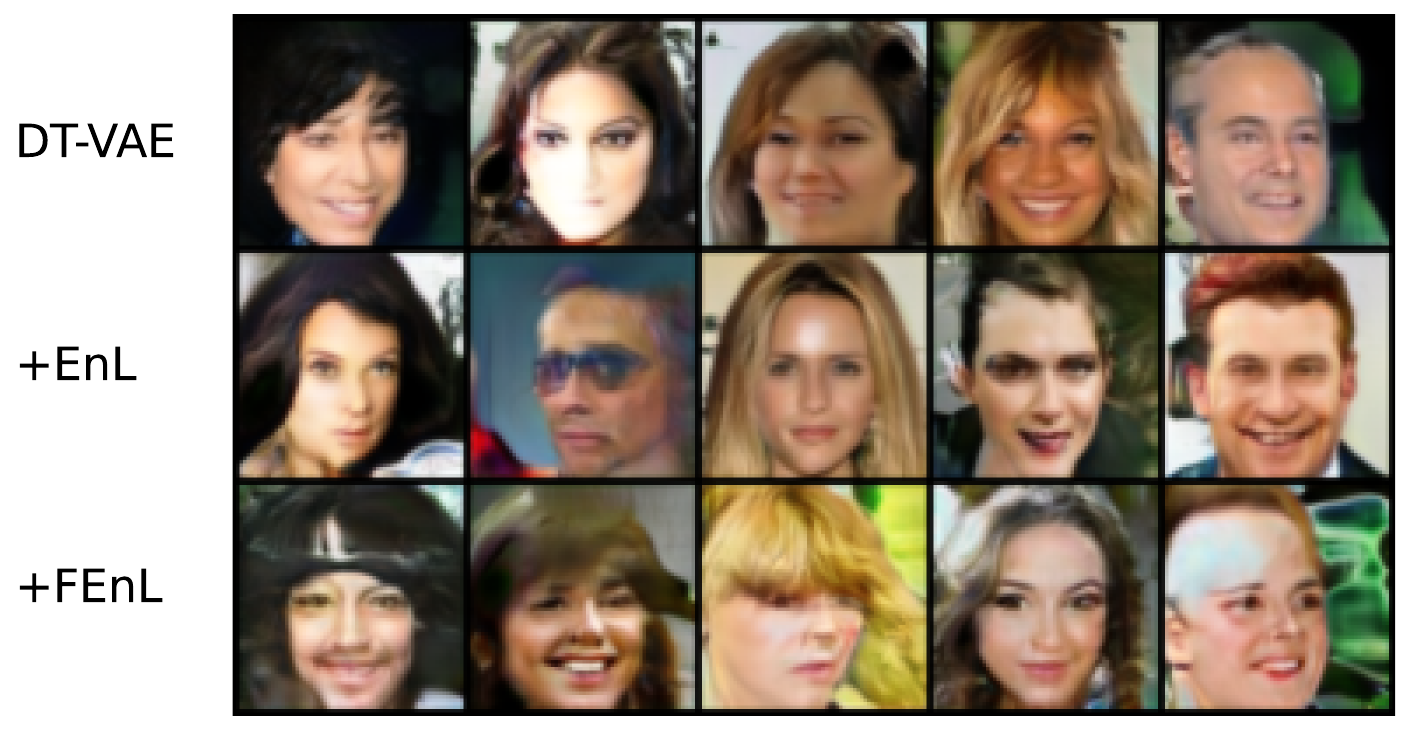}}
\subfigure[Based on NCP-VAE. ]{\includegraphics[width=1.81in]{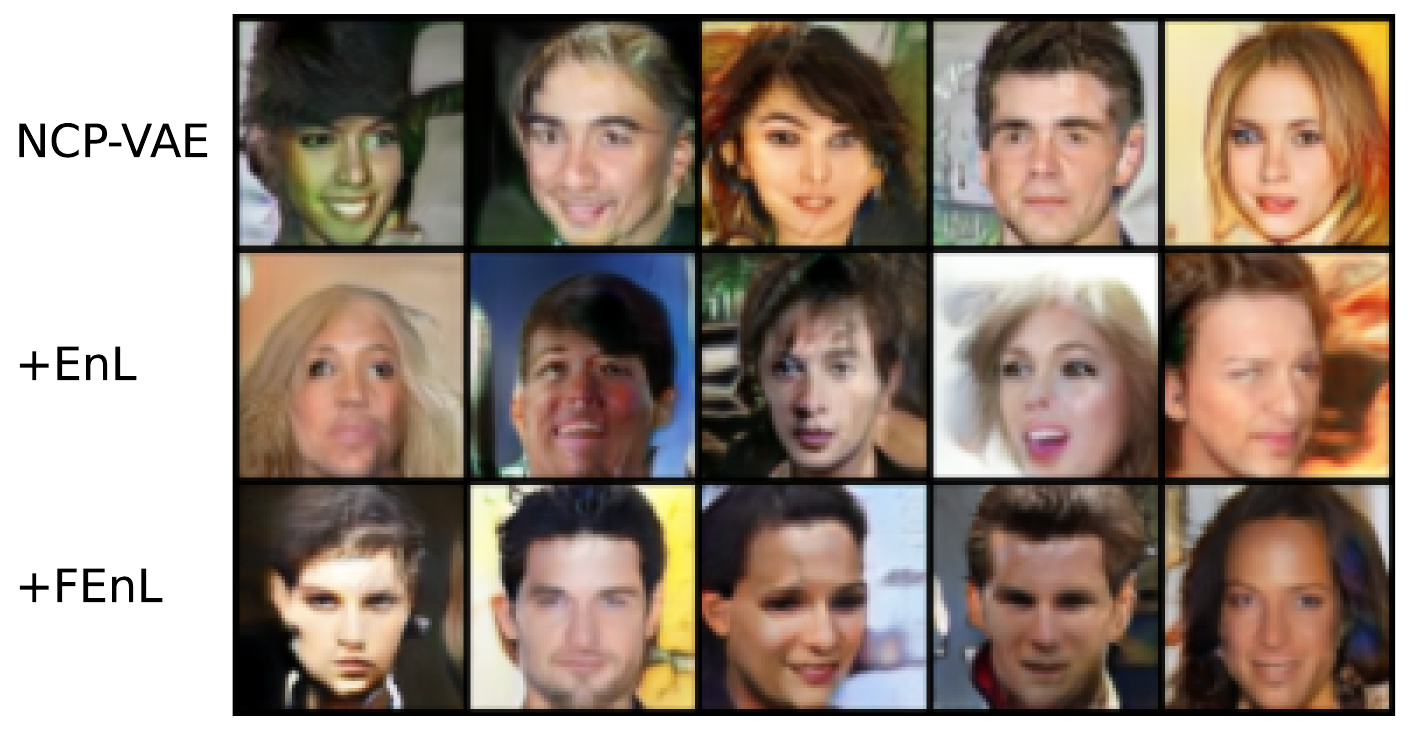}}
\subfigure[ Based on EC-VAE.]{\includegraphics[width=1.81in]{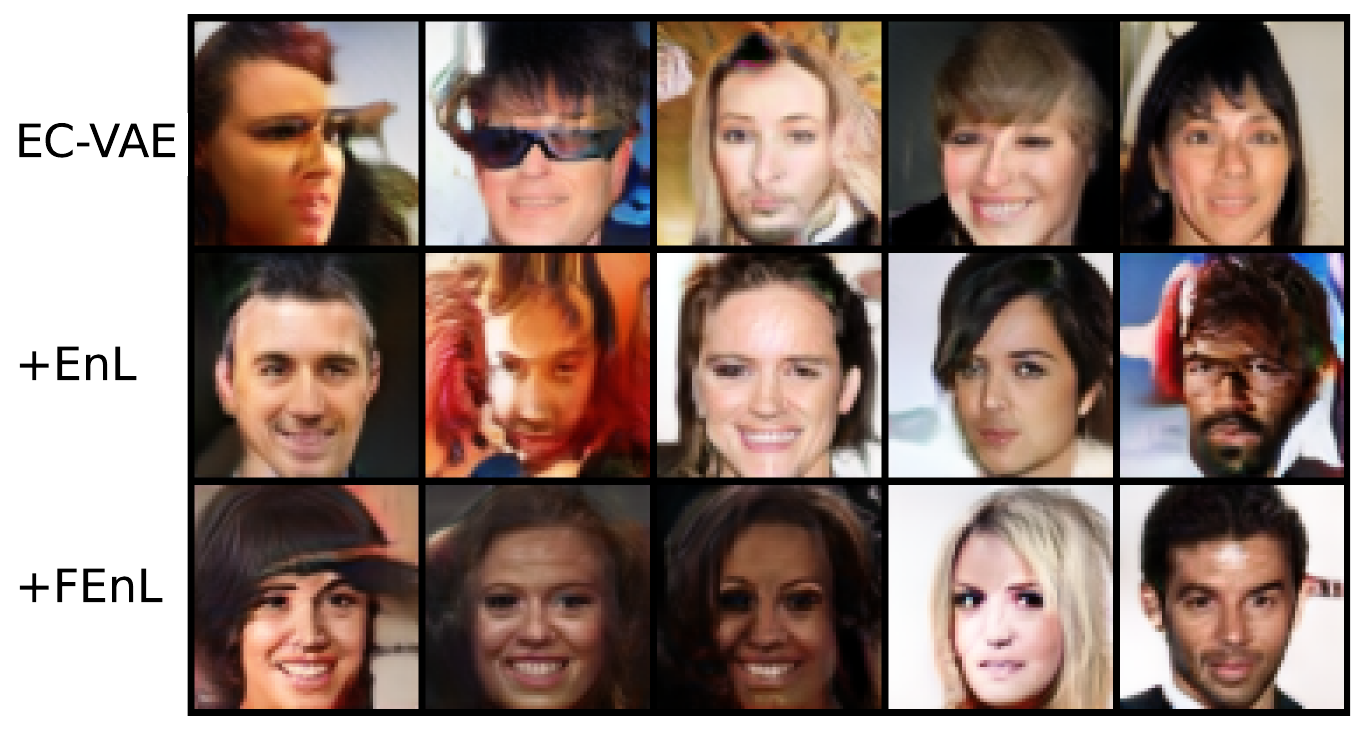}}
\vspace{-0.4cm}
\caption{Additional generation images of CelebA 64 dataset after integration.}
\end{figure}

\begin{figure}[H]
\centering
\vspace{-0.1cm}
\subfigure[ Based on DT-VAE.]{\includegraphics[width=1.81in]{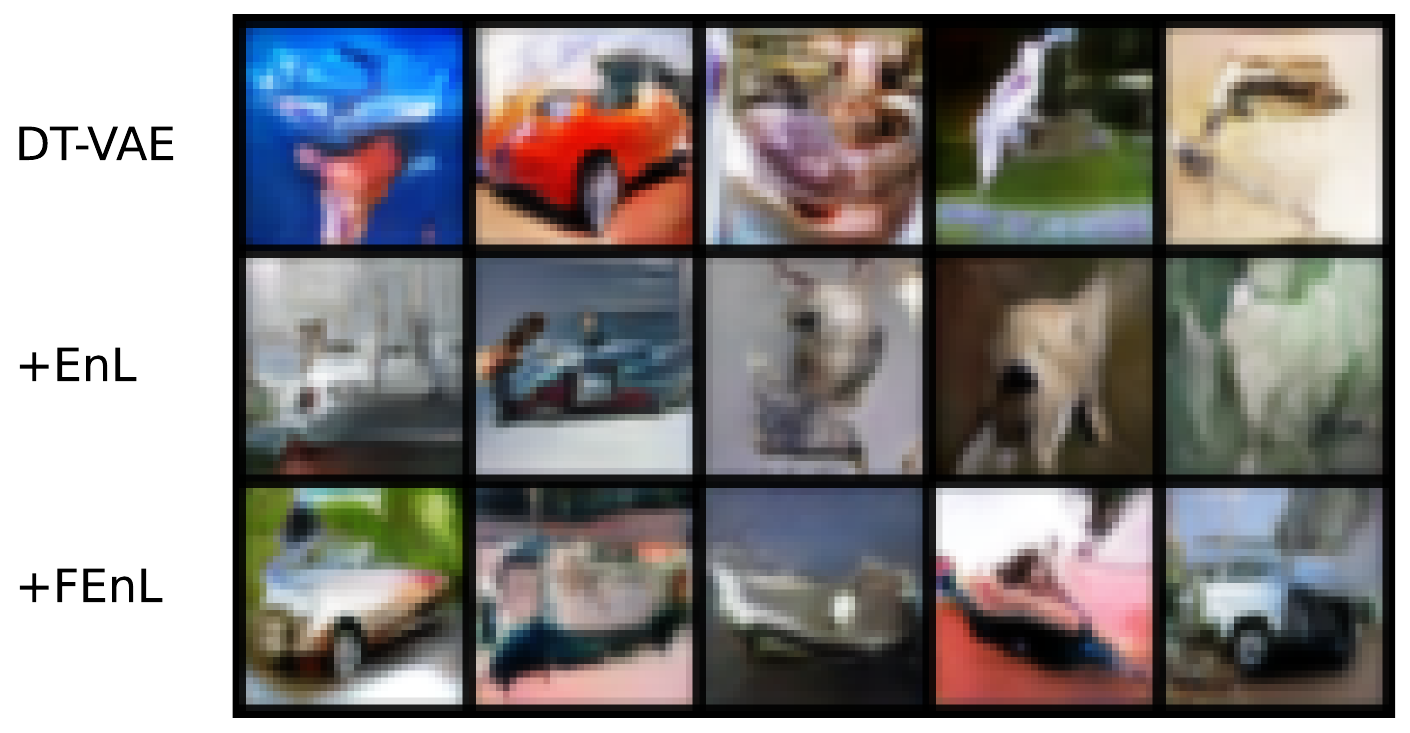}}
\subfigure[ Based on NCP-VAE.]{\includegraphics[width=1.81in]{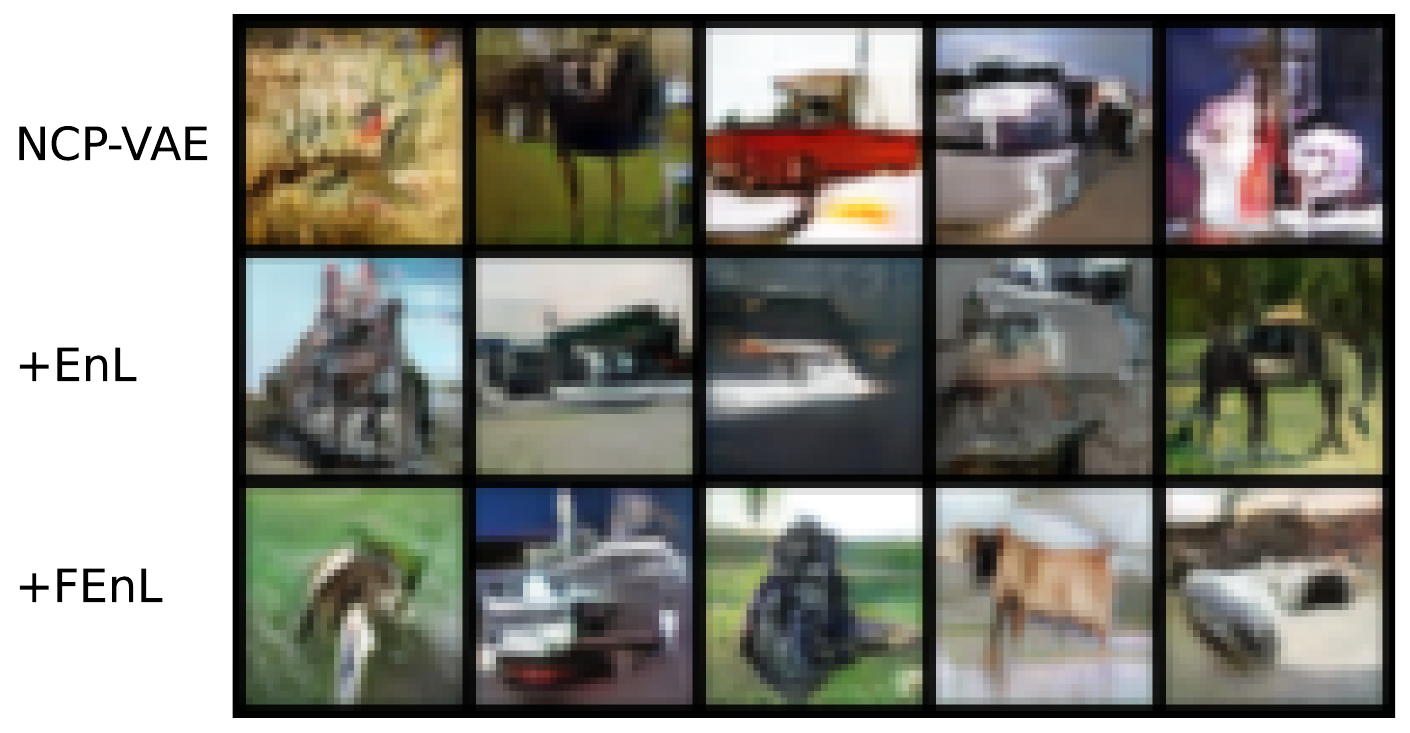}}
\subfigure[Based on EC-VAE. ]{\includegraphics[width=1.81in]{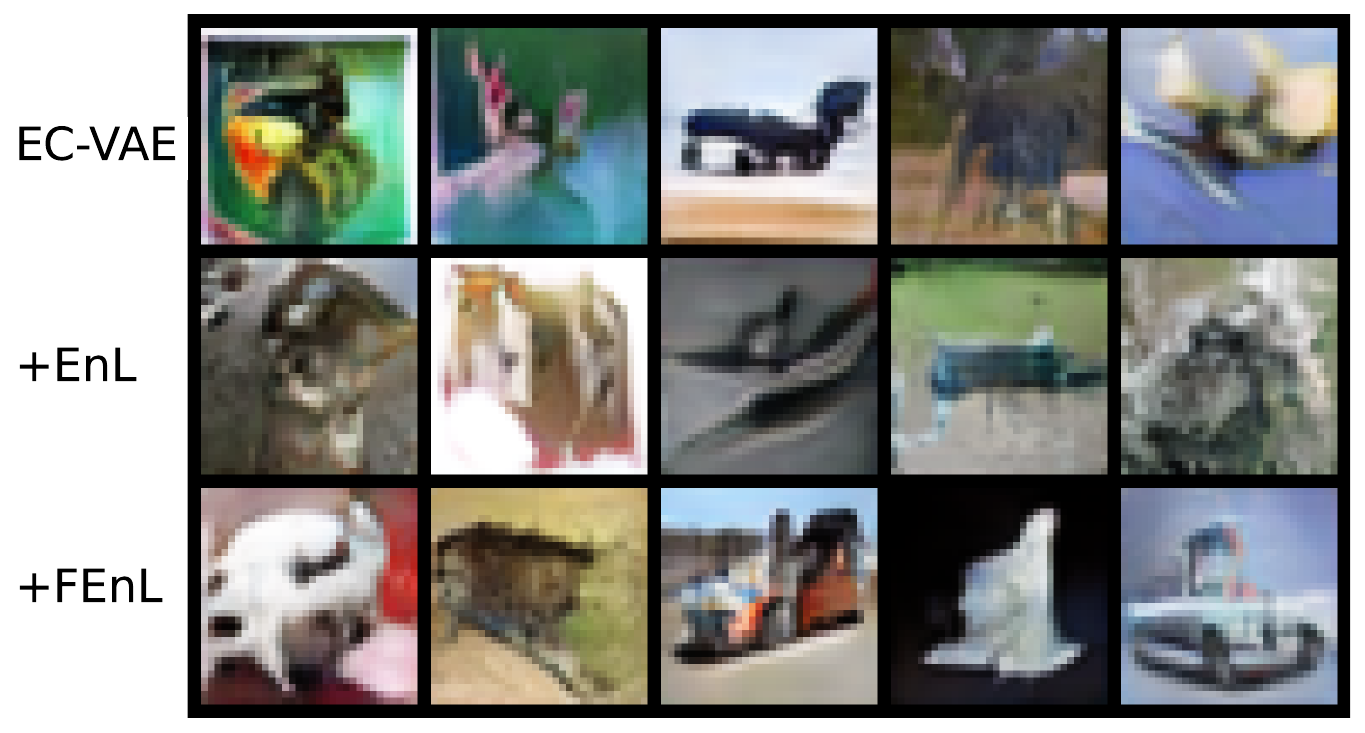}}
\vspace{-0.4cm}
\caption{Additional generation images of CIFAR-10 dataset after integration.}
\end{figure}

\begin{figure}[H]
\centering
\vspace{-0.1cm}
\subfigure[Based on DT-VAE. ]{\includegraphics[width=1.81in]{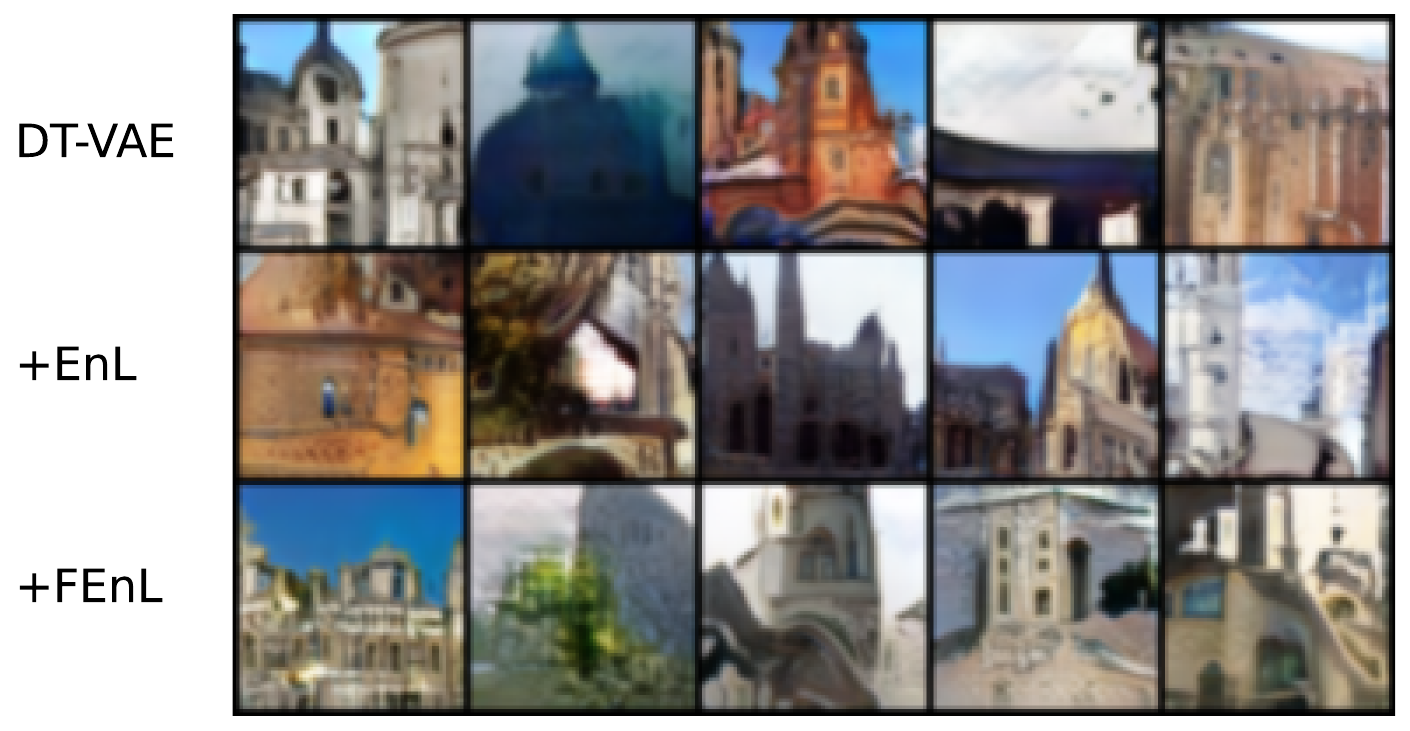}}
\subfigure[  Based on NCP-VAE.]{\includegraphics[width=1.81in]{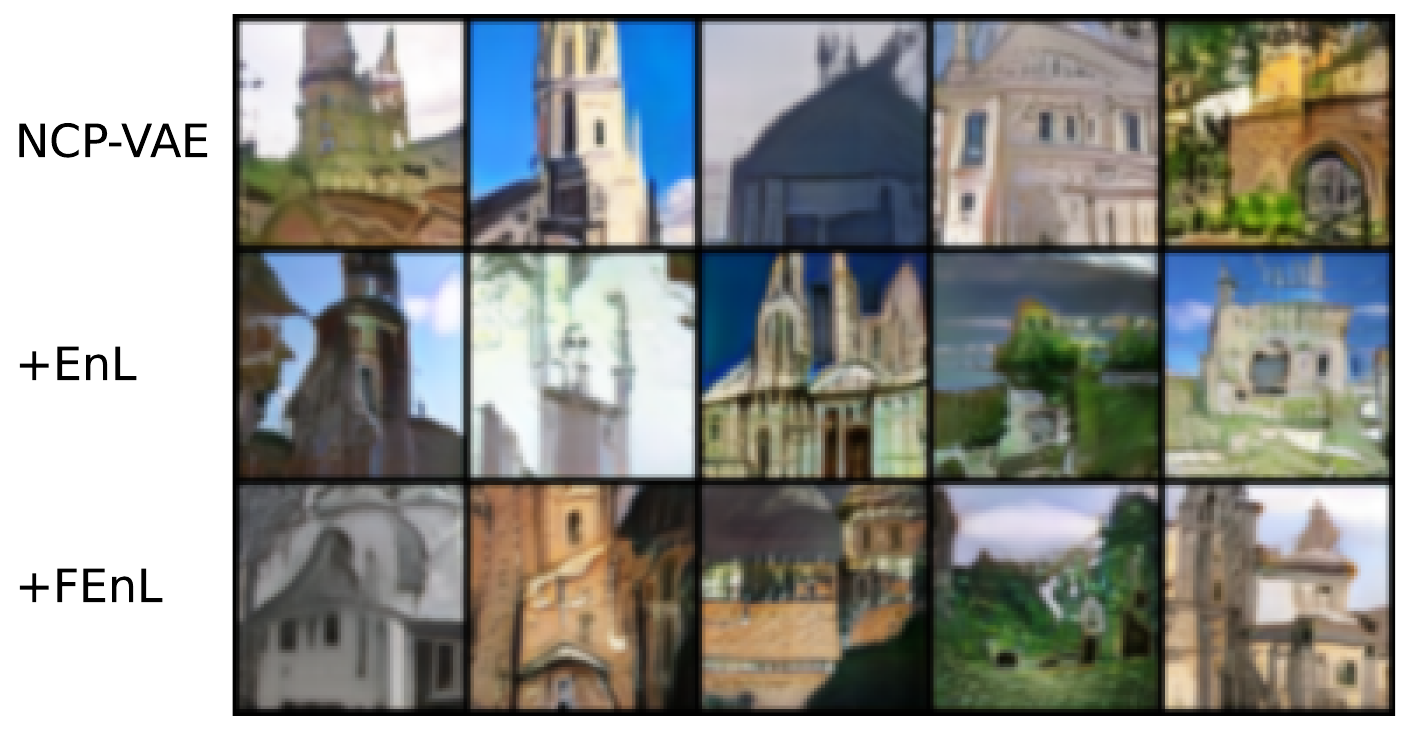}}
\subfigure[Based on EC-VAE. ]{\includegraphics[width=1.81in]{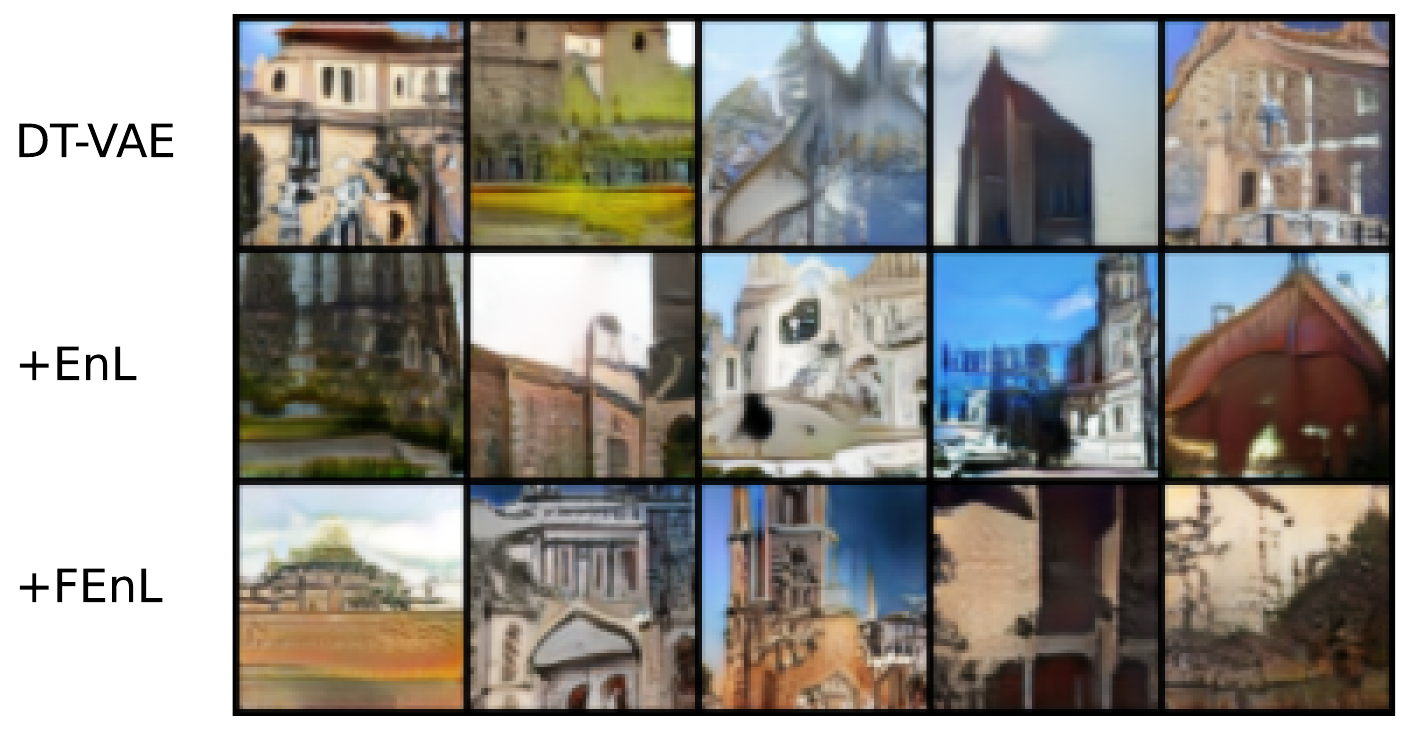}}
\vspace{-0.4cm}
\caption{Additional generation images of LSUN 64 dataset after integration.}
\end{figure}

\subsection{Spectral Visualization}

\begin{figure}[H]
\centering
\vspace{-0.3cm}
\includegraphics[width=3.5in]{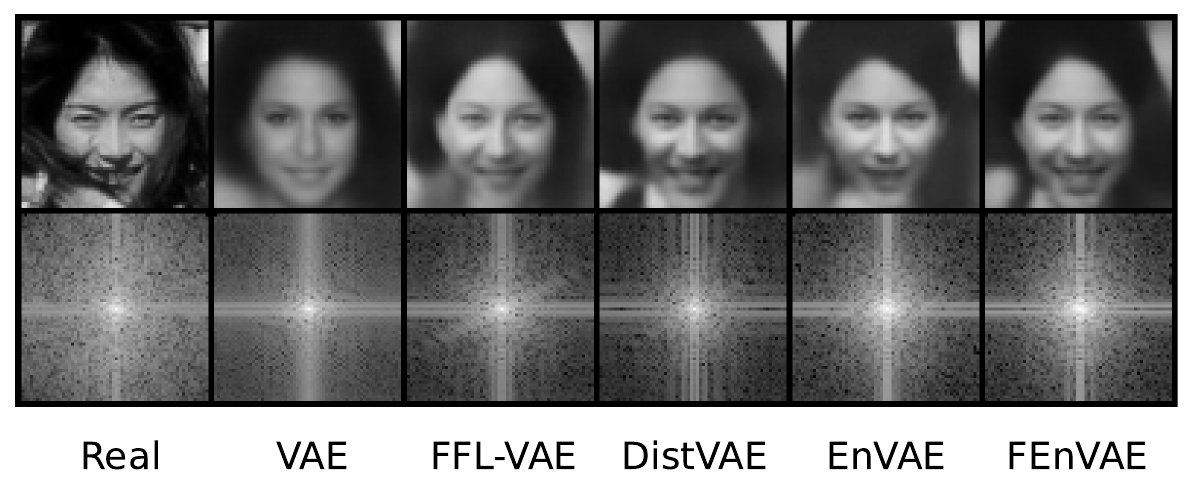}
\vspace{-0.3cm}
\caption{Comparison of spectral characteristics.}
\vspace{-0.2cm}
\label{fig:4}
\end{figure}
We visualize the spectral plots of the real and reconstructed images from Vanilla VAE, FFL-VAE, DistVAE, and our proposed models, as shown in Figure \ref{fig:4}. The spectra of our models exhibit increased intensity in both high-frequency components (peripheral regions) and low-frequency components (central regions) compared to baselines, with a particularly pronounced enhancement in the high-frequency region. The transition between high and low frequencies is relatively smooth. Overall, the spectral representation of our models aligns more closely with that of the real image, suggesting that our model effectively captures the full-spectrum characteristics of the image. Thus, the reconstructed and generated images from our models demonstrate enhanced detail.

\begin{figure}[H]
\centering
\vspace{-0.1cm}
\subfigure[ ]{\includegraphics[width=2.5in]{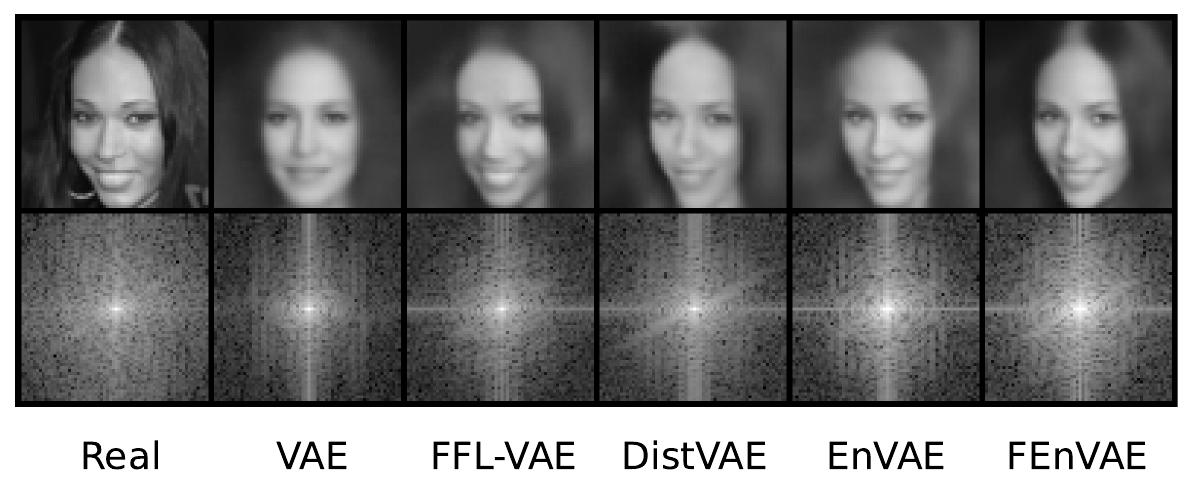}}
\subfigure[ ]{\includegraphics[width=2.5in]{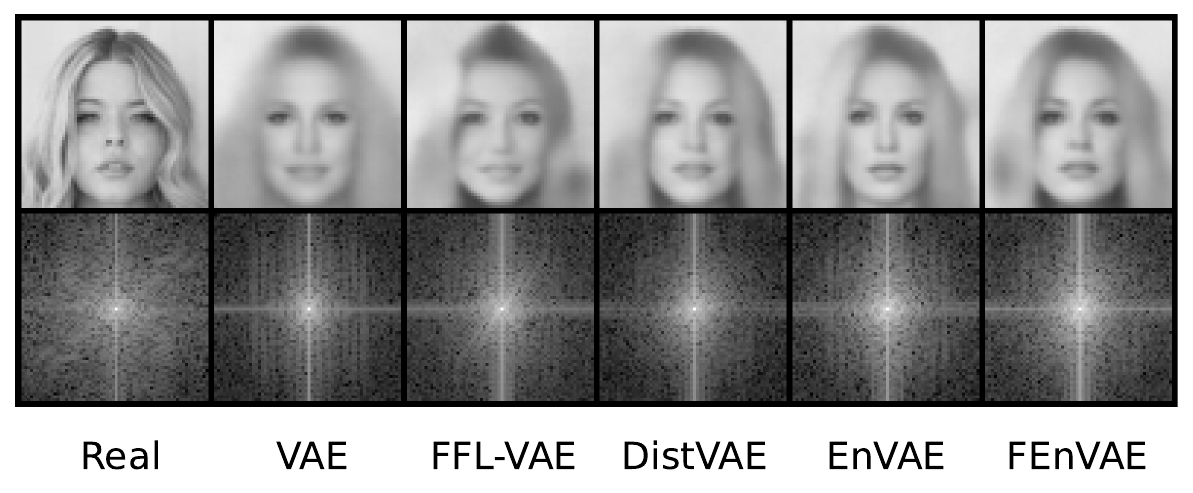}}
\vspace{-0.2cm}
\caption{Additional spectral visualization of CelebA 64 dataset.}
\end{figure}

\begin{figure}[H]
\centering
\vspace{-0.1cm}
\subfigure[ ]{\includegraphics[width=2.5in]{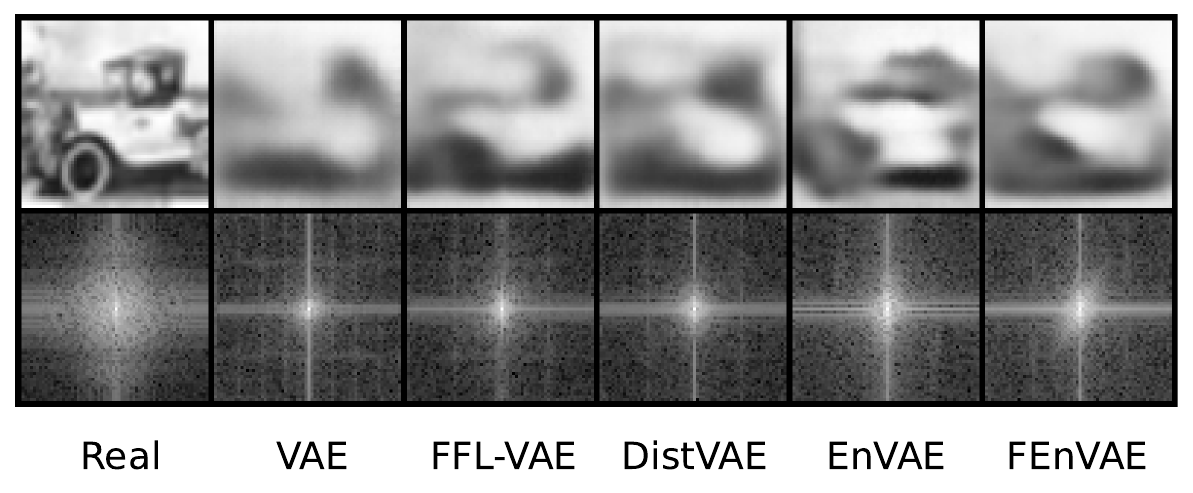}}
\subfigure[ ]{\includegraphics[width=2.5in]{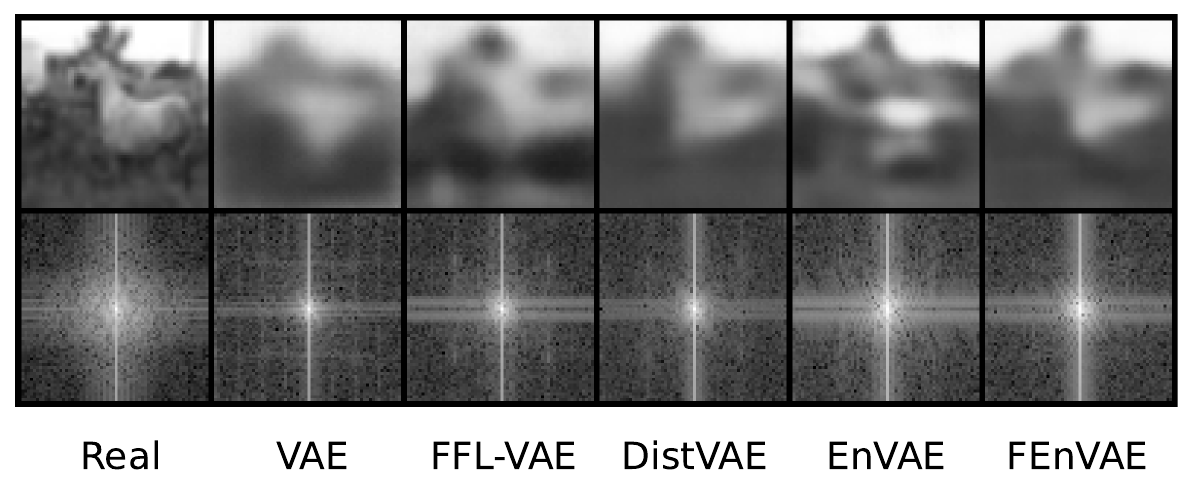}}
\vspace{-0.2cm}
\caption{Additional spectral visualization of CIFAR-10 dataset.}
\end{figure}

\begin{figure}[H]
\centering
\vspace{-0.1cm}
\subfigure[ ]{\includegraphics[width=2.5in]{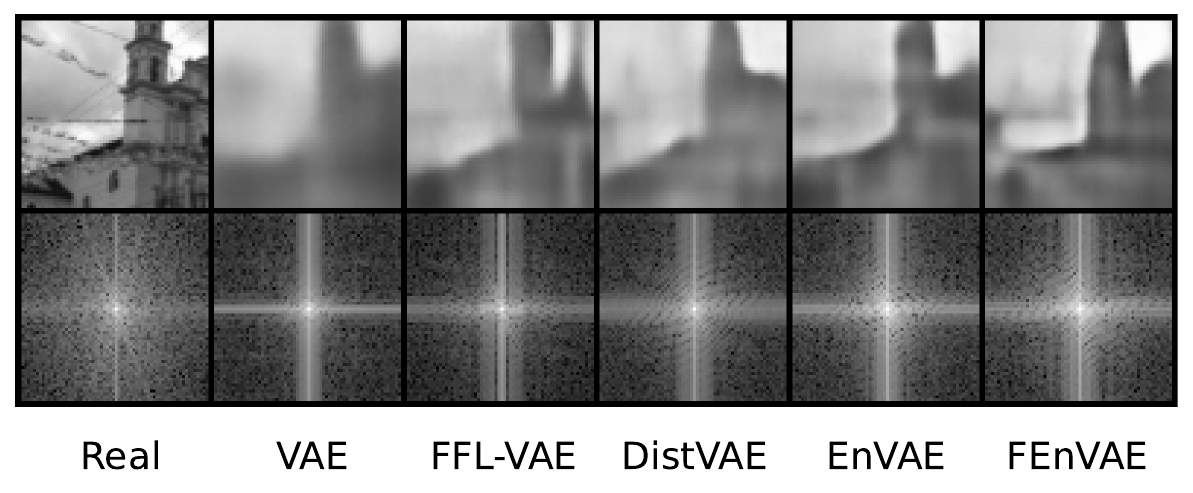}}
\subfigure[ ]{\includegraphics[width=2.5in]{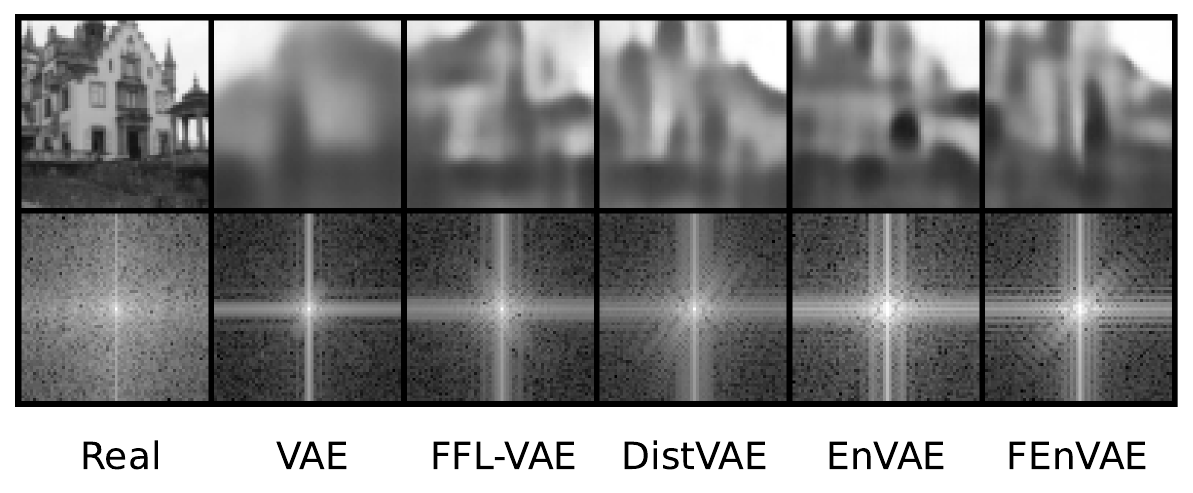}}
\vspace{-0.2cm}
\caption{Additional spectral visualization of LSUN 64 dataset.}
\end{figure}

\subsection{Performance of the single-sampling scheme compared to the original \textit{FEnVAE}}

\begin{table}[H]
\centering
\caption{Image generation performance.}
\label{tab:4}
\renewcommand {\arraystretch}{1.2}
\begin{tabular}{l c c c c c c c c c}
\hline
\hline
\multirow{2}{*}{Model} & \multicolumn{2}{c}{CelebA 64} & & \multicolumn{2}{c}{CIFAR-10} & & \multicolumn{2}{c}{LSUN Church 64} \\
\cline{2-3} \cline{5-6} \cline{8-9}
& FID$\downarrow$ & IS$\uparrow$ & & FID$\downarrow$ & IS$\uparrow$ & & FID$\downarrow$ & IS$\uparrow$ \\
\hline
FEnVAE\_single & $80.012$ & $2.150$ & & $92.167$ & $2.544$ & & $145.613$ & $2.617$
\\
FEnVAE\_double & $80.194$ & $2.211$ & & $92.320$ & $2.537$ & & $145.552$ & $2.602$ \\
\hline
\end{tabular}
\end{table}

In the \textit{FEnVAE} model, we adopt a single-sampling strategy to replace the original double sampling. We have verified the impact of this simplified sampling approach on performance through experiments, as shown in Table \ref{tab:4}. The results show that under the single-sampling strategy, the model's performance is comparable to that of the double-sampling approach. Therefore, it can be concluded that the correlation between the mean term and the uncertainty term in the loss function (since the samples come from the same distribution) does not significantly impact the model's optimization and performance.

\subsection{Raw performance data of \textit{EnVAE} and \textit{FEnVAE} under different sample sizes}

We provide the raw data related to Figure \ref{fig:6}.

\begin{table}[H]
\caption{Image generation performance.}
\label{tab:3}
\renewcommand {\arraystretch}{1.2}
\resizebox{\textwidth}{!}{
\begin{tabular}{l c c c c c c c c c c c c c c c }
\hline
\hline
\multirow{2}{*}{Dataset} & \multirow{2}{*}{Metrics}&\multirow{2}{*}{VAE} &\multirow{2}{*}{FEnVAE} & \multicolumn{10}{c}{EnVAE (\# of samples)} \\
\cline{5-14}
&  &  & &  10& 20 &  30 & 40&50&60&70&80&90&100 \\
\hline

\multirow{3}{*}{CelebA} & FID & $98.691$ &$80.012$ &$85.298$ &$ 84.543$ &$ 82.261$ &$ 81.628$ &$ 78.941$ &$ 78.269$ &$ 77.669$ &$ 78.169$ &$ 77.409$ &$ 77.499$ \\
 & IS &$1.826$ & $2.150$ & $1.990$ &$ 2.022$ &$ 2.112$ &$ 2.131$ &$ 2.177$ &$ 2.187$ &$ 2.185$ &$ 2.190$ &$ 2.197$ &$ 2.199$\\
  & Time (s) & $115$ &$125$& $132$ & $156$ & $203$&$265$&$332$&$414$&$504$&$614$&$724$&$834$ \\
\hline
\multirow{3}{*}{CIFAR} & FID & $107.630$ & $92.167$ & $97.338$ &$ 95.043$ &$ 93.061$ &$ 92.328$ &$ 90.132$ &$ 90.069$ &$ 89.669$ &$ 89.369$ &$89.469$ &$ 89.671$ \\
 & IS & $2.440$ & $2.544$& $2.509$ &$ 2.521$ &$ 2.537$ &$ 2.541$ &$ 2.553$ &$ 2.554$ &$ 2.559$ &$ 2.553$ &$ 2.552$ &$ 2.554$ \\
  & Time (s) & $40$ &$42$& $46$&$ 60$&$ 75$&$ 86$&$ 102$&$ 125$&$ 140$&$ 154$&$ 169$&$ 185$ \\
\hline
\multirow{3}{*}{LSUN} & FID & $180.375$ &$145.613$  & $154.338$ &$ 152.043$ &$ 150.061$ &$ 144.328$ &$ 143.064$ &$ 141.369$ &$ 141.669$ &$ 141.609$ &$ 142.069$ &$ 141.599$ \\
 & IS & $2.361$ &$2.617$ &$2.493$ &$ 2.52$ &$ 2.570$ &$ 2.585$ &$ 2.630$ &$ 2.636$ &$ 2.645$ &$ 2.641$ &$ 2.652$ &$ 2.643$ \\
  & Time (s) & $78$ &$83$& $90$&$ 112$&$ 150$&$ 176$&$ 233$&$ 284$&$ 340$&$ 390$&$ 439$&$ 475$ \\
\hline
\end{tabular}}
\end{table}

\subsection{Result and Visualization of Sample Variance }\label{a:sa}

We visualize the distribution of a randomly selected single sample in the latent space $ p_{\boldsymbol{\theta}}(\boldsymbol{z}_i | \boldsymbol{x}_i) $ (colored distribution), the overall posterior distribution $ p_{\boldsymbol{\theta}}(\boldsymbol{z} | \boldsymbol{x}) $ (background gray distribution), and the prior distribution $ p_{\boldsymbol{\theta}}(\boldsymbol{z}) $ (background blue distribution).

From the figure, it can be observed that the latent space variance of a single sample is significantly smaller than that of the entire space. When the latent space dimension is 64, the variance of the latent variables for individual samples is relatively large. However, when the latent space dimension is reduced to 32, the variance of the latent variables becomes very small.

\begin{figure}[H]
\centering
\subfigure[Latent space dimension of 64 in the CelebA 64 dataset. ]{\includegraphics[width=1.8in]{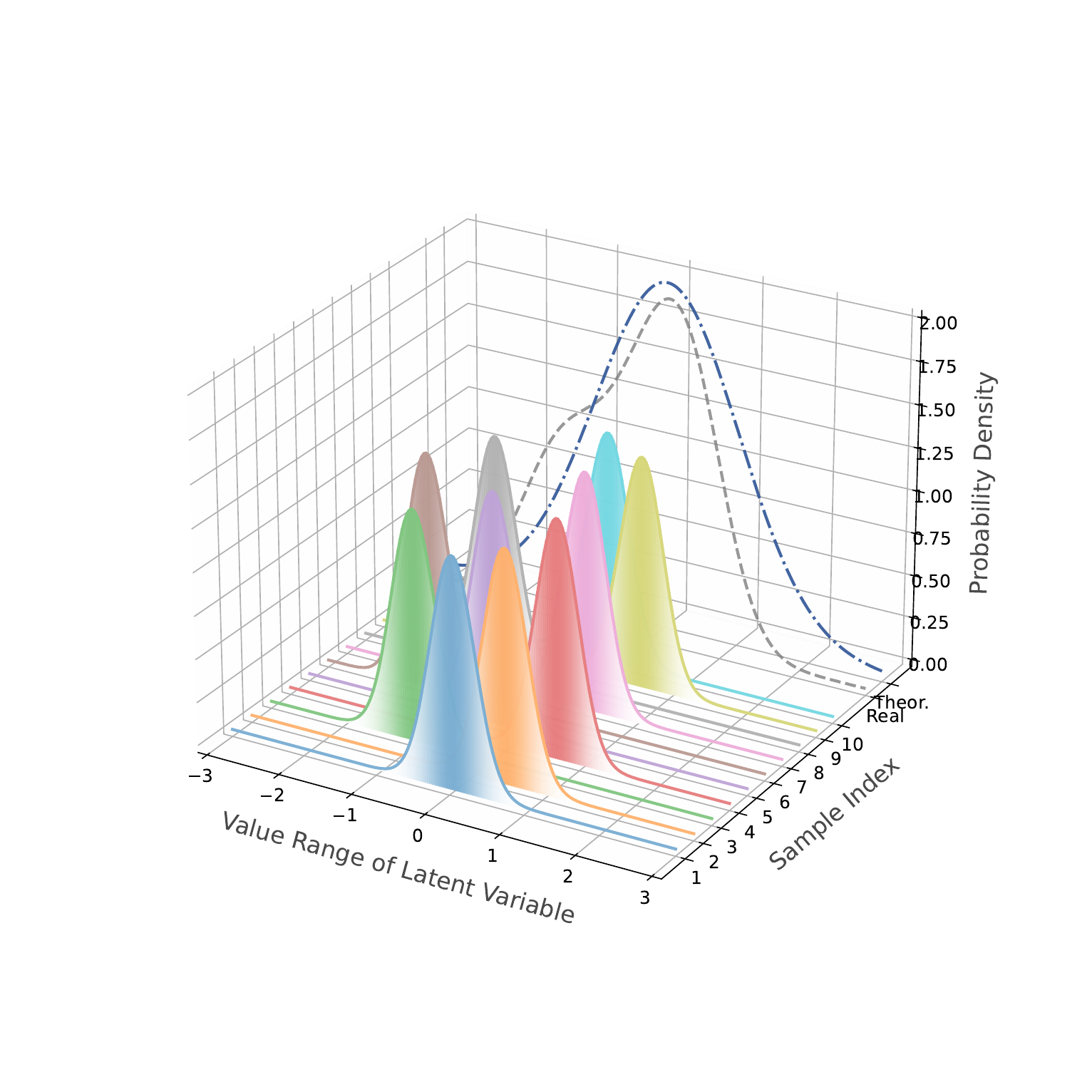}}
\subfigure[Latent space dimension of 64 in the CIFAR-10 dataset. ]{\includegraphics[width=1.8in]{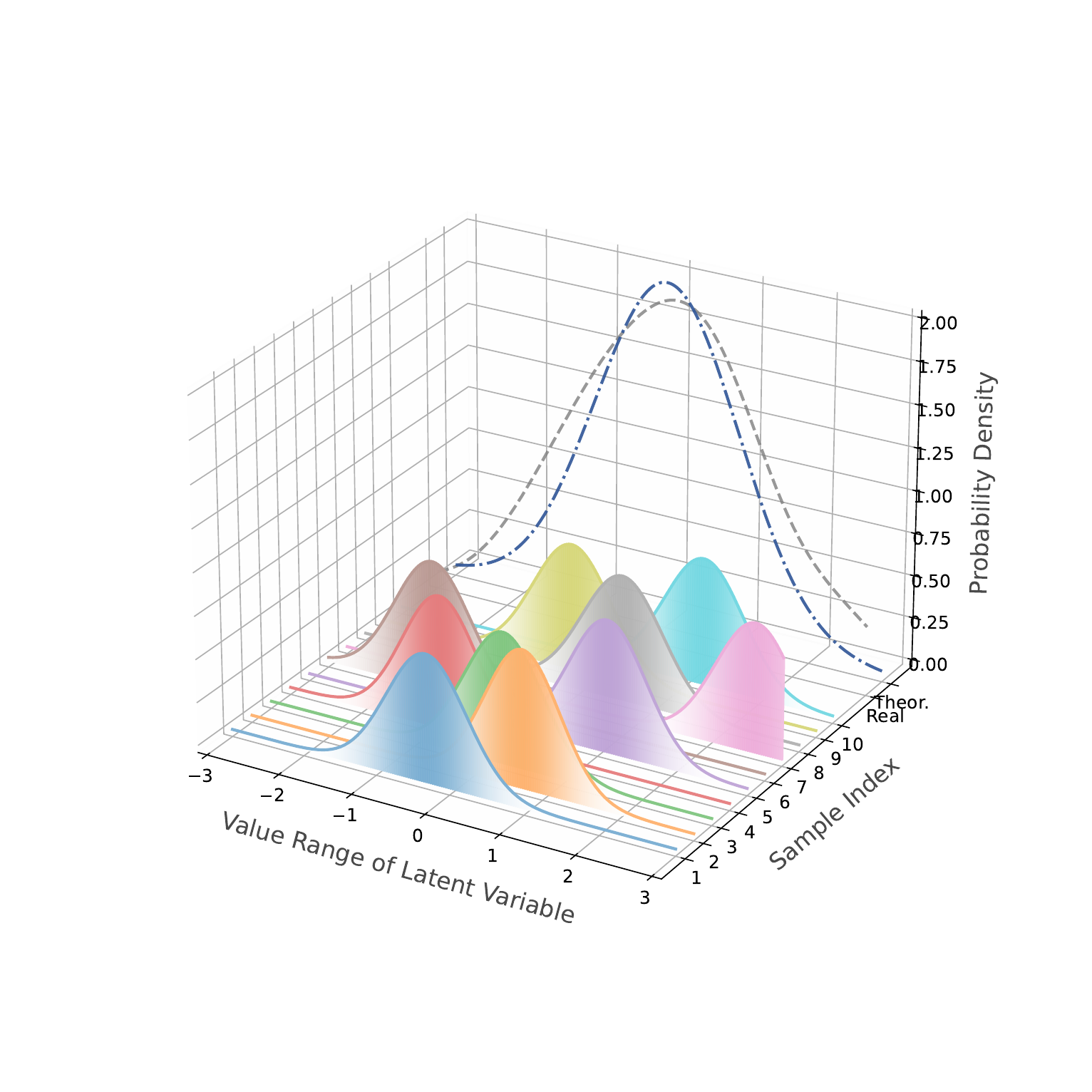}}
\subfigure[Latent space dimension of 64 in the LSUN 64 dataset. ]{\includegraphics[width=1.8in]{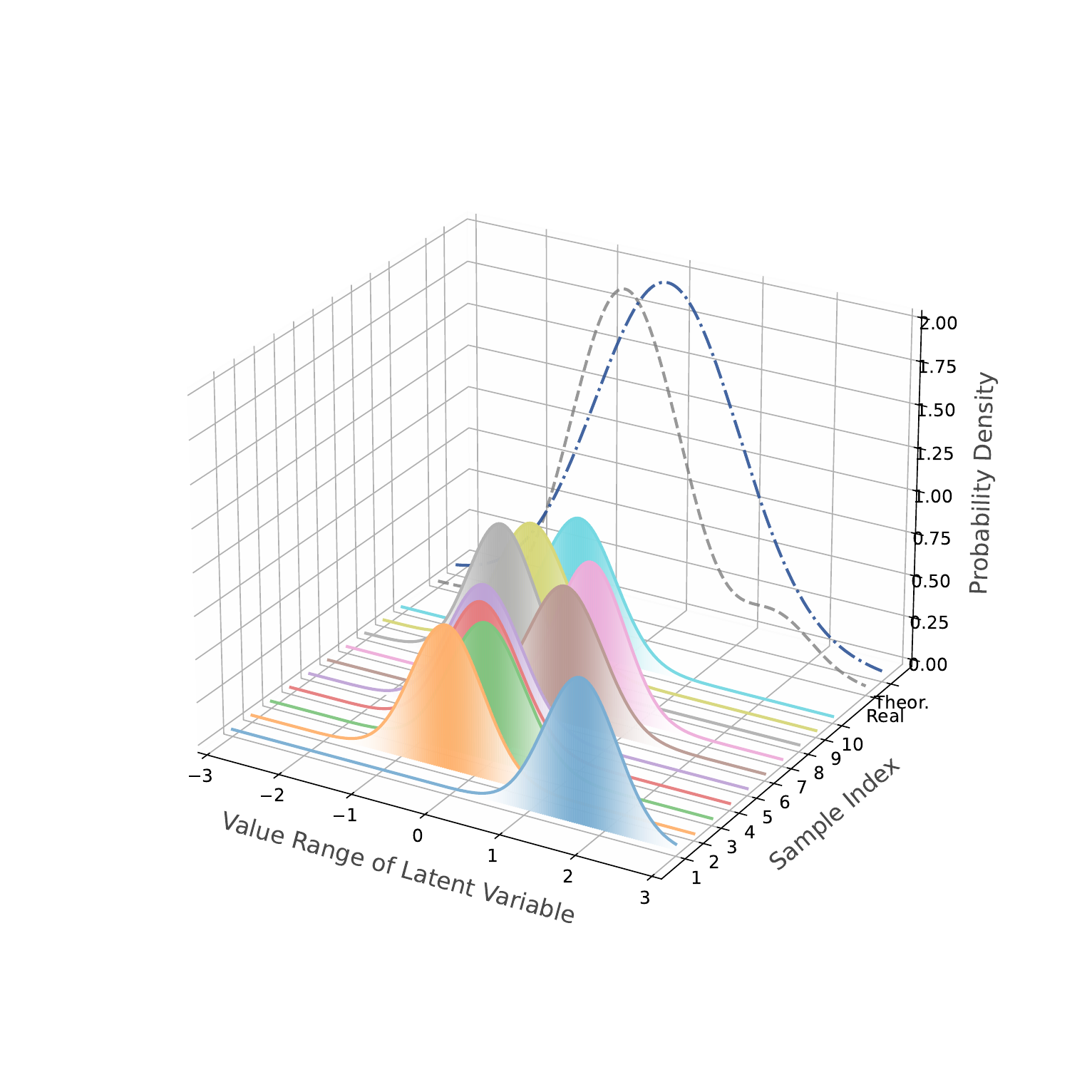}}
\subfigure[Latent space dimension of 32 in the CelebA dataset. ]{\includegraphics[width=1.8in]{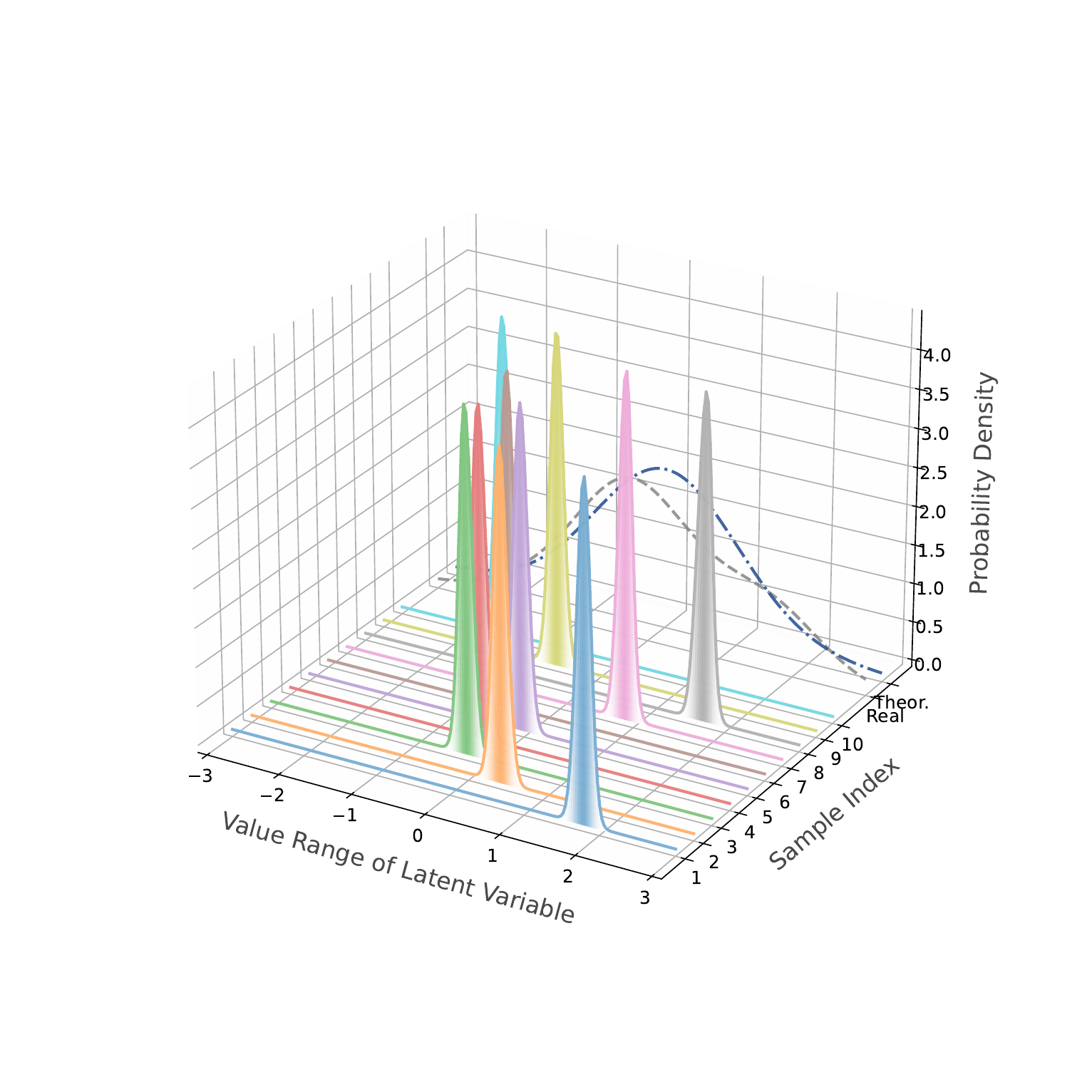}}
\subfigure[Latent space dimension of 32 in the CIFAR-10 dataset. ]{\includegraphics[width=1.8in]{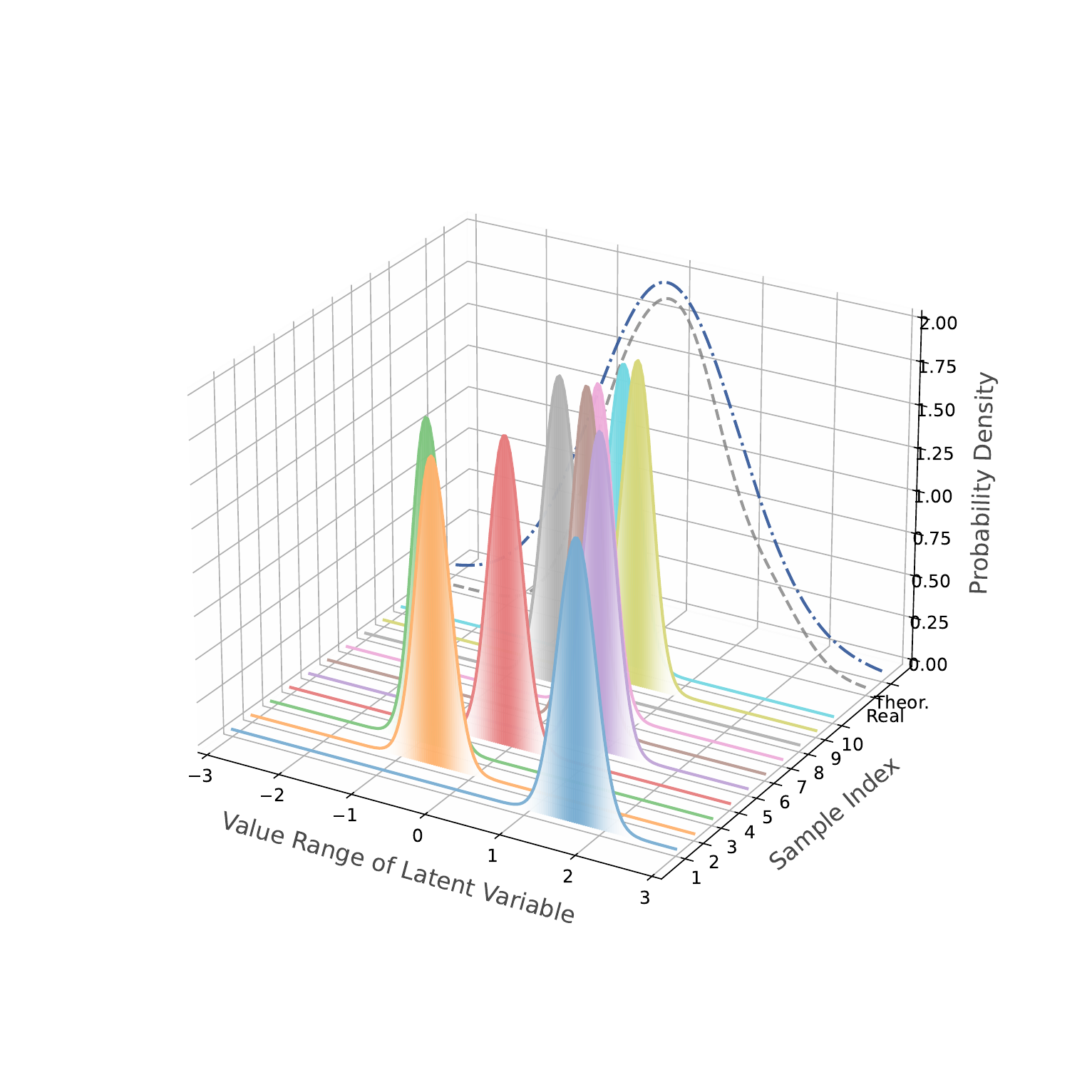}}
\subfigure[Latent space dimension of 32 in the LSUN 64 dataset. ]{\includegraphics[width=1.8in]{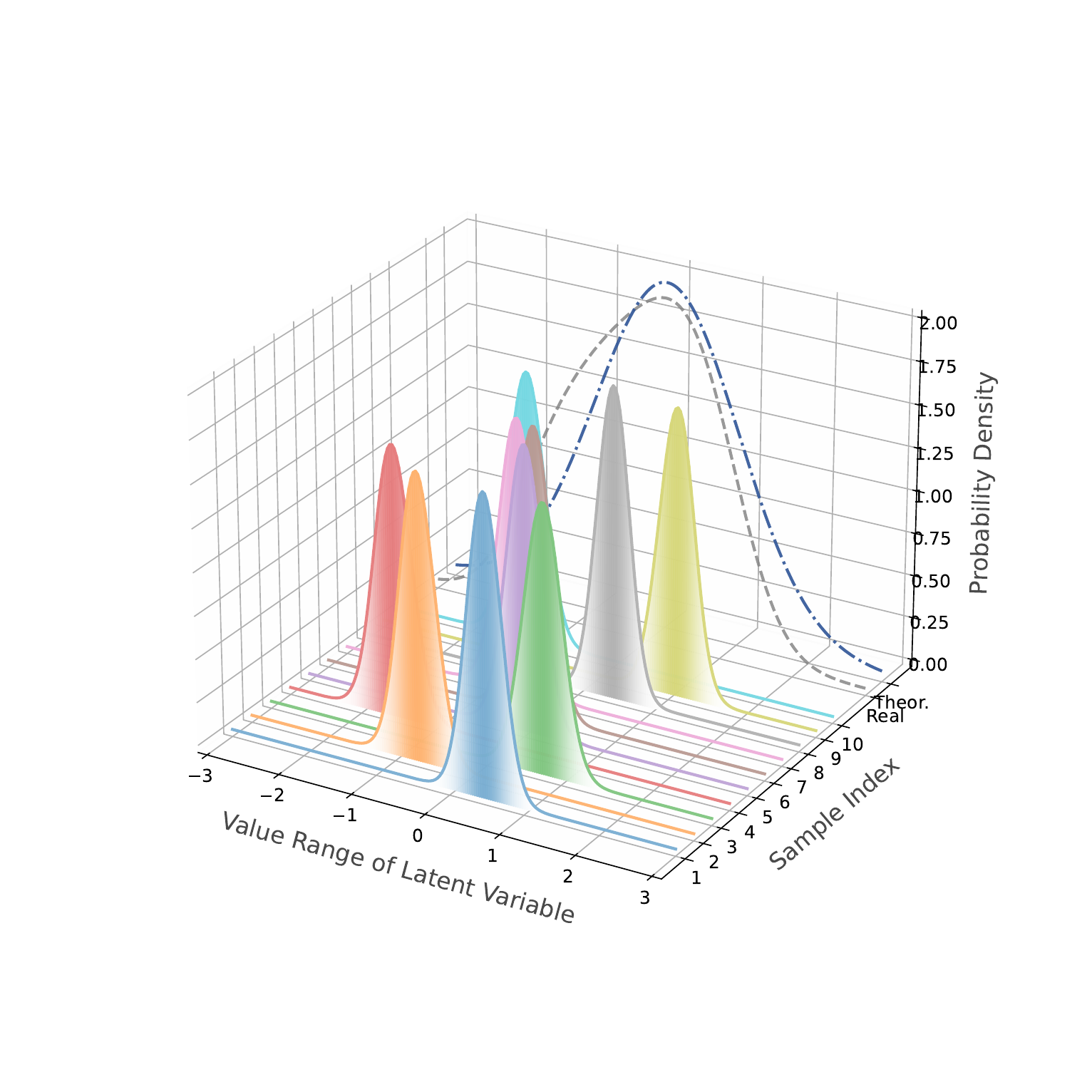}}
\caption{Visualization of sample distribution in the latent space. }
\end{figure}

We provide the variance and performance differences of the two models on two additional datasets (CIFAR-10 and LSUN 64) under different latent space dimensions as a supplement to Table \ref{tab:32}.

\begin{table}[H]
\centering
\small
\caption{Image generation performance comparison.}
\label{tab:3}
\renewcommand {\arraystretch}{1.2}
\resizebox{\textwidth}{!}{
\begin{tabular}{ c c c c c c c c c c c c}
\hline
\hline
\multirow{2}{*}{Dimension} &  & \multicolumn{4}{c}{CIFAR-10} & & \multicolumn{4}{c}{LSUN Church 64} \\
 \cline{2-6} \cline{8-12}
 & Var.&Lip& E\_FID$\downarrow$ &F\_FID$\downarrow$ &Impr.& & Var.&Lip & E\_FID$\downarrow$ &F\_FID$\downarrow$ &Impr. \\
\hline
8 &$0.0102$ &$44.32$&$146.621$  & $146.445$&$+0.12\%$& &$0.0159$ &$72.65$&$162.990$ &$161.914$ &$+0.66\%$ \\
\hline
16  &$0.0272$ &$34.61$&$123.991$  &$124.165$ &$-0.14\%$& &$0.0289$&$28.29$&$155.409$&$155.052$&$+0.23\%$ \\
\hline
32 &$0.0522$&$15.72$&$105.821$  & $107.054$&$-1.17\%$& &$0.0729$&$22.86$&$149.357$&$149.506$& $-0.10\%$ \\
\hline
64 & $0.3025$&$10.97$&$90.132$&$92.167$&$-2.70\%$ & &$0.2601$ &$14.27$&$143.064$ &$145.613$ &$-1.78\%$\\
\hline
128 &$0.5329$ &$8.20$&$86.151$  &$89.026$ &$-3.34\%$ & &$0.5062$ &$5.99$&$134.771$ &$138.200$ &$-2.54\%$  \\
\hline
\end{tabular}}
\end{table}

\subsection{Latent Space Walking}\label{lsw}
We conducted a latent space walking experiment to demonstrate the generative performance of our model. Specifically, we randomly sampled two latent space points, performed linear interpolation between them with 10 intermediate steps, resulting in 12 samples, and generated the corresponding images. This experiment helps to visualize how the model interpolates between different points in the latent space, showcasing the smoothness and continuity of the generated samples.

Experimental results demonstrate that our model generates significantly sharper images and more effectively handles variations in lighting, shadow, and object contours during latent-space walking, producing smooth and visually coherent transitions. In contrast, the Vanilla VAE consistently yields blurrier outputs and exhibits less natural scene changes.

\begin{figure}[H]
\centering
\vspace{-0.5cm}
\includegraphics[width=5.2in]{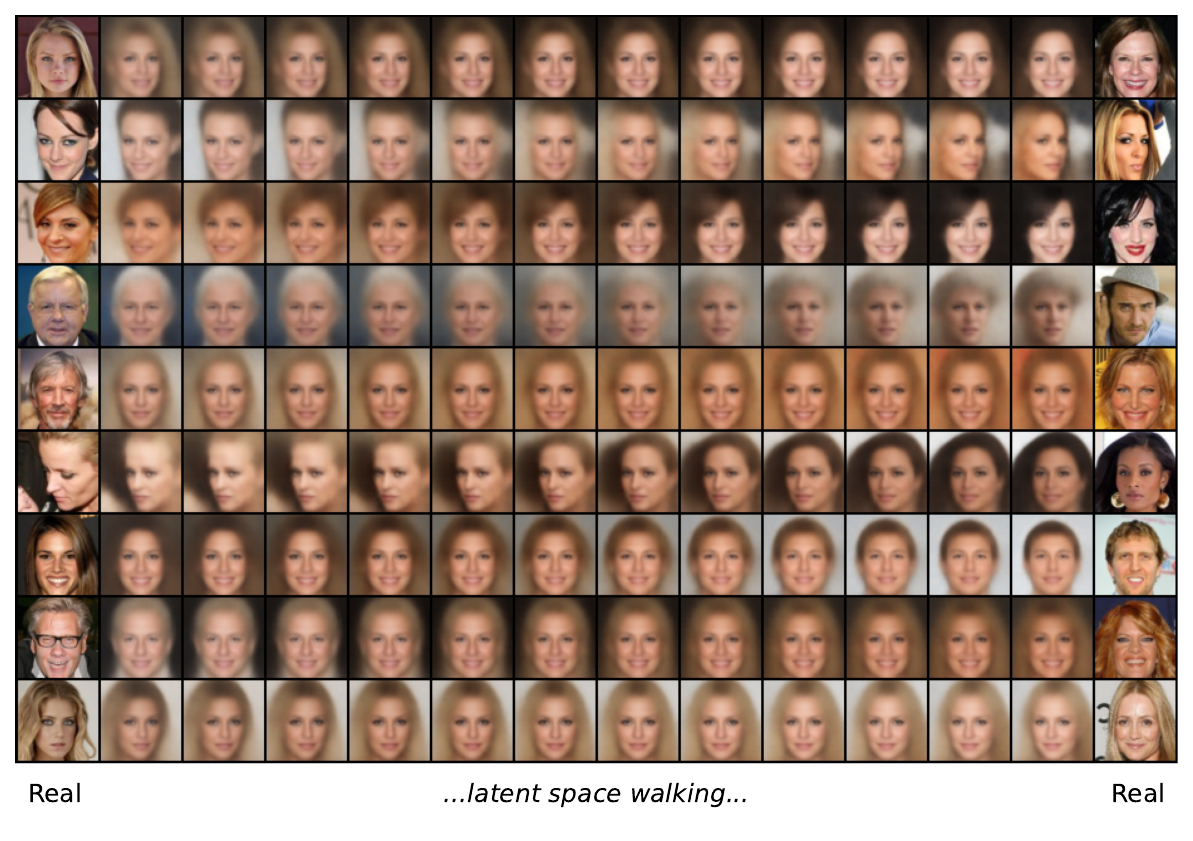}
\vspace{-0.3cm}
\caption{Latent space walking of VAE.}
\label{fig:cont}
\end{figure}

\begin{figure}[H]
\centering
\vspace{-0.5cm}
\includegraphics[width=5.2in]{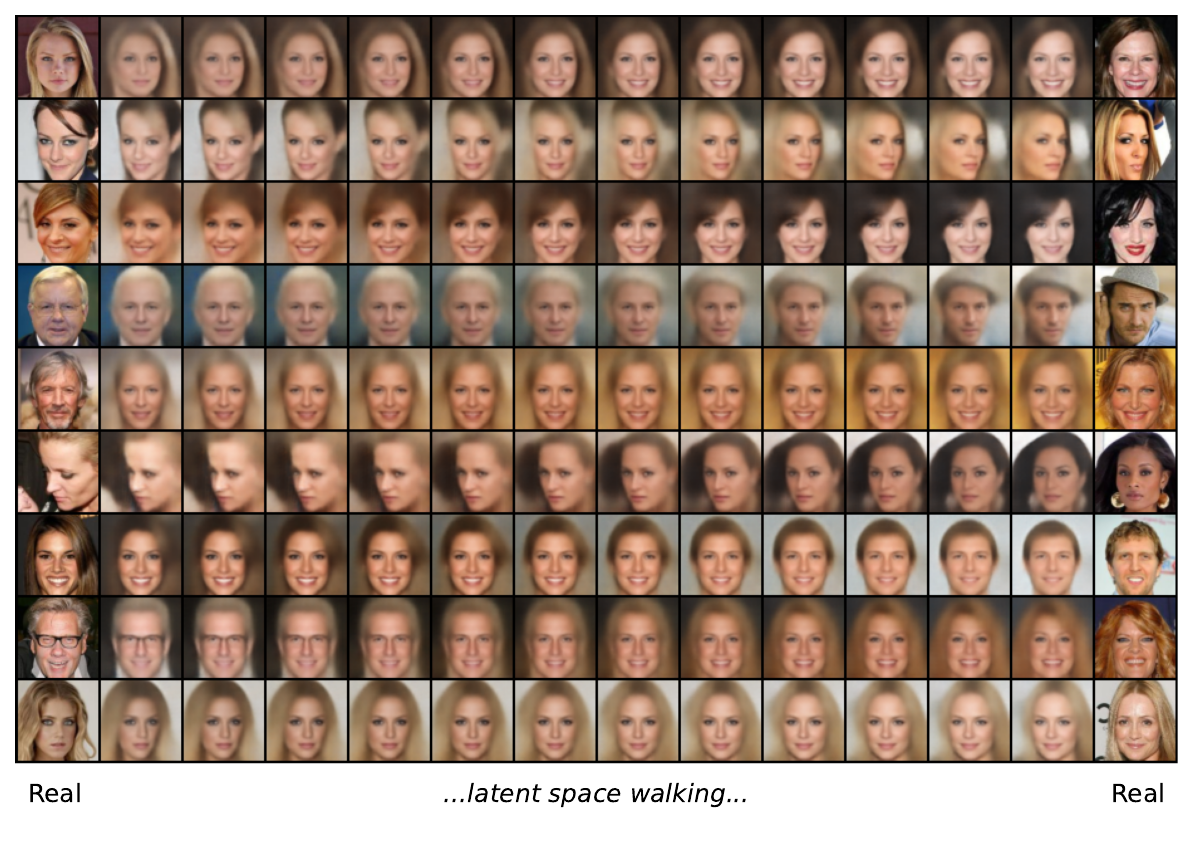}
\vspace{-0.3cm}
\caption{Latent space walking of \textit{EnVAE}.}
\label{fig:cont}
\end{figure}

\begin{figure}[H]
\centering
\vspace{-0.5cm}
\includegraphics[width=5.2in]{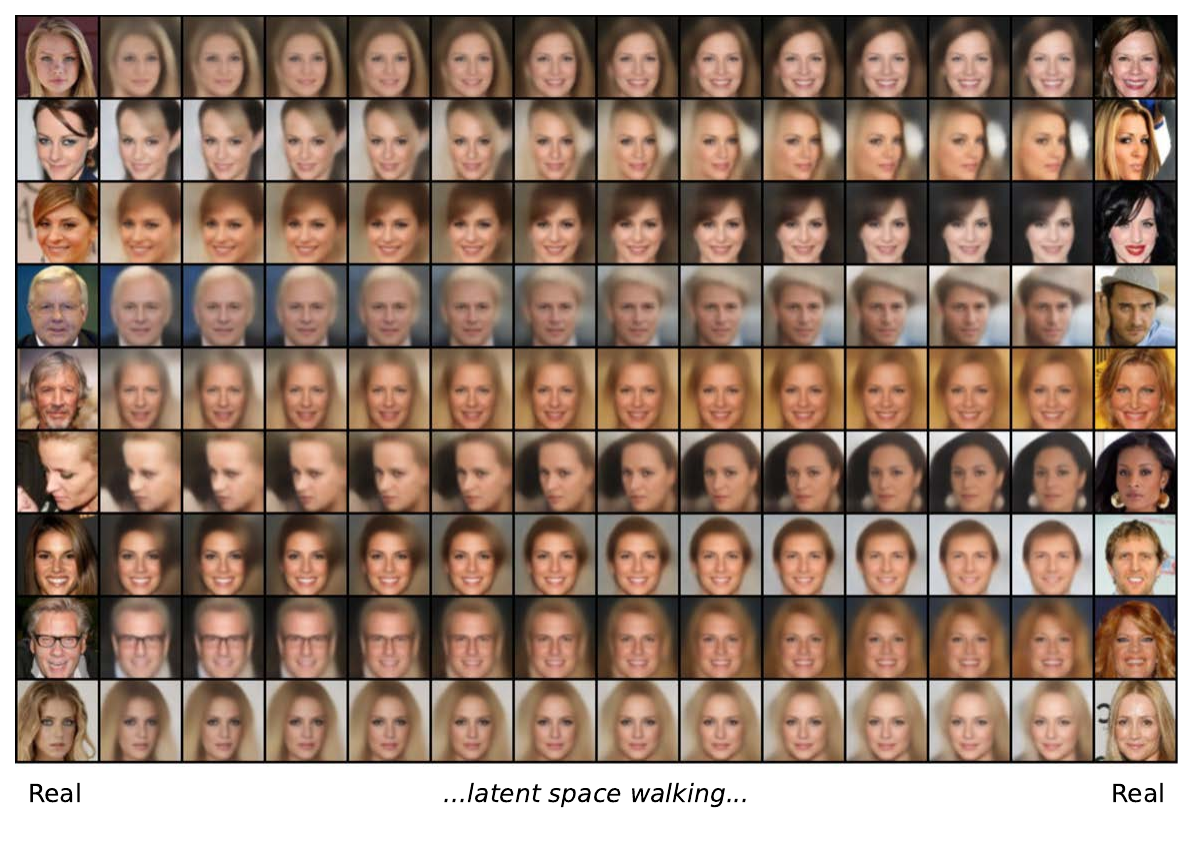}
\vspace{-0.3cm}
\caption{Latent space walking of \textit{FEnVAE}.}
\label{fig:cont}
\end{figure}

\subsection{Hyperparameter Walking Visualization Analysis}

In \textit{FEnVAE}, $\beta$ is the exponent of the reconstruction loss, determining the extent to which the loss function emphasizes the errors in different scales. A larger $\beta$ increases the weight assigned to large-scale errors.

To analyze their effects, we designed a hyperparameter walking experiment for $\beta$, visualizing how different values within a certain range impact both reconstruction quality and generated images.

\begin{figure}[H]
\centering
\includegraphics[width=3.2in]{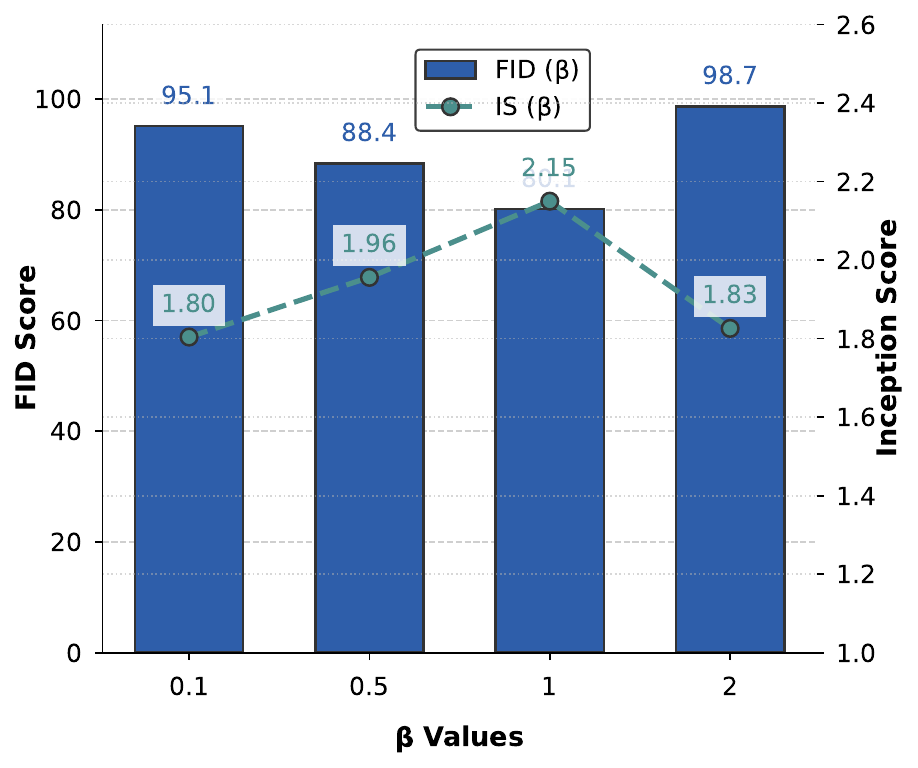}
\caption{FEnVAE performance of hyperparameter walking.}\label{f:24beta}
\end{figure}

We select 4 different $\beta$ values within the range $(0,2]$, $\beta \in \{0.1, 0.5, 1, 2\}$ (we cannot take 0, as it would make the model lose error perception), and visualize the reconstructed and generated images.

From Figure~\ref{f:24beta}, it can be observed that both excessively large and small values of $\beta$ negatively impact image generation performance. When $\beta$ is small, the generated images tend to be globally blurred, with the edges of the main subjects blending into the background, making them difficult to distinguish. Conversely, when $\beta$ is large, the images appear blurry with excessive local smoothing, and the background is prone to unreasonable shadow artifacts. From the FID and IS metrics, it can also be observed that the model's generation performance is more sensitive to excessively large values of $\beta$.

We believe that different values of $\beta$ influence the model’s emphasis on different types of image features, as $\beta$ is the exponent of the frequency domain energy score. A larger $\beta$ tends to make the model focus on frequency components where the generated image exhibits the largest discrepancy from the real image. Unlike time-domain distances, this discrepancy can occur in either high-frequency or low-frequency regions, depending on the characteristics of the data. Conversely, an excessively small $\beta$ leads the model to prioritize learning frequency components that are easier to refine while treating frequency components with large discrepancies as outliers and ignoring their associated information.

\begin{figure}[H]
\centering
\includegraphics[width=3.4in]{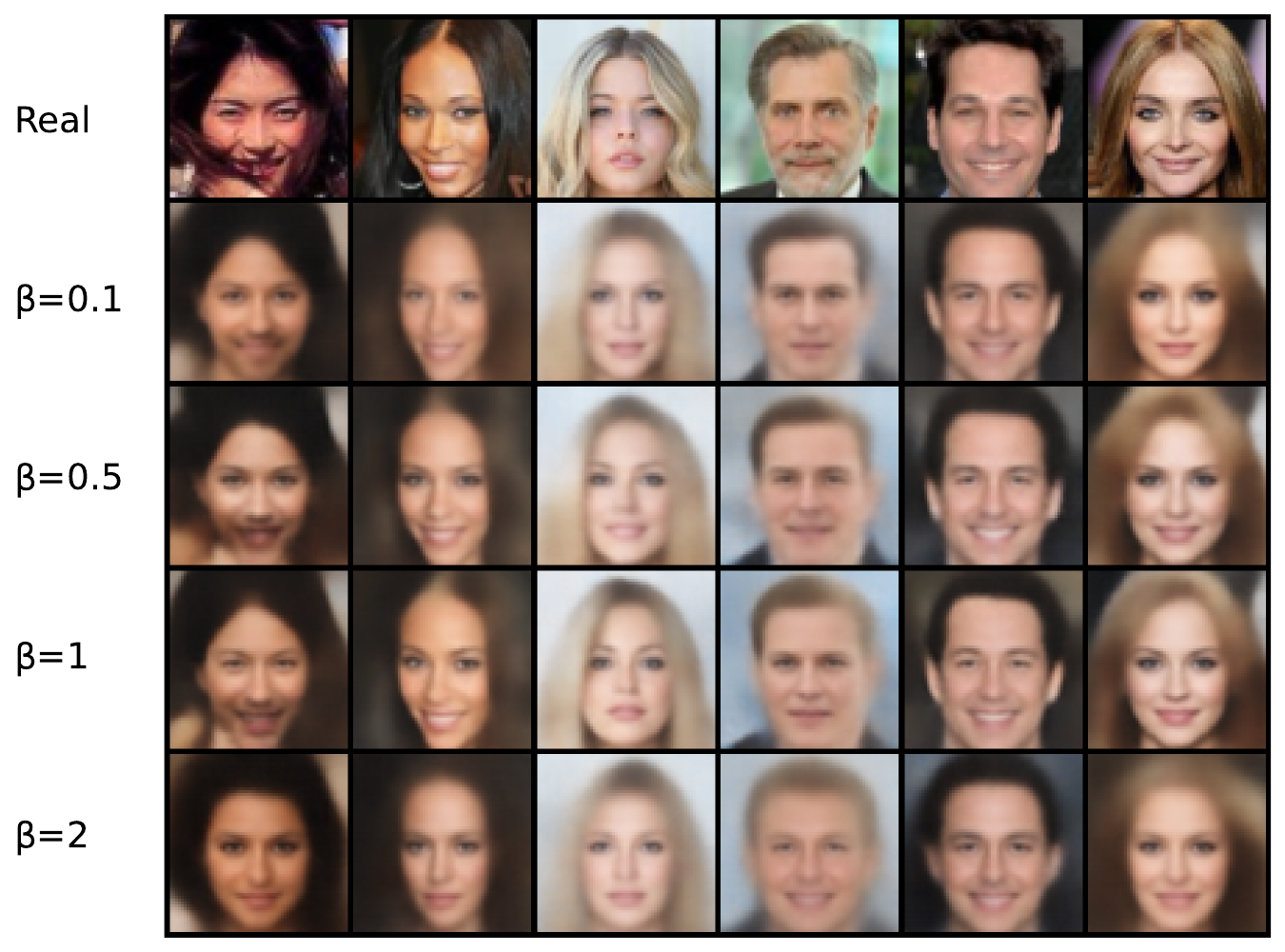}
\caption{Reconstructed images of $\beta$ walking.}
\label{fig:cont}
\end{figure}

\begin{figure}[H]
\centering
\includegraphics[width=3.4in]{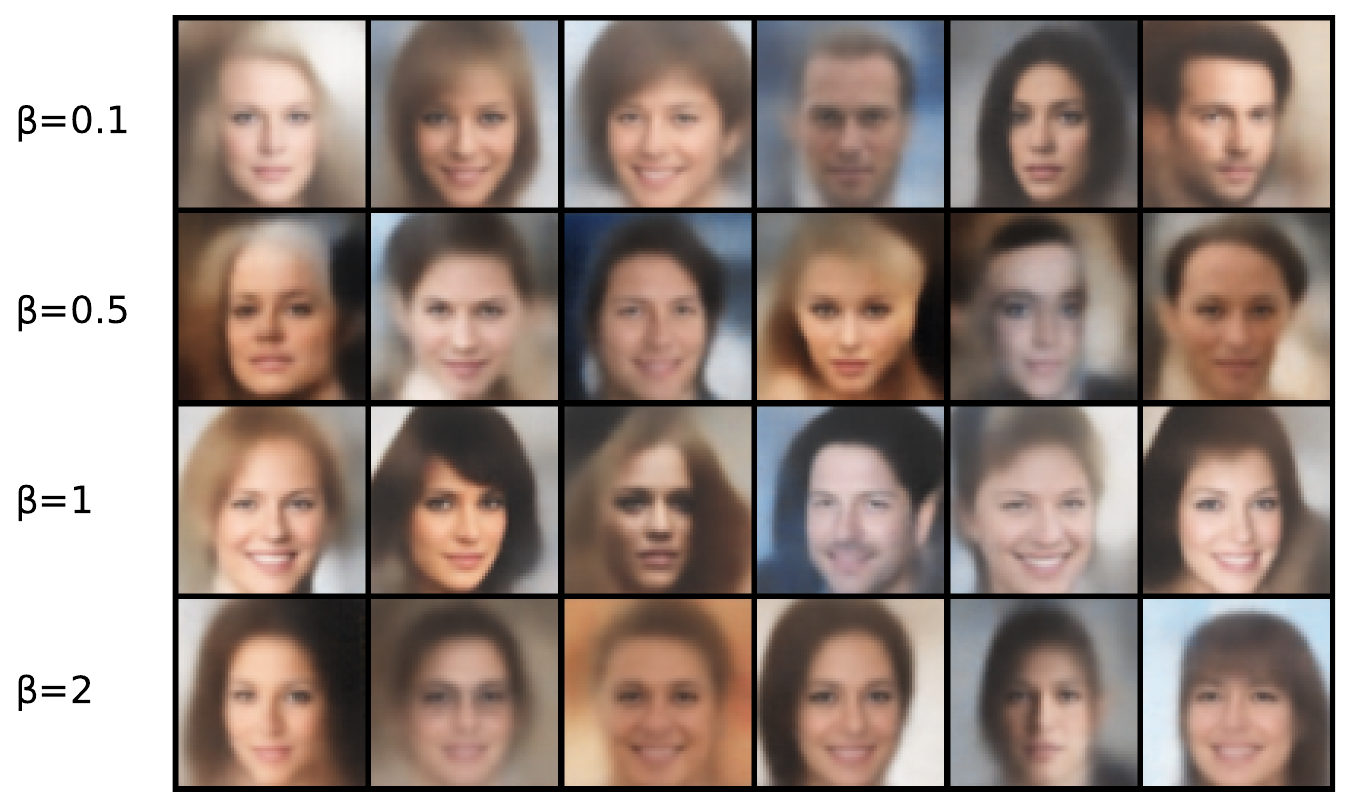}
\caption{Generated images of $\beta$ walking.}
\label{fig:cont}
\end{figure}

\end{document}